\documentclass[10pt,letterpaper]{article}
\usepackage[T1]{fontenc}
\usepackage{times}
\usepackage[textwidth=5.5in,textheight=9in,centering]{geometry}
\usepackage{amsmath,amssymb,booktabs,graphicx,array}
\usepackage{longtable}
\usepackage[authoryear,round]{natbib}
\usepackage{microtype}
\usepackage{hyperref,url}
\hypersetup{pdftitle={MultiEcho: An Experimental Science of Learned Worlds},pdfauthor={Meng Zhu, Airui Zhang},hidelinks}

\title{MultiEcho: An Experimental Science of Learned Worlds}
\author{%
Meng Zhu\\
SHIHAO XINRUI Group\\
\texttt{\href{mailto:mzqef@outlook.com}{mzqef@outlook.com}}
\and
Airui Zhang\\
SHIHAO XINRUI Group\\
\texttt{\href{mailto:maseath3927@outlook.com}{maseath3927@outlook.com}}
}
\date{}

\begin{document}
\maketitle

\begin{abstract}
World models can be studied as experimental systems with response laws of
their own. We introduce MultiEcho, a framework for estimating these laws
through controlled counterfactual interventions, delimiting their applicability,
and separately testing their physical correspondence. Across nine simulated
physical systems and seven frozen model configurations, three-reference
estimators predict complete intervention responses and recover intervention
parameters. Estimator selection uses discovery data only; frozen fits are
evaluated on validation and confirmation contexts. The experiments distinguish
response predictability, intervention readability and physical accuracy.
Responses can be locally describable yet poorly match physical effects
in the same target coordinates. Event-window, visibility and camera interventions
reveal conditional applicability, and paired generator configurations show reduced
readability under a scene prompt with stronger guidance. Magnitude sweeps expose
small image errors alongside large relative effect errors. An exact-reset
material experiment separates registered visible-response success from fixed-readout
failure on material-dependent futures at matched positions and velocities.
Exact finite-scale identities resolve odd and even response errors; first-order
remainder bounds specify when refined calibration converges. MultiEcho provides
an experimental basis for studying learned-world laws independently of, and in
relation to, physical laws.
\end{abstract}

\section{Introduction}
\label{sec:introduction}

World models support prediction, generation, and planning, and make learned worlds
available for experiment. Since stable, unfamiliar laws are themselves legitimate objects
of scientific inquiry, we may ask: what laws govern a learned world's responses to changing
conditions?

In experiments on the physical world, we typically control a small number of conditions and
observe selected effects of interest, while treating the remaining changes as
background. This works when prior knowledge tells us which observables are
relevant to the phenomenon under study. For a learned world, however, the
relevant effects may be numerous, interdependent, unfamiliar, or unknown in
advance. The MultiEcho framework therefore treats the complete
recorded intervention response as the experimental object and develops quantitative
analyses of its magnitude, geometry, predictability, and intervention identification.

We study seven frozen configurations of five released models, spanning
encoders, latent predictors and a video generator. The interface is
universal at both ends: every subject receives the same rendered 16-frame
clip, and its native output, whether slots, latents, tokens or pixels, is
recorded as a fixed-size array that the same fits and statistics operate on
directly. Encoders save the encoded clip, predictors and generators save
eight future frames, and each interface records the final input encoding or
the first future step. VideoSAUR
\citep{zadaianchuk2023}, an object-centric encoder trained on CLEVRER, gives
seven 128-D slots per frame, and C-JEPA \citep{nam2026} predicts the next
frame's slots from the last 15. LeWM \citep{maes2026}, a JEPA world model
trained on Reacher, encodes frames as 192-D states and predicts from three
states under zero actions. V-JEPA 2-AC \citep{assran2025} encodes two-frame
tubelets as 256 tokens of 1,408 channels and predicts frames 16--17 under
zero actions and states. Cosmos3-Nano \citep{nvidia2025}, a 16B diffusion
world model, encodes the clip into five 48-channel Wan VAE latent slices and
generates frames 16--23 from them. The Cosmos tokenizer keeps the last slice,
covering frames 12--15. Cosmos null uses an empty prompt at guidance~1,
Cosmos text a scene prompt at guidance~6, and both keep generated frame~16.
Simulator frames 15 and 16 serve as RGB references
(Appendices~\ref{app:design} and \ref{app:simulatorreference}).

We realize interventions through simulated input histories, since simulation
makes matched counterfactual conditions reproducible and controllable, including
changes that are subtle or complex to implement physically. At fixed context
$c$, we compare the frozen subject's outcome $Y=F(c,u)$ with its reference at
$u_0$. Three labeled observations $\mathcal D_c$ supply the reference and
the endpoints of the intervention range in each query context. The forward fit
$\widehat{\Delta Y}=f(\mathcal D_c,u)\approx F(c,u)-F(c,u_0)$
describes how its complete output changes with the intervention. The reverse
fit $\widehat u=g(\mathcal D_c,Y)$ asks whether the parameter can be recovered
from an observed response.

An experimental description must establish its range of applicability.
Event-window studies locate when an intervention becomes readable and show
that encoder and predictor windows can differ. Spatial conditions include
viewpoint, object geometry, and scene configuration. These dimensions locate
response laws within an experimental domain.

Along the intervention axis, we probe locality by successively halving the
log-parameter change in spring, heat, and bounce. Smaller effects can make
absolute errors decrease without improving relative accuracy. Finite
differences separate sensitivity errors from effects shared by opposite
interventions. Remainder bounds state when refined calibration converges.

Response laws differ strongly and unevenly across subjects, scenes and
contexts. Median reverse $R^2$ ranges from 0.208 at VideoSAUR to 0.986 at
the Cosmos tokenizer. The LeWM encoder reaches forward $R^2$ of 0.975 in
heat and 0.231 in mechanics plus optics, and C-JEPA heat distance
association is 0.861 for context-averaged responses and 0.186 within a
typical context. Cosmos null reaches reverse $R^2$ of 0.776 in discovery and
0.836 in confirmation, Cosmos text 0.530 and 0.391, and the text
configuration's heat effect errors plateau 2.5 times higher.

Physical correspondence requires a separate comparison with simulator RGB
or encoded true futures. Cosmos heat effect errors approach an absolute
plateau as the true effect shrinks. C-JEPA and LeWM effect errors remain about
as large as the true effects at all sampled magnitudes. Crucially, the separation persists in
one observation space: LeWM heat responses can be described substantially
better than their physical effects are matched.

A separate viscoelastic exact-reset experiment makes a specific distinction
between geometric and physical evidence. With positions and velocities matched
but material damping retained, C-JEPA and AC meet registered visible-response
criteria with forward $R^2=0.125/0.031$, yet fail the material-future criterion
under the fixed readout (Appendix~\ref{app:material}). This contrast supports the
conclusion: geometric regularities alone are insufficient evidence of
physical-law emergence.

Through measurements, MultiEcho estimates learned worlds' laws, investigates
their applicability and locality, and compares them with references from the
physical world. We aim to provide MultiEcho as a multidimensional experimental
framework spanning time, space, and counterfactual interventions for
investigating inherent laws in learned worlds.

\section{Related Work}

\paragraph{Interrogating learned representations.}
Counterfactual World Modeling (CWM) extracts flow, segmentation, and depth from
a masked video predictor using visual prompts and Jacobians \citep{bear2023}.
Physical CWM evaluates contact and future plausibility \citep{venkatesh2024}.
Physics Emergence Zone (PEZ) probes encoder layers and steers decoded motion
\citep{joseph2026}, and PhyIP fits equations to energy and force readouts from
numerical world models \citep{interno2026}. Board-state probes,
violation-of-expectation tests, sequence metrics, and inductive-bias probes ask
whether predictors form world models, and orbit-trained predictors fail to
apply Newtonian mechanics \citep{li2023,garrido2025,vafa2024,vafa2025}.
MultiEcho varies physical
conditions and describes complete responses before choosing a physical readout.

\paragraph{Counterfactual physical prediction.}
CoPhy and Filtered-CoPhy learn alternative futures from factual histories and
edited initial conditions \citep{baradel2020,janny2022}. PhysEditWorld evaluates
gravity-conditioned generation before and after adaptation \citep{hu2026}.
Physion++ extends Physion's contact task with property-revealing histories and
matched visible starts \citep{bear2021,tung2023}, a precedent for our exact-reset
design. CRONOS compares generation quality across matched visual and physical
variants \citep{begiristain2026}. Here the subject is already frozen: the
fitted function describes its response.

\paragraph{Physical behavior and generalization.}
IntPhys 2, VideoPhy, and PhyGenBench evaluate event plausibility and generated
physical behavior \citep{bordes2025,bansal2024,meng2025}.
Physics-IQ, WorldBench, and Morpheus evaluate continuation fidelity, physical
parameters, and prescribed equations \citep{motamed2026,upadhyay2026,tragoudaras2026}.
PhyWorld and PISA study generalization over physical conditions
\citep{kang2025,li2025}. We distinguish these physical
targets from learnability of a subject's own intervention-response map.

\paragraph{Law discovery and system identification.}
Symbolic regression distills laws from experimental data and from trained
networks \citep{schmidt2009,cranmer2020}. SINDy selects sparse equation terms
\citep{brunton2016}, and later methods infer them jointly with coordinates,
constants, or modes \citep{gao2024,liu2024}. Visual system identification
recovers video-generating geometry, physical properties, and constitutive laws
\citep{kaneko2024,zhao2025}. MultiEcho targets the frozen learned system's own
intervention-response law.

\section{Experimental Method}
\label{sec:method}

A MultiEcho experiment begins with a physical question and a family of alternative
conditions that share a matched context while varying the condition of interest.
We run the frozen model on each branch and record observations through the same
interface. Differences from the reference branch define the intervention response.
An intervention can produce coupled changes across the recorded output, so a small
set of predefined summary observables may capture only part of its effect. We therefore
treat the complete recorded response as the experimental object and retain all
coordinates when evaluating forward predictions. Each response occupies a fixed
number of coordinates set by its interface, such as one frame of pixels, however
many or complex the interventions are. Although the present experiments
use simple interventions, the paired construction can also accommodate more complex
changes when matched contexts and comparable observations are available.

\begin{figure}[t]
\centering
\includegraphics[width=\linewidth]{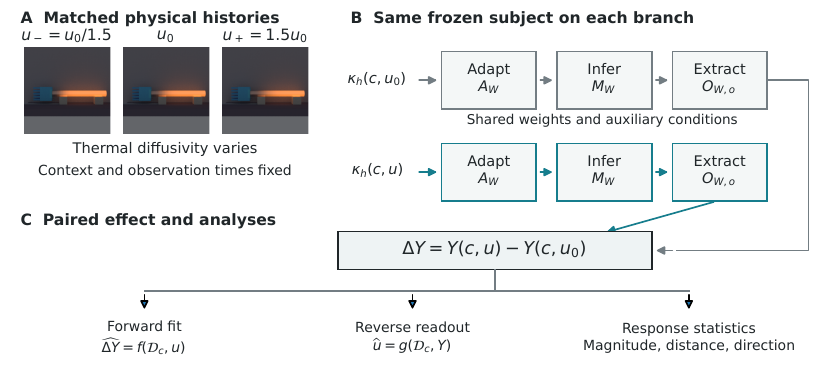}
\caption{Constructing a counterfactual experiment. (A) Final input frames at
diffusivities $u_-$, $u_0$ and $u_+$, one set of coarse heat references from
Section~\ref{sec:scales}. (B) Each branch input
$\kappa_h(c,u)$ passes through the same adapter $A_W$, frozen model $M_W$ and
observation map $O_{W,o}$ (Equation~\ref{eq:response}). (C) The paired effect
$\Delta Y$ enters the forward fit, reverse readout and response
statistics.}
\label{fig:method}
\end{figure}

\paragraph{The experimental unit.}
Let $c$ collect the conditions held fixed, including non-target physical and
initial-state coordinates, environment, and seed policy. The intervention
coordinate $u$ has reference value $u_0$. An intervention handle $h$ can target
inputs, actions, or internal states such as selected neuron activations,
depending on the subject's access. For the video-input handle studied here,
the compiler $\kappa_h(c,u)$ simulates and renders each branch history.
Subject $W$ includes the frozen model, adapter, and inference settings.
The adapter $A_W(x;c)$ prepares input
$x$ and auxiliary conditions, $M_W$ performs inference, and $O_{W,o}$ extracts
observation $o$. The paired measurement is
\begin{equation}
\begin{aligned}
Y_{W,o}(c,u)&=O_{W,o}\!\left(M_W\!\left(A_W(\kappa_h(c,u);c)\right)\right),\\
\Delta Y_{W,o}(c;u,u_0)&=Y_{W,o}(c,u)-Y_{W,o}(c,u_0).
\end{aligned}
\label{eq:response}
\end{equation}
The response $\Delta Y$ is the change across runs at a fixed observation
interface. Varying $u$ samples its dependence on the intervention. Action,
state, and text conditions remain fixed within a pair,
so the physical change reaches the model through the input history
(Figure~\ref{fig:method}).

\paragraph{Response functions and observables.}
For a fixed subject and observation, write the observable law as $Y=F(c,u)$.
Reference and intervention share one observation interface and time. Let
$\mathcal D_c=\{(u_-,Y_-),(u_0,Y_0),(u_+,Y_+)\}$ contain exactly three
measurements, with $u_-<u_0<u_+$. The forward fit approximates this law in
counterfactual response form:
\begin{equation}
\widehat{\Delta Y}=f(\mathcal D_c,u)\approx F(c,u)-F(c,u_0),
\qquad Y_0=F(c,u_0).
\label{eq:lawfit}
\end{equation}

\paragraph{Reverse intervention readout.}
A reverse readout $\widehat u=g(\mathcal D_c,Y)$ estimates the intervention
from the query and the same three labeled references. It asks how precisely
an intervention parameter can be recovered from the subject's response. Given
a query observation $Y$ and the same three labeled references $\mathcal D_c$,
we estimate $\hat u = g(\mathcal D_c,Y)$. The references provide a local
calibration linking response geometry to parameter values. Recovery
accuracy on held-out queries measures intervention readability under
this calibration and readout. This complements forward fitting: a response
may contain information that distinguishes parameter values even when predicting
all of its components is difficult. A common geometric descriptor allows the same
readout procedure to assess readability across observation interfaces with
different output dimensions.

\paragraph{Statistics of the full response.}
Distance association measures whether larger parameter separations produce
larger response separations. We compute it within each context and separately
for responses averaged across contexts. Direction consistency measures whether
increasing the parameter moves observations in a common direction across
contexts (Appendix~\ref{app:geometrydefinitions}). These describe organization,
while total squared response magnitude measures the size of the effect.

\paragraph{Experimental program.}
The atlas covers 24 physical axes in nine scenes, each with 24 discovery,
8 validation and 8 confirmation contexts, and re-renders every discovery
context from a moved camera (Section~\ref{sec:relations}). Bounce
event-window and visibility studies, magnitude sweeps against true futures
(Section~\ref{sec:scales}), an RGB-trained inverse and a viscoelastic
exact reset (Section~\ref{sec:physical}) complete the program, for 268,506
saved frozen-subject runs. The atlas yields 105,600 response records and
84,183 eligible query records (Table~\ref{tab:accounting}). The
event-window, visibility and material studies retain their original
readouts (Appendix~\ref{app:historical}), and Appendix~\ref{app:tools}
collects reusable tools.

\section{Approximating World-Local Response Laws}
\label{sec:relations}

\subsection{Response Organization across Contexts}

Averaging across contexts can make responses look more ordered while
discarding most of their separation. For thermal diffusivity, C-JEPA's mean
response has distance association 0.861, while the median association within
individual contexts is only 0.186. The mean response keeps only 8.2\% of the
average squared separation between levels within a context. LeWM's encoder
and predictor remain organized within contexts, with median associations of
0.995 and 0.996, and their mean responses keep 80.0\% and 79.1\% of this
separation.

Regular distances can likewise hide changing directions. For thermal
diffusivity, AC's responses at
different levels are nearly equidistant, and consecutive steps along the
parameter turn by about $119^\circ$, compared with $16^\circ$ at LeWM
(Appendix~\ref{app:contextgeometry}). Each new level thus moves AC's response
in a new direction. AC's distances are the most uniform among the learned
interfaces on 21 of 24 axes, and its median turns exceed $93^\circ$ on every axis.
An organized mean or a regular distance pattern
therefore does not show that individual responses are predictable.

\subsection{From Response Structure to Predictive Laws}

Table~\ref{tab:scenes} gives each observation's forward and reverse $R^2$
from six context-held-out folds of the common three-reference procedures,
with every native coordinate in the forward loss, together with the two
geometric statistics. Each statistic is also compared with
first-future RGB on the same axes. (Table~\ref{tab:neuralcontrols}).

\begin{table}[h]
\caption{Response statistics of every observation. Entries are medians over the same 24 physical axes. Forward and reverse $R^2$ come from the common three-reference neural procedures under six context-held-out folds. Within $\rho$ is the median within-context distance association, and Direction is direction consistency. Below FF gives the percentage of axes on which the observation scores lower than simulator first-future RGB, compared before rounding. Negative scores are included.}
\label{tab:scenes}
\centering\small
\setlength{\tabcolsep}{3pt}
\begin{tabular}{@{}lrrrrrrrr@{}}
\toprule
& \multicolumn{2}{c}{Forward $R^2$} & \multicolumn{2}{c}{Reverse $R^2$} & \multicolumn{2}{c}{Within $\rho$} & \multicolumn{2}{c}{Direction}\\
\cmidrule(lr){2-3}\cmidrule(lr){4-5}\cmidrule(lr){6-7}\cmidrule(l){8-9}
Observation & Median & Below FF & Median & Below FF & Median & Below FF & Median & Below FF\\
\midrule
VideoSAUR enc. & 0.344 & 91.7 & 0.208 & 100.0 & 0.392 & 100.0 & 0.315 & 75.0\\
C-JEPA pred. & 0.363 & 83.3 & 0.455 & 95.8 & 0.524 & 91.7 & 0.427 & 41.7\\
LeWM enc. & 0.619 & 33.3 & 0.659 & 100.0 & 0.667 & 70.8 & 0.508 & 33.3\\
LeWM pred. & 0.742 & 16.7 & 0.760 & 87.5 & 0.782 & 58.3 & 0.475 & 25.0\\
AC enc. & 0.359 & 95.8 & 0.898 & 83.3 & 0.848 & 45.8 & 0.401 & 41.7\\
AC pred. & 0.350 & 95.8 & 0.885 & 83.3 & 0.822 & 54.2 & 0.409 & 41.7\\
Cosmos tokenizer & 0.439 & 58.3 & 0.986 & 29.2 & 0.882 & 33.3 & 0.634 & 29.2\\
Cosmos null & 0.431 & 66.7 & 0.776 & 95.8 & 0.750 & 83.3 & 0.380 & 58.3\\
Cosmos text & 0.410 & 79.2 & 0.530 & 100.0 & 0.508 & 95.8 & 0.340 & 75.0\\
\midrule
RGB last input & 0.433 & 75.0 & 0.877 & 79.2 & 0.682 & 79.2 & 0.319 & 50.0\\
RGB first future & 0.469 & -- & 0.972 & -- & 0.798 & -- & 0.364 & --\\
\bottomrule
\end{tabular}
\end{table}

Each observation has its own profile, and its output form suggests possible
explanations.

LeWM's encoder and predictor are the only interfaces whose complete responses are more predictable
than first-future RGB on most axes. Their forward $R^2$ is 0.619 and 0.742,
below first-future RGB on only 33.3\% and 16.7\% of axes. Their reverse
$R^2$ is 0.659 and 0.760, below first-future RGB on 100\% and 87.5\% of
axes. Their compact global states may change smoothly with the intervention
while compressing parameter-specific detail.

AC recovers the parameter well and predicts complete responses poorly.
Reverse $R^2$ is 0.898 at the encoder and 0.885 at the predictor. Forward
$R^2$ is 0.359 and 0.350, below first-future RGB on 95.8\% of axes. In
heat, each new level moves AC's response in a new direction. Dense spatial
tokens may keep parameter-specific detail that separates the levels, while
such direction changes make complete responses hard to interpolate from
three references.

VideoSAUR's final slots have the lowest median on all four statistics,
including reverse $R^2$ 0.208 and within-context association 0.392. Both
fall below first-future RGB on every axis. Slot competition can reassign
image regions between slots when the input changes, which would break the
correspondence of slot coordinates across branches. C-JEPA reaches 0.455
and 0.524.

The Cosmos tokenizer is the only observation whose reverse exceeds
first-future RGB on most axes. Its reverse $R^2$ is 0.986, below the
reference on 29.2\% of axes. It also has the highest within-context
association, 0.882, and the highest direction consistency, 0.634. Its last
latent slice covers the final four of 17 input frames, so it can carry
motion that a single image lacks.

Cosmos null and text generate the first future frame, so they can be
compared directly with the true frame at the same time. Their reverse
$R^2$ is 0.776 and 0.530, below first-future RGB on 95.8\% and 100\% of
axes. Text adds one branch-invariant scene prompt and stronger guidance,
each configuration at a fixed seed, and its within-context association is
0.508 against 0.750. Paired on identical queries, text lowers reverse $R^2$
by a median of 0.219 in discovery, 0.396 in validation, 0.232 in
confirmation and 0.294 under the moved camera, with every interval above
zero, and forward $R^2$ by 0.026--0.064,
with intervals that include zero (Table~\ref{tab:promptpairing}).

\paragraph{Observation time changes what can be recovered.}
Last-input RGB has lower aggregate forward/reverse medians, 0.433/0.877,
than first-future RGB, 0.469/0.972. The mechanical atlas aligns the last
input to a key physical event: full extension in spring, B--C contact in collision,
and the second ground contact in bounce. The event frame can hide parameter
differences that appear in the very next frame. For spring stiffness, reverse
rises from 0.453 at the event frame to 0.998 at the next. For bounce
restitution, it rises from $-0.089$ to 0.966. Heat diffusivity remains near
0.999 at both frames. The pattern is consistent with time-dependent
observability. The complete input history can retain information absent
from its final image.

\subsection{The Domain of a Response Description}

Selection and optimization use discovery data only. We apply its six fits
unchanged to eight validation and eight confirmation contexts per scene
(Table~\ref{tab:stageprofiles}). Each
context supplies its own three references, with no refitting or multiplier
selection. LeWM remains the only model whose forward $R^2$ exceeds
first-future RGB on most axes, and the tokenizer the only observation whose
reverse $R^2$ does. AC keeps its gap between forward and reverse $R^2$.
The LeWM predictor's forward median falls from 0.742 in discovery to 0.692
in validation and 0.622 in confirmation. Reverse medians fall from 0.208
to 0.056 and 0.050 at VideoSAUR and from 0.530 to 0.431 and 0.391 at
Cosmos text.

\begin{table}[h]
\caption{Forward and reverse $R^2$ of the frozen discovery fits in eight validation and eight confirmation contexts per scene. Medians and Below FF follow Table~\ref{tab:scenes}, with first-future RGB from the same stage. Table~\ref{tab:neuralstages} adds intervals.}
\label{tab:stageprofiles}
\centering\small
\setlength{\tabcolsep}{3pt}
\begin{tabular}{@{}lrrrrrrrr@{}}
\toprule
& \multicolumn{4}{c}{Validation} & \multicolumn{4}{c}{Confirmation}\\
\cmidrule(lr){2-5}\cmidrule(l){6-9}
Observation & Forward & Below FF & Reverse & Below FF & Forward & Below FF & Reverse & Below FF\\
\midrule
VideoSAUR enc. & 0.351 & 79.2 & 0.056 & 100.0 & 0.332 & 79.2 & 0.050 & 100.0\\
C-JEPA pred. & 0.398 & 66.7 & 0.382 & 91.7 & 0.412 & 75.0 & 0.401 & 87.5\\
LeWM enc. & 0.541 & 37.5 & 0.586 & 100.0 & 0.543 & 33.3 & 0.703 & 91.7\\
LeWM pred. & 0.692 & 20.8 & 0.822 & 91.7 & 0.622 & 16.7 & 0.829 & 87.5\\
AC enc. & 0.363 & 91.7 & 0.904 & 79.2 & 0.359 & 95.8 & 0.877 & 75.0\\
AC pred. & 0.359 & 91.7 & 0.860 & 83.3 & 0.357 & 95.8 & 0.868 & 79.2\\
Cosmos tokenizer & 0.447 & 54.2 & 0.986 & 37.5 & 0.441 & 50.0 & 0.984 & 33.3\\
Cosmos null & 0.432 & 58.3 & 0.852 & 95.8 & 0.440 & 66.7 & 0.836 & 95.8\\
Cosmos text & 0.383 & 75.0 & 0.431 & 100.0 & 0.418 & 79.2 & 0.391 & 100.0\\
\midrule
RGB last input & 0.446 & 75.0 & 0.783 & 70.8 & 0.453 & 66.7 & 0.899 & 66.7\\
RGB first future & 0.468 & -- & 0.973 & -- & 0.477 & -- & 0.977 & --\\
\bottomrule
\end{tabular}
\end{table}

The descriptions are local: each context's three references set its output
coordinates and response scale, and the queries cover the eight interior
levels between its endpoints. Contexts vary visual and non-target
physical conditions, and the ensembles reuse nine heat baselines, three
mechanics-plus-optics clusters, three viscoelastic compression settings, and
one flow realization per Reynolds level. Fluid contexts thus differ only in
tracers, camera and appearance. Across them, reverse $R^2$ in all three
stages reaches 0.941--0.980 for the tokenizer, both Cosmos configurations and
first-future RGB, 0.870--0.881 for the AC encoder, 0.551--0.805 for LeWM and
the AC predictor, 0.117--0.481 for C-JEPA, and at most 0.198 for VideoSAUR.

\subsection{Temporal and Visual Applicability}

The separate bounce event-window study asks when a condition can be read
from a history. C-JEPA, both AC interfaces, and tokenizer satisfy the
registered pattern in discovery, validation, and confirmation: post-contact
restitution is readable, pre-contact restitution is unsupported, and pre-contact
gravity and height are readable. LeWM's gravity readout has active-only
support at its encoder and event-free-only support at its predictor.
These original within-window tests
locate an interface's temporal applicability (Appendix~\ref{app:temporal}).
A bounce visibility study lowers the ball's Weber contrast from 0.3 to 0
while keeping its shadow (Table~\ref{tab:visibility}). Post-contact restitution reaches
$R^2=0.778$ from the shadow alone at the tokenizer and 0.19--0.23 at AC,
which saturates at 0.66--0.71 from the first visible level. C-JEPA stays
negative until the third level. Different interfaces thus read the same
parameter through different visual channels.

Every discovery context was also rendered with unchanged physics from a
camera moved by $45^\circ$ in five scenes and 7--$12^\circ$ in three, with
a combined background, lighting and camera change in heat. Scored by the
fold that held it out, with three references from the moved view, every
interface matches or exceeds its canonical forward $R^2$ on at least 13 of
24 axes, and forward medians change by $-0.06$ to $+0.08$
(Table~\ref{tab:viewpoint}). The forward descriptions thus carry across
these camera changes. Reverse is lower on 12 to 15 axes per interface and
falls most at the LeWM encoder, from 0.659 to 0.456, and at Cosmos text,
from 0.530 to 0.417.

\section{Response Laws across Physical Scales}
\label{sec:scales}

The spring-stiffness, thermal-diffusivity, and bounce-gravity experiments
compare model predictions with the actual physical continuation. Each scene
has 24 contexts and three reference values. For positive parameter $u$ and
reference $u_\ast$, let $s=\log(u/u_\ast)$. Queried
offsets are $\pm\delta_0/j$, with $\delta_0=\log(1.5)$ and
$j\in\{2,4,8,16,32\}$. Observation times and image dimensions stay fixed.

\begin{figure}[t]
\centering
\includegraphics[width=\linewidth]{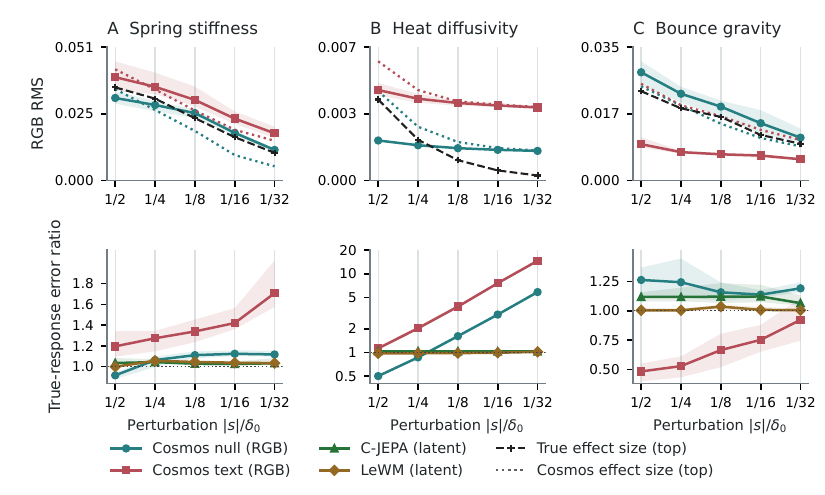}
\caption{Model predictions compared with true physical futures across
intervention magnitudes. Top: true effect size $\|\Delta T_W\|_{\rm RMS}$
(dashed, no model involved), Cosmos's own effect size $\|\Delta Z_W\|_{\rm RMS}$
(dotted) and the error of its effect, $\|\Delta Z_W-\Delta T_W\|_{\rm RMS}$
(solid). Per query, error over true effect size is $e_W$ (bottom), in
RGB or native target coordinates. Each point summarizes 144 queries.
Bands are conditional 95\% context-block intervals.}
\label{fig:scales}
\end{figure}

\paragraph{Whole futures and counterfactual effects.}
Let $Z_W(s)$ be the first future predicted by the frozen model itself and
$T_W(s)$ the true next frame in the same observation space: simulator RGB for
Cosmos null/text, and its encoding in the predictor's native target space for
C-JEPA and LeWM (Appendix~\ref{app:directfuture}). With
$\Delta Z_W(s)=Z_W(s)-Z_W(0)$ and $\Delta T_W(s)=T_W(s)-T_W(0)$, we measure the
whole-future error $E_W$ and the relative effect error $e_W$:
\begin{equation}
E_W(s)=\|Z_W(s)-T_W(s)\|_{\rm RMS},\qquad
e_W(s)=\frac{\|\Delta Z_W(s)-\Delta T_W(s)\|_{\rm RMS}}
{\|\Delta T_W(s)\|_{\rm RMS}},
\label{eq:trueeffect}
\end{equation}
where $\|x\|_{\rm RMS}=\|x\|_2/\sqrt d$ over the $d$ native coordinates.
The numerator of $e_W$ is the absolute effect error.

\paragraph{Absolute plateaus and precision.}
Cosmos heat effect errors level off near 0.00146/0.00359 RGB RMS for
null/text as the true effect halves at each step to 0.00025. The models' own
effects level off at 0.00148/0.00361, and their cosine with the true effect
falls from 0.90/0.65 to 0.17/0.07, so the fine-scale error is essentially the
model's own change (Figure~\ref{fig:scales}). The finest ratios $e_W$ of
5.843/14.605 reflect this plateau. Re-rendering the same reference
input in an independent stream changes the generated frame by
0.00329/0.00532 RMS, above the plateau. Rounding the true frames to
8-bit, the precision of the generated frames, already changes their effect
by 0.00033 RMS, so an exact 8-bit copy of the truth has $e_W=1.335$.
Discounting one 8-bit step
per coordinate leaves 0.00017 of null's plateau and 0.00175 of text's
(Table~\ref{tab:precision}). Null's plateau thus lies almost entirely within
this step, and text's exceeds it.

\paragraph{Amplitude, direction, and offsets.}
C-JEPA's $e_W$ spans 1.025--1.121 and LeWM's 0.972--1.056 across all three
scenes and magnitudes. At the finest heat scale, their amplitude ratios
$A=\|\Delta Z_W\|/\|\Delta T_W\|$ are 0.248/0.468 and their cosines $c$ with
the true effect are 0.000/0.216, so the identity $e^2=(A-1)^2+2A(1-c)$ of
Section~\ref{sec:theory} places $e_W$ near one. Predicting that the effect in the last input frame persists
gives smaller ratios, 0.798/0.368 for these latents and 0.126 in RGB.
In heat, the whole-future errors $E_W$ stay near 0.416--0.418 and 0.426 at
every magnitude. Subtracting the reference branch's error $Z_W(0)-T_W(0)$,
which uses the true reference future, leaves exactly the absolute effect
error, falling with the intervention from 0.0279 to 0.00305 and from 0.0051
to 0.000327. Most of $E_W$ is thus an offset shared by all branches. After
this correction, C-JEPA's heat error lies below the 0.053 of predicting the
last observed latent, and its spring and bounce errors of 0.141--0.240 stay
above the corresponding 0.042--0.068 (Table~\ref{tab:parityoffset}).

\begin{figure}[t]
\centering
\includegraphics[width=\linewidth]{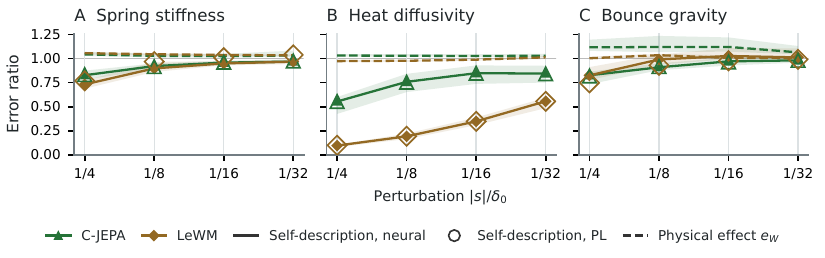}
\caption{Self-description and physical-effect errors in identical latent
coordinates. Solid lines give the held-out error of the neural law fitted to
each model's own responses at references $0,\pm\delta_0/2$, relative to its own
response norm. Open markers give PL, and dashed lines the physical effect
error $e_W$. Each point summarizes 144 queries. Bands are
conditional 95\% context-block intervals.}
\label{fig:samespace}
\end{figure}

\paragraph{A separation in identical coordinates.}
We fit self-response laws anew on the saved target-space observations,
holding out whole contexts and using references $0,\pm\delta_0/2$.
At the finest LeWM heat scale, the neural self-description has error ratio
0.556 $[0.484,0.594]$ relative to the model's own response, and the physical
effect error $e_W$ is 1.013 $[0.994,1.026]$, both in the same post-projection
coordinates. PL interpolation of the same three references
gives 0.555 $[0.481,0.600]$, so the self-description rests on simple local
response structure. LeWM's heat response is thus substantially more
describable than its physical effect is matched, and the gap widens at
coarser magnitudes (Figure~\ref{fig:samespace}). In spring and bounce,
neural self-description errors rise from 0.73--0.83 at $\delta_0/4$ to
0.97--1.01 at $\delta_0/32$ (Table~\ref{tab:samespace}).

\section{Transferring an RGB-Trained Inverse}
\label{sec:physical}

We freeze the reverse fitted and selected on future-RGB training
contexts and apply it to each subject's own three references and query,
without target-side regression (Appendix~\ref{app:simulatorreference}).
The target's PL inverse is already part of this readout, so the transferred
correction is measured by $S_{RGB}=1-\mathrm{SSE}_{RGB}/\mathrm{SSE}_{PL}$.
Four learned interfaces have positive median paired skills and five negative.
Only Cosmos null's interval lies above zero, at 0.102 $[0.018,0.262]$.
High raw scores at tokenizer or AC therefore do not establish a transferred
RGB increment (Table~\ref{tab:alignedreference}, Figure~\ref{appfig:rgbgain}).

\paragraph{A specific test at matched mechanical states.}
The separate viscoelastic exact-reset experiment retains branch-specific
material damping while matching positions and velocities before the future
to be predicted. In its preregistered 24-context extension, C-JEPA and AC
satisfy the registered visible-compression criteria with forward
$R^2=0.125/0.031$. Their material-future scores are
$-0.066/0.339$ against a 0.817
visual-history benchmark under the frozen ridge readout, with Holm-adjusted
inferiority $p=0.0004$ for both
(Appendix~\ref{app:material}). This contrast supports the conclusion that
geometric regularities alone are insufficient evidence of physical-law
emergence.

\section{Interpreting Local Response Error}
\label{sec:theory}

Fix a context and reference, let $Y(s)$ be the observation at log-offset
$s$, and let $\Delta_\pm=Y(\pm h)-Y(0)$,
$D_h=(\Delta_+-\Delta_-)/(2h)$ and
$K_h=(\Delta_++\Delta_-)/(2h)$. A linear law with slope $L$ predicts
$\pm hL$, and its paired error obeys the exact identity
\begin{equation}
\mathcal A_h(L)^2:=\tfrac12\big(\|hL-\Delta_+\|^2+\|{-hL}-\Delta_-\|^2\big)
=h^2(\|L-D_h\|^2+\|K_h\|^2).
\label{eq:absolutescale}
\end{equation}
For predicted effects $\widehat\Delta_\pm=\Delta Z_W(\pm h)$ and true effects
$\Delta_\pm=\Delta T_W(\pm h)$, define $\widehat D_h,\widehat K_h$ in the same
way. Their paired error is
\begin{equation}
\tfrac12\big(\|\widehat\Delta_+-\Delta_+\|^2+\|\widehat\Delta_--\Delta_-\|^2\big)
=h^2(\|\widehat D_h-D_h\|^2+\|\widehat K_h-K_h\|^2).
\label{eq:neuralpaired}
\end{equation}
We compute the even share of this error for each of the 72 $\pm h$ pairs per
magnitude, one per context and reference, before taking medians. At the
finest heat scale, the median even share is 0.730/0.741 for Cosmos null/text
and 0.164/0.053 for C-JEPA/LeWM (Table~\ref{tab:parityoffset}). Likewise,
the pointwise identity
$e^2=(A-1)^2+2A(1-c)$ separates amplitude and directional mismatch
(Table~\ref{tab:effectdiagnostics}).

Suppose $Y$ has derivative $J$ at the reference, so that
$\|Y(s)-Y(0)-sJ\|\le|s|\,\omega(|s|)$ for $|s|\le r$, with $\omega$
nondecreasing and $\omega(t)\to0$ as $t\to0$. Equivalently, $D_h\to J$ and
$K_h\to0$. For $2h\le r$,
\begin{equation}
\|D_h-J\|\le\omega(h),\qquad
\mathcal A_h(D_{2h})\le h[\omega(h)+\omega(2h)].
\label{eq:modulusrefinement}
\end{equation}
A slope re-estimated at $\pm2h$ thus predicts the responses at $\pm h$ with
$o(h)$ error. Step-normalized error $\mathcal A_h/h$, proofs, and
counterexamples appear in Appendix~\ref{app:scalerecovery}.

These splits locate each failure. At the finest heat scale the true effect is
mostly odd, with even share at most 0.158,
C-JEPA and LeWM mostly miss this odd part with small, weakly
aligned effects, and Cosmos null and text mostly add a frame change shared by
opposite interventions, which no slope represents.

\section{Discussion and Conclusion}

MultiEcho treats a frozen world model's complete recorded response to
counterfactual interventions as the experimental object. Our experiments
repeatedly establish testable facts about a learned world, each valid under
declared conditions of subject, intervention, contexts, observation time and
readout. Such facts are part of the science of a learned world.

These experiments also provide a concrete form of communication between
simulated and learned worlds: the simulator encodes controlled interventions
as input histories, each model responds through its own interface, and
simulated futures provide references for interpreting those responses.
A natural extension would connect physical and learned worlds, and enable
exchanges between learned worlds. The reference library already supports
messages and declared translations (Appendix~\ref{app:tools}). Counterfactual
experimentation can also extend to training through data augmentation and to
internal mechanisms through interventions on selected neuron activations
\citep{geiger2021,meng2022}.

We do not claim an exhaustive characterization of any subject, optimal
estimators or fits, or a ranking of models and architectures; nor do we
evaluate the strongest current systems. Broader scene coverage and joint
interventions on multiple parameters would extend the physical phenomena
and interactions under study. MultiEcho lays the groundwork for this
research direction through a common experimental object, a reproducible
protocol, and initial measurements and error identities that future work
can refine and extend.

\label{sec:mainend}

\clearpage
\section*{AI Use Statement}
AI assistants supported methodology refinement, mathematical arguments and
proof writing, implementation of synthetic experiments and analysis code,
source checking, result interpretation, scientific figures, and manuscript
drafting and editing. Model-generated clips are experimental observations
from the frozen subject configurations described in Appendix~\ref{app:design}.
Checks of AI-assisted work included comparisons with saved experimental
artifacts, numerical checks of identities, and executable validation of code
and figures. The authors are responsible for the final text, claims, and
artifacts.

\section*{Reproducibility Statement}
The methods and appendices specify the intervention families, model access,
estimators, and evaluation stages. Mathematical results include assumptions
and proofs. Figures are generated from the completed measurements, and the
reference library provides the experimental interfaces described in
Appendix~\ref{app:tools}.

\bibliographystyle{plainnat}
\bibliography{multiecho_counterfactual_responses_paper_final}

\clearpage
\appendix
\section{Experimental Design and Observation Access}
\label{app:design}

\subsection{Scenes and Sample Units}

The 24 intervention axes cover material parameters, initial conditions, and
geometry (Table~\ref{tab:axes}). Each has ten levels and 40 contexts per scene,
split into 24 discovery, eight validation and eight confirmation contexts.
This gives $24\times10\times40=9{,}600$ axis-level-context cases. Recording
nine learned interfaces and two simulator RGB observations yields 105,600
response records: 63,360 in discovery and 21,120 in each external stage.
Each stage retains the same 264 scene-axis-observation combinations.
Inputs can supply several interfaces, and axes can share a reference branch.

\paragraph{Collection-wide run count.}
The nine main campaigns contain 220,006 saved frozen-subject run records:
131,996 discovery, 43,970 validation and 44,040 confirmation records.
Each scene has 40 contexts split 24/8/8 across these stages, giving 360
scene-context settings. By scene, the run counts are 25,640 for spring,
28,362 for collision, 31,324 for bounce, 32,200 for refraction, 29,400 for
beam splitting, 20,720 for mechanics plus optics, 20,160 for heat, 12,040
for fluid, and 20,160 for viscoelastic recovery. These include reference,
intervention, appearance, event-window and control branches.
The repeatability and magnitude studies add 24,032 completed runs
(Appendix~\ref{app:scales}), the 12-context bounce visibility study adds
17,556 runs, 2,508 at each of seven subjects (Appendix~\ref{app:visibility}),
and the separate 24-context viscoelastic
confirmation extension adds 6,912 runs: 72 branches per context at four
subjects. The acquisition total is therefore
$220{,}006+24{,}032+17{,}556+6{,}912=268{,}506$ saved runs.
A run supplying both encoder and predictor observations is counted once.
Simulator RGB references, target-encoding and prediction replays, estimator
fits, duplicate downloads and superseded campaigns are excluded from this
acquisition count. Repeated runs and reused physical settings contribute
to data volume without constituting independent physical realizations.
The three-reference estimators are fitted only in discovery and evaluated
across all three stages. The temporal, visibility and material studies retain their original analyses
(Appendix~\ref{app:historical}).

\begin{table}[h]
\caption{Physical interventions in the atlas. Each axis is recorded through
eleven distinct observations. Coordinates marked ``log'' use logarithmic physical
parameters. The remaining coordinates use the recorded units.}
\label{tab:axes}
\centering
\small
\begin{tabular}{@{}lp{3.1in}r@{}}
\toprule
Scene & Intervention axes & Rows\\
\midrule
Two-mass spring & Stiffness, two masses, initial amplitude (all log) & 44\\
Collision & Two masses and speed (log), restitution & 44\\
Bounce & Gravity and initial height (log), restitution & 33\\
Refraction & Refractive index, slab thickness (log), incidence & 33\\
Beam splitting & Refractive index, plate angle, incidence & 33\\
Mechanics + optics & Gravity (log), initial height & 22\\
Heat & Diffusivity (log), initial temperature & 22\\
Fluid & Reynolds number (log) & 11\\
Viscoelastic recovery & Material $\beta$, compression & 22\\
\bottomrule
\end{tabular}
\end{table}

\begin{table}[h]
\caption{Observation and query counts used by the revised neural analysis.
Query counts exclude the two endpoint references and reference-parameter
duplicates. The same physical query identities are used at every interface.
External-stage counts do not multiply queries by the six frozen fits.}
\label{tab:accounting}
\centering
\small
\begin{tabular}{@{}llrrr@{}}
\toprule
Study & Observation set & Contexts & Units & Queries\\
\midrule
Discovery & Nine learned interfaces & 24/scene & 216 & 41,328\\
Discovery & Two simulator RGB frames & 24/scene & 48 & 9,184\\
Validation & Eleven interfaces & 8/scene & 264 & 16,841\\
Confirmation & Eleven interfaces & 8/scene & 264 & 16,830\\
Fine scale & Seven learned interfaces & 24/scene & 21 & 15,120\\
Fine scale & Two simulator RGB frames & 24/scene & 6 & 4,320\\
\midrule
Total & & & & 103,623\\
\bottomrule
\end{tabular}
\end{table}

Discovery has 4,592 query cases per interface after excluding
16 viscoelastic compression cases that equal the reference parameter. Validation
and confirmation have 1,531 and 1,530 queries per interface, respectively,
after excluding five and six such cases. Exclusions match at all eleven
interfaces. Each query is scored under five procedures: neural and PL forward,
neural and PL reverse, and frozen future-RGB reverse. The atlas therefore
contains 420,915 method-query scores, including 168,355 external-stage scores
averaged over six fixed fits. Fine-scale evaluation adds 97,200 method-query
scores. Repeated procedure or fold evaluations are not independent observations.

The current analysis uses context holdouts with local calibration. The lower
and upper references are the actual minimum and maximum transformed physical
coordinates.
The physical settings
underlying these ensembles are reused: nine heat baselines, three
mechanics-plus-optics clusters, three viscoelastic compression settings, and
one cached flow per Reynolds number. Context holdouts can therefore share
physical trajectories across folds.

\paragraph{Joint fitting across contexts.}
For each scene, physical axis, and interface, the forward and reverse
estimators share a training set spanning multiple contexts. The canonical
24-context evaluation has six outer folds: each fits 20 contexts and evaluates
the remaining four. All contexts provide three references. In the training
contexts, the remaining eligible levels provide supervised queries. The last
four training contexts in the recorded order are reserved for multiplier
selection. The other sixteen fit the selection model. After selection, the
same procedure refits on all twenty. Only response estimators are trained.
The studied world model remains frozen.

Contexts vary both visual and non-target physical conditions. In bounce,
the design varies camera azimuth, elevation, and distance, texture seed,
ball/floor colors, radius, and reference gravity/height. Each context is
evaluated at every intervention level. Thus canonical training already
includes visual diversity. It does not establish invariance to a separately
imposed appearance intervention. The nine-scene atlas reports independently
fitted scene-axis tools side by side. Fine-scale
evaluation freezes a declared atlas fit, then supplies three references in
each new context-anchor family.

\subsection{Adapters and Saved Interfaces}

VideoSAUR, LeWorldModel, V-JEPA 2, Cosmos, and Causal-JEPA supply the
encoders, predictors and generators of the frozen experimental subjects
\citep{zadaianchuk2023,maes2026,assran2025,nvidia2025,nam2026}.
Each receives the same 16-frame source stimulus. VideoSAUR, C-JEPA, and LeWM
adapt it to $224\times224$ images, while AC and Cosmos use $256\times256$.
Resizing is bilinear and antialiased with \texttt{align\_corners=False}.
Normalization follows the model pipeline. Table~\ref{tab:interfaces} gives
both the recorded observation and the history used during inference.

\begin{table}[h]
\caption{Model access and recorded observations. The inference column specifies
the history and conditioning used to produce each measurement.}
\label{tab:interfaces}
\centering
\small
\setlength{\tabcolsep}{3pt}
\begin{tabular}{@{}p{.84in}p{1.08in}p{1.27in}p{2.00in}@{}}
\toprule
Interface & Training domain & Saved observation & Input and inference policy\\
\midrule
VideoSAUR & CLEVRER object videos \citep{yi2020}, pretrained DINOv2 \citep{oquab2024} & Last-time seven 128-D slots & Fixed slot initialization seed 0.\\
C-JEPA predictor & CLEVRER slot sequences & First of eight predicted slot frames & 15 slot frames. Adapted 15-to-1 partition differs from original 6-to-10 training partition.\\
LeWM encoder & Released Reacher checkpoint \citep{tassa2018} & Final-frame projected 192-D CLS state & Same visual stimulus as its predictor.\\
LeWM predictor & Released Reacher checkpoint & First of eight predicted latents, before prediction projection & Three-latent history, constant zero actions.\\
AC encoder & Natural-video pretraining, robot transfer & Last tubelet, 256 tokens of 1,408 channels & Eight two-frame tubelets, zero action/state inputs.\\
AC predictor & Robot interactions, 62-hour post-training & First of four predicted tubelets, frames 16--17 & Same fixed conditions. Later rollout exceeds two-step automatic setting.\\
Cosmos tokenizer & Cosmos3-Nano Wan VAE \citep{wan2025}, corpus not audited & Last 48-channel latent slice, frames 12--15 & Repeat first source frame to give 17 input frames and five latent slices.\\
Cosmos null & Cosmos3-Nano 16B diffusion generator, corpus not audited & Frame 16 of generated frames 16--23 & Five conditioning latent slices, empty prompt, guidance 1, fixed seed, 35 steps.\\
Cosmos text & Same released generator & Frame 16 of generated frames 16--23 & Five conditioning latent slices, branch-invariant scene prompt, guidance 6, fixed seed, 35 steps.\\
Simulator last input & Not learned & Last input RGB frame & Frame 15 of the rendered branch history.\\
Simulator first future & Not learned & First future RGB frame & Frame 16 under identical physical conditions and sampling interval.\\
\bottomrule
\end{tabular}
\end{table}

VideoSAUR encoder, its predictor alias, and C-JEPA encoder share one reported
representation. Cosmos tokenizer encoder/predictor aliases likewise count once.
Their stored predictor outputs reproduce the final encoding and are not
separate future predictions.
Action paths remain active with constant inputs. Text prompts are branch-invariant.
Cosmos null/text differ in both prompt and guidance, precluding a text-only
attribution. Generated observations describe one fixed realization.

The generated RGB arrays are used directly, with stored byte values divided
by 255. Baseline and intervention use the same first generated frame, at the
first-future observation time. No additional encoder is applied to generated
images. The frozen VideoSAUR/C-JEPA decoder reconstructs features, and the
LeWM and AC configurations have no native pixel decoder.
Their original latent observations are therefore retained. Cosmos tokenizer
is explicitly a separate input-encoding subject.

The current atlas and fitted-scale results use the neural procedure described
below. Geometry is computed directly from observations. The separate
event-window, visibility and material-future studies retain their original frozen readouts
and stage-specific criteria (Appendix~\ref{app:historical}).

\subsection{Reusable Experimental Operations}
\label{app:tools}

The reference library implements the experiment through a
\texttt{WorldAdapter}: an application opens a subject, forks a context,
applies an intervention, records an observation, and closes the branch.
\texttt{PairRunner} executes the paired branches under a frozen revision.
Response fitters learn candidate functions, and evaluation routines compare
their predictions on independent queries with declared controls. This permits
compiled physical inputs and native action interfaces to use the same paired
construction while retaining their own observation spaces.

Applications supply scenes, checkpoints, observation maps, and estimators.
The core uses Python 3.11 or later and NumPy, with optional Torch integration.
It provides equal-group summaries, percentile bootstrap, and Bonferroni
intervals for new studies. The paper's results use the campaign estimators
specified below. Additional interfaces implement observation translation,
controlled message reception, sequential and path-based candidate laws, and
training-data export. These capabilities support extensions beyond the video
experiments reported here.

\subsection{Related Experimental Questions}
\label{app:related}

\paragraph{Experiments on trained systems.}
Machine behaviour proposes studying AI systems with the experimental methods
of the behavioral sciences \citep{rahwan2019}. Cognitive-psychology
experiments probe GPT-3 with vignettes and controlled tasks \citep{binz2023}.
Applied to a microprocessor, standard neuroscience analyses such as lesions
and tuning curves reveal structure in the recordings and miss the known
hierarchy of its information processing \citep{jonas2017}. Computer
experiments design inputs to deterministic simulators and fit statistical
predictors to their outputs \citep{sacks1989}. MultiEcho brings this
experimental stance to frozen world models, with designed physical inputs,
matched references and complete recorded responses.

\paragraph{World representations and their evaluation.}
World models learn predictive dynamics that support policy learning and
control \citep{ha2018,hafner2025}. Probes and interventions on a sequence
model trained on Othello moves reveal an emergent board-state representation
\citep{li2023}. Metrics derived from the Myhill--Nerode theorem show that
accurate next-token predictors can hold incoherent world models
\citep{vafa2024}, and inductive-bias probes show that models trained on
orbital trajectories fail to apply Newtonian mechanics to new physics tasks
\citep{vafa2025}. Violation-of-expectation tests find intuitive-physics
understanding in video models that predict in a learned representation space
\citep{garrido2025}. These studies compare a model with a declared world
model. MultiEcho measures the learned world's own intervention responses and
compares them with physical references as a separate step.

\paragraph{Interventions on internal states.}
Causal abstraction aligns network components with high-level causal models
through interchange interventions \citep{geiger2021}. Causal tracing restores
hidden activations to locate factual associations in GPT, and rank-one edits
change them \citep{meng2022}. MultiEcho's intervention handle admits such
internal states (Section~\ref{sec:method}). The reported experiments intervene
on rendered inputs, so every subject receives the same physical change.

\paragraph{Visual prompts and representation interventions.}
CWM perturbs visual inputs to extract flow, segmentation and depth. Its
Jacobian concerns the image input. Physical CWM uses frozen features for
contact tasks and reconstruction errors for plausibility
\citep{bear2023,venkatesh2024}. PEZ localizes and steers motion representations
through tuning fits, attention ablations and decoded-direction probes.
Its layerwise plausibility study also trains predictors \citep{joseph2026}.
PhyIP fits symbolic equations to energy/force readouts from numerical
world models \citep{interno2026}. A physically controllable world model
estimates any visual variable conditioned on others and extracts objects and
their physical relations \citep{venkatesh2026}. MultiEcho varies physical inputs and measures
complete responses before selecting components. Probe control tasks and
accuracy--complexity comparisons inform this design \citep{hewitt2019,pimentel2020}.

\paragraph{Learning alternative physical futures.}
CoPhy and Filtered-CoPhy train a map from factual history and edited initial
conditions to alternative futures. Filtered-CoPhy also constrains hidden-property
identifiability \citep{baradel2020,janny2022}. PhysEditWorld compares frozen
and adapted gravity-conditioned generation, including quantitative acceleration
ordering and qualitative first-person results \citep{hu2026}. CWMDT conditions
a video diffusion model on digital twins that a language model edits to
propagate an intervention \citep{shen2025}. MultiEcho fits
the intervention-response function of an already frozen subject.

\paragraph{Counterfactual consistency.}
CRONOS compares generation quality across variants of an event type
\citep{begiristain2026}. Viewpoint and appearance preserve physical parameters,
whereas scene and object changes can alter dynamics. Its sensitivity is a
range of quality scores. MultiEcho studies the response vector, whose large
magnitude can be an expected intervention effect.

\paragraph{Inferring properties from history.}
Physion++ supplies property-revealing histories before matched visible starts
and compares contact readouts with and without history \citep{tung2023}.
Exact reset applies this logic to matched positions and velocities with
different damping, using quantitative future trajectories.

\paragraph{Evaluating physical behavior.}
IntPhys introduced matched possible and impossible videos for evaluating
physical expectations \citep{riochet2018}. IntPhys 2 uses prediction errors
and VLM judgments. Its continuity concerns persistence \citep{bordes2025}. VideoPhy/PhyGenBench assess generated-video
plausibility, and VideoPhy-2 adds action-centered analysis
\citep{bansal2024,meng2025,bansal2025}. PhyWorldBench includes anti-physics
instructions \citep{gu2026}. PhysicsMind combines question answering with
generation evaluated against mechanical constraints \citep{mak2026}.
Physics-IQ compares recorded continuations and repeat variation
\citep{motamed2026}. WorldBench estimates physical parameters from trajectories.
Morpheus evaluates prescribed equations and invariants
\citep{upadhyay2026,tragoudaras2026}. These physical targets complement
descriptions of the learned world's own responses.

\paragraph{Video question answering.}
CausalVQA and MMWorld test multimodal answers to descriptive, counterfactual
and predictive video questions \citep{foss2025,he2024}. MultiEcho instantiates
paired interventions and measures changes in representations or generated
outputs, a different experimental endpoint.

\paragraph{Discovering dynamical equations.}
Symbolic regression searches expression spaces for free-form laws that fit
experimental data \citep{schmidt2009,udrescu2020}. Graph networks trained with
sparse messages can be distilled into symbolic force laws
\citep{cranmer2020}. SINDy selects sparse functions of state to describe time derivatives
\citep{brunton2016}. Bayesian SINDy autoencoders learn coordinates and dynamics,
including gravity from pendulum video \citep{gao2024}. AMORE discovers switching
modes and equations \citep{liu2024}. MultiEcho's numerical tools
describe finite intervention contrasts of a frozen subject, without inferring
a temporal differential equation.

\paragraph{System identification from video.}
\citet{kaneko2024} improves PAC-NeRF with Lagrangian particle optimization,
refining geometry and material properties. MASIV learns neural constitutive
models from particle trajectories \citep{zhao2025}. These identify the
video-producing physical system. MultiEcho identifies the learned subject's
response. Both symbolic and numerical descriptions can support either goal.

\paragraph{Continuous physical variation.}
PhyWorld studies parameter and combinatorial generalization. PISA examines
physical post-training and dropping-time distributions \citep{kang2025,li2025}.
MultiEcho shrinks interventions at fixed image resolution and observation times.
Causal identification supplies assumptions connecting observations to effects
\citep{pearl2009}. Sobolev training combines value and derivative constraints
\citep{czarnecki2017}.

\section{Estimators and Additional Atlas Results}
\label{app:estimators}

\subsection{A Common Neural Tool across Observation Dimensions}

Let $Y_{\rm PL}(u)$ linearly interpolate the two references bracketing $u$,
and let $t\in[0,1]$ denote its position within that interval. We predict
\begin{equation}
\widehat Y(u)=Y_{\rm PL}(u)+4t(1-t)R_\theta(\mathcal D_c,u),
\qquad \widehat{\Delta Y}=\widehat Y-Y_0.
\label{eq:neuralforward}
\end{equation}
A coordinate-shared network supplies the residual, with a zero-initialized
head and a gate that preserves all three reference observations. Reverse
starts from projected inversion of the reference polyline and learns a scalar
correction. Both use two 24-unit hidden layers, without subject-specific
image, slot, or vector architectures. Their fitted weights may differ.

For a context, let $a^2=(\|Y_--Y_0\|_{\rm RMS}^2+
\|Y_+-Y_0\|_{\rm RMS}^2)/2$. Normalize endpoint contrasts as
$\ell=(Y_--Y_0)/a$ and $r=(Y_+-Y_0)/a$, and the reference parameter as
$m=(u_0-u_-)/(u_+-u_-)$. When $a=0$, use one for numerical division and
retain the zero-amplitude flag. This convention uses only calibration data.

For coordinate $j$, the forward network has four local inputs:
$\operatorname{asinh}(\ell_j)$, $\operatorname{asinh}(r_j)$,
$\operatorname{asinh}(\ell_jr_j)$, and the inverse-hyperbolic-sine of the
coordinate's centered, RMS-normalized baseline value. Eight pooled inputs
are $m$, both endpoint RMS norms, their cosine, both coordinate means,
$\|r-\ell\|_{\rm RMS}$, and the fraction of coordinates with both contrasts
zero. Zero norms use finite zero conventions. These twelve values feed a
12--24--24--4 multilayer perceptron with SiLU hidden activations and a linear
head, shared across every coordinate.

The four outputs are $(A_j^-,B_j^-,A_j^+,B_j^+)$. On the bracketing interval
with side $b\in\{-,+\}$, the residual in
Equation~\ref{eq:neuralforward} is
\begin{equation}
R_{\theta,j}(\mathcal D_c,u)=\lambda a
\left[A_j^b+B_j^b(2t-1)\right].
\label{eq:residualbasis}
\end{equation}
Thus normalized query position enters through the gated polynomial basis.
There are 1,012 learned parameters regardless of output width. A constant
coordinate can receive a nonzero correction if other calibration coordinates
establish response amplitude. Fully flat calibration yields the
unchanged forward output, an explicit information limitation of this tool.
No output encoder, dimension reduction, or basis derived from query outputs
is used for neural training.

Reverse projects the complete normalized query contrast onto the two
calibration-polyline segments, with each segment fraction clipped to $[0,1]$.
The nearest projected point provides normalized estimate $z_{\rm PL}$.
Nineteen geometric inputs comprise this estimate, the two candidate positions,
the two unbounded segment fractions, query norm, two segment residual norms,
distances to all three references, and eight calibration descriptors. The
latter replace coordinate means and zero counts by endpoint energy fractions
and one minus squared cosine. Consequently reverse inputs are invariant to
orthogonal coordinate changes and isometric embeddings into wider outputs
after RMS normalization. Unbounded fractions and query-dependent norms use
the inverse-hyperbolic-sine transform.

The reverse network is 19--24--24--1, with 1,105 parameters, the same two
24-unit SiLU hidden layers, and a zero-initialized output head. If $d$ is
the normalized distance to the nearest reference, its correction is
$\lambda d/(1+d)$ times the network output. The final estimate is clipped
to the calibrated parameter range and converted back to recorded log/raw
units. The correction vanishes at an exactly observed reference. Equal
segment errors choose the lower segment. Fully coincident references use the reference
parameter as the polyline estimate. All cases remain in scoring.

\subsection{Training, Selection and Scores}

Both tasks use AdamW, learning rate 0.002, weight decay 0.001, gradient norm
clipping at 1, and 160 updates. Forward samples 1,024 uniformly chosen
eligible query-coordinate pairs per update. Reverse uses all eligible
training queries per update. Sampling does not reduce the evaluation target:
every native coordinate contributes to full-response error. The same
architecture, initialization policy, candidate set, and budget apply to every
subject. Forward and reverse are distinct tasks and have fixed task-specific
input/output dimensions. Seeds are $20260921+1009k$ for outer fold $k$.

An inner context holdout selects $\lambda\in\{0,0.25,0.5,1\}$ by forward
calibration-RMS-normalized mean squared error or reverse normalized-parameter
mean squared error. Ties prefer the smaller multiplier. We then refit from
the same initialization on all twenty outer-training contexts. Of 1,584 folds, forward
selects zero residual in 56 and reverse in 65. The test contexts never
select a multiplier, architecture, or training duration.

We report $R^2=1-\mathrm{SSE}/\mathrm{SST}$ within each physical axis and
observation. SSE sums squared errors over held-out queries. SST uses each
fold's mean eligible training-query target, with no test centering. Forward
targets are complete response vectors. Reverse targets are signed log/raw
parameter offsets. Sums of squares are pooled across folds before division.
Tables summarize medians of axis scores. Undefined zero-energy pointwise
relative errors retain their absolute error and explicit missing count.

The revised analysis is descriptive. It does not inherit earlier estimator
support decisions or permutation tests. Its new conditional block-bootstrap
intervals are defined below. Negative
scores and unsuccessful residual corrections are retained. An unchanged
output defines the relative-error baseline. PL forward and projected-polyline
reverse are equal-three-observation controls. No nearest-neighbor score is
part of this analysis.

\subsection{Subject-Resolved Atlas}

The frozen reverse
uses the same first-future RGB source for every observation, including
encoders, tokenizer and last-input RGB. Complete per-axis records retain
the scores underlying every median.

Table~\ref{tab:subjectscene} preserves scene and observation identities, including
both RGB references in all nine scenes. The companion results retain 264
axis-observation units, all weights and selected residual multipliers, pointwise
losses and negative scores. Main comparisons fix the common neural procedure
and weight the same 24 axes equally. No held-out best-family selection occurs.

\small
\begin{longtable}{@{}llrrrr@{}}
\caption{Scene-resolved neural response laws. Each entry is a median over that scene's physical axes. The last column applies the fixed future-RGB reverse to the subject's three-reference query geometry.}\label{tab:subjectscene}\\
\toprule
Scene & Observation & Axes & Forward $R^2$ & Reverse $R^2$ & RGB reverse $R^2$\\
\midrule
\endfirsthead
\toprule
Scene & Observation & Axes & Forward $R^2$ & Reverse $R^2$ & RGB reverse $R^2$\\
\midrule
\endhead
Bounce & VideoSAUR enc. & 3 & 0.310 & -0.128 & -1.035\\
Collision & VideoSAUR enc. & 4 & 0.396 & 0.413 & -0.046\\
Refraction & VideoSAUR enc. & 3 & 0.387 & 0.234 & -0.068\\
Mechanics + optics & VideoSAUR enc. & 2 & 0.304 & 0.056 & -1.129\\
Heat & VideoSAUR enc. & 2 & 0.219 & 0.048 & -0.606\\
Fluid & VideoSAUR enc. & 1 & 0.359 & 0.198 & -1.085\\
Viscoelastic & VideoSAUR enc. & 2 & 0.387 & 0.191 & -0.231\\
Splitting & VideoSAUR enc. & 3 & 0.325 & 0.175 & -0.273\\
Spring & VideoSAUR enc. & 4 & 0.303 & 0.147 & -0.266\\
Bounce & C-JEPA pred. & 3 & 0.379 & 0.457 & -0.346\\
Collision & C-JEPA pred. & 4 & 0.431 & 0.727 & 0.305\\
Refraction & C-JEPA pred. & 3 & 0.331 & 0.590 & 0.215\\
Mechanics + optics & C-JEPA pred. & 2 & 0.308 & -0.052 & -0.682\\
Heat & C-JEPA pred. & 2 & 0.338 & 0.300 & -0.213\\
Fluid & C-JEPA pred. & 1 & 0.364 & 0.277 & -0.381\\
Viscoelastic & C-JEPA pred. & 2 & 0.455 & 0.486 & 0.295\\
Splitting & C-JEPA pred. & 3 & 0.362 & 0.316 & -0.067\\
Spring & C-JEPA pred. & 4 & 0.384 & 0.510 & 0.079\\
Bounce & LeWM enc. & 3 & 0.404 & 0.042 & -0.780\\
Collision & LeWM enc. & 4 & 0.473 & 0.635 & 0.004\\
Refraction & LeWM enc. & 3 & 0.768 & 0.824 & 0.626\\
Mechanics + optics & LeWM enc. & 2 & 0.231 & 0.192 & -1.337\\
Heat & LeWM enc. & 2 & 0.975 & 0.995 & 0.989\\
Fluid & LeWM enc. & 1 & 0.466 & 0.723 & 0.184\\
Viscoelastic & LeWM enc. & 2 & 0.810 & -0.954 & -1.173\\
Splitting & LeWM enc. & 3 & 0.919 & 0.948 & 0.934\\
Spring & LeWM enc. & 4 & 0.568 & 0.652 & -0.078\\
Bounce & LeWM pred. & 3 & 0.873 & 0.951 & 0.822\\
Collision & LeWM pred. & 4 & 0.387 & 0.652 & 0.117\\
Refraction & LeWM pred. & 3 & 0.765 & 0.783 & 0.562\\
Mechanics + optics & LeWM pred. & 2 & 0.377 & 0.614 & -0.157\\
Heat & LeWM pred. & 2 & 0.967 & 0.994 & 0.982\\
Fluid & LeWM pred. & 1 & 0.399 & 0.551 & 0.150\\
Viscoelastic & LeWM pred. & 2 & 0.875 & 0.961 & 0.903\\
Splitting & LeWM pred. & 3 & 0.930 & 0.956 & 0.939\\
Spring & LeWM pred. & 4 & 0.639 & 0.690 & 0.485\\
Bounce & AC enc. & 3 & 0.374 & 0.890 & 0.443\\
Collision & AC enc. & 4 & 0.354 & 0.933 & 0.883\\
Refraction & AC enc. & 3 & 0.398 & 0.943 & 0.899\\
Mechanics + optics & AC enc. & 2 & 0.326 & 0.571 & 0.461\\
Heat & AC enc. & 2 & 0.331 & 0.524 & 0.368\\
Fluid & AC enc. & 1 & 0.342 & 0.870 & 0.620\\
Viscoelastic & AC enc. & 2 & 0.466 & 0.937 & 0.890\\
Splitting & AC enc. & 3 & 0.354 & 0.810 & 0.733\\
Spring & AC enc. & 4 & 0.374 & 0.971 & 0.840\\
Bounce & AC pred. & 3 & 0.360 & 0.869 & 0.409\\
Collision & AC pred. & 4 & 0.348 & 0.930 & 0.886\\
Refraction & AC pred. & 3 & 0.403 & 0.948 & 0.899\\
Mechanics + optics & AC pred. & 2 & 0.317 & 0.379 & 0.292\\
Heat & AC pred. & 2 & 0.312 & 0.160 & -0.100\\
Fluid & AC pred. & 1 & 0.334 & 0.762 & 0.548\\
Viscoelastic & AC pred. & 2 & 0.493 & 0.917 & 0.875\\
Splitting & AC pred. & 3 & 0.336 & 0.374 & 0.123\\
Spring & AC pred. & 4 & 0.371 & 0.968 & 0.830\\
Bounce & Cosmos tokenizer & 3 & 0.748 & 0.973 & 0.901\\
Collision & Cosmos tokenizer & 4 & 0.437 & 0.984 & 0.835\\
Refraction & Cosmos tokenizer & 3 & 0.597 & 0.997 & 0.908\\
Mechanics + optics & Cosmos tokenizer & 2 & 0.370 & 0.952 & 0.879\\
Heat & Cosmos tokenizer & 2 & 0.370 & 0.991 & 0.884\\
Fluid & Cosmos tokenizer & 1 & 0.470 & 0.978 & 0.965\\
Viscoelastic & Cosmos tokenizer & 2 & 0.758 & 0.995 & 0.961\\
Splitting & Cosmos tokenizer & 3 & 0.400 & 0.986 & 0.971\\
Spring & Cosmos tokenizer & 4 & 0.419 & 0.971 & 0.834\\
Bounce & Cosmos null & 3 & 0.320 & 0.120 & 0.112\\
Collision & Cosmos null & 4 & 0.437 & 0.681 & 0.639\\
Refraction & Cosmos null & 3 & 0.434 & 0.878 & 0.830\\
Mechanics + optics & Cosmos null & 2 & 0.342 & 0.860 & 0.768\\
Heat & Cosmos null & 2 & 0.911 & 0.999 & 0.999\\
Fluid & Cosmos null & 1 & 0.567 & 0.949 & 0.954\\
Viscoelastic & Cosmos null & 2 & 0.587 & 0.786 & 0.770\\
Splitting & Cosmos null & 3 & 0.468 & 0.963 & 0.963\\
Spring & Cosmos null & 4 & 0.399 & 0.542 & 0.426\\
Bounce & Cosmos text & 3 & 0.422 & 0.134 & -0.743\\
Collision & Cosmos text & 4 & 0.303 & 0.624 & 0.434\\
Refraction & Cosmos text & 3 & 0.444 & 0.793 & 0.779\\
Mechanics + optics & Cosmos text & 2 & 0.316 & 0.378 & 0.032\\
Heat & Cosmos text & 2 & 0.769 & 0.977 & 0.975\\
Fluid & Cosmos text & 1 & 0.511 & 0.954 & 0.929\\
Viscoelastic & Cosmos text & 2 & 0.418 & 0.127 & -0.242\\
Splitting & Cosmos text & 3 & 0.272 & 0.277 & -0.010\\
Spring & Cosmos text & 4 & 0.400 & 0.116 & -0.242\\
Bounce & RGB last input & 3 & 0.229 & -0.089 & -1.083\\
Collision & RGB last input & 4 & 0.413 & 0.741 & 0.717\\
Refraction & RGB last input & 3 & 0.436 & 0.981 & 0.981\\
Mechanics + optics & RGB last input & 2 & 0.340 & 0.459 & 0.395\\
Heat & RGB last input & 2 & 0.867 & 0.999 & 0.999\\
Fluid & RGB last input & 1 & 0.551 & 0.971 & 0.971\\
Viscoelastic & RGB last input & 2 & 0.596 & -0.700 & -0.757\\
Splitting & RGB last input & 3 & 0.469 & 0.991 & 0.991\\
Spring & RGB last input & 4 & 0.371 & 0.675 & 0.660\\
Bounce & RGB first future & 3 & 0.755 & 0.925 & 0.925\\
Collision & RGB first future & 4 & 0.423 & 0.918 & 0.918\\
Refraction & RGB first future & 3 & 0.438 & 0.981 & 0.981\\
Mechanics + optics & RGB first future & 2 & 0.374 & 0.948 & 0.948\\
Heat & RGB first future & 2 & 0.870 & 0.999 & 0.999\\
Fluid & RGB first future & 1 & 0.543 & 0.971 & 0.971\\
Viscoelastic & RGB first future & 2 & 0.716 & 0.994 & 0.994\\
Splitting & RGB first future & 3 & 0.469 & 0.991 & 0.991\\
Spring & RGB first future & 4 & 0.426 & 0.880 & 0.880\\
\bottomrule
\end{longtable}
\normalsize

\subsection{Frozen Fits across Experimental Stages}
\label{app:neuralstages}

For every axis and interface, we restore all six discovery-fold forward and
reverse networks, their selected residual multipliers, and the corresponding
future-RGB reverse networks. Each validation or confirmation context provides
its own baseline and actual parameter-range endpoints. All other eligible
levels are scored. No external-stage query selects a weight, multiplier,
training duration or fold, and no target-side regression is used for RGB
transfer. All 264 axis-interface combinations are evaluated in each stage.

For a fixed axis, interface and method, let $E_{kq}$ be the squared error
of discovery fit $k$ on external query $q$. Let $B_{kq}$ be the squared
error of that fit's original discovery training-query mean, in complete
response coordinates for forward or signed parameter coordinates for reverse.
We report
\begin{equation}
R^2_{\rm stage}=1-
\frac{\sum_q\frac16\sum_{k=1}^{6}E_{kq}}
{\sum_q\frac16\sum_{k=1}^{6}B_{kq}}.
\end{equation}
This averages losses across fixed fits before forming the pooled score.
Predictions are not averaged into an ensemble and no best fold is selected.
Each query contributes once to coverage counts. Table~\ref{tab:neuralstages}
then takes medians over the same 24 axis scores at each interface. Per-axis
records retain both PL controls, all six fold scores, all point losses and
negative outcomes. Future-RGB self-transfer equals its own reverse score
exactly on every axis in both stages.

The evaluation uses the existing eight-context validation and confirmation
collections. Low discovery scores motivated revisiting the initial response
estimators. Validation and confirmation were first evaluated before time was
available for estimator optimization; that later optimization used discovery
data only, not the external-stage outcomes. The revised discovery fits were
then applied unchanged to both external stages. Prior evaluation of these
cohorts did not make them inputs to estimator selection or optimization.
Earlier results remain separate and unchanged.

For each stage independently, the table also reports 1,999 hierarchical
scene/context-block draws: sample nine scenes with replacement, retain all
axes of each selected scene, and jointly resample that scene's complete
contexts across axes and interfaces. Discovery has 24 such contexts and
each external stage eight. The six discovery-fit losses have already been
averaged per external query. Intervals
are marginal and conditional on the saved fits. Reused
physical trajectories limit interpretation as independent dynamical evidence.

\begingroup\small
\setlength{\tabcolsep}{3pt}
\begin{longtable}{@{}lrrr@{}}
\caption{Response laws across stages. Entries are medians across all 24 axes with marginal 95\% scene/context-block percentile intervals from 1,999 draws. Discovery uses six-fold context cross-validation and 24 contexts per scene. Each external stage has eight contexts per scene and three labeled references per context. Its six unchanged discovery fits contribute averaged squared losses. Intervals condition on saved fits and the examined cohorts, without refitting, selection correction or multiplicity adjustment. Reused physical settings are not independent dynamical replicates.}\label{tab:neuralstages}\\
\toprule
Observation & Forward $R^2$ & Own reverse $R^2$ & RGB reverse $R^2$\\
\midrule
\endfirsthead
\toprule
Observation & Forward $R^2$ & Own reverse $R^2$ & RGB reverse $R^2$\\
\midrule
\endhead
\textbf{Discovery CV} & & &\\
RGB first future & 0.469 [0.432, 0.678] & 0.972 [0.953, 0.993] & 0.972 [0.953, 0.993]\\
RGB last input & 0.433 [0.370, 0.495] & 0.877 [0.458, 0.991] & 0.873 [0.002, 0.991]\\
VideoSAUR enc. & 0.344 [0.286, 0.385] & 0.208 [-0.015, 0.338] & -0.334 [-0.776, -0.130]\\
C-JEPA pred. & 0.363 [0.319, 0.409] & 0.455 [0.272, 0.681] & -0.061 [-0.273, 0.190]\\
LeWM enc. & 0.619 [0.420, 0.804] & 0.659 [0.229, 0.845] & 0.158 [-0.508, 0.767]\\
LeWM pred. & 0.742 [0.474, 0.889] & 0.760 [0.655, 0.955] & 0.645 [0.323, 0.913]\\
AC enc. & 0.359 [0.340, 0.385] & 0.898 [0.802, 0.945] & 0.780 [0.599, 0.878]\\
AC pred. & 0.350 [0.327, 0.380] & 0.885 [0.493, 0.938] & 0.721 [0.313, 0.873]\\
Cosmos tokenizer & 0.439 [0.396, 0.653] & 0.986 [0.978, 0.992] & 0.914 [0.873, 0.956]\\
Cosmos null & 0.431 [0.381, 0.516] & 0.776 [0.547, 0.947] & 0.768 [0.457, 0.950]\\
Cosmos text & 0.410 [0.308, 0.456] & 0.530 [0.187, 0.701] & 0.266 [-0.164, 0.695]\\
\midrule
\textbf{Validation} & & &\\
RGB first future & 0.468 [0.422, 0.676] & 0.973 [0.945, 0.994] & 0.973 [0.945, 0.994]\\
RGB last input & 0.446 [0.356, 0.544] & 0.783 [0.525, 0.988] & 0.800 [0.363, 0.988]\\
VideoSAUR enc. & 0.351 [0.287, 0.411] & 0.056 [-0.208, 0.275] & -0.610 [-0.935, -0.188]\\
C-JEPA pred. & 0.398 [0.301, 0.477] & 0.382 [0.265, 0.639] & -0.032 [-0.271, 0.187]\\
LeWM enc. & 0.541 [0.423, 0.899] & 0.586 [0.255, 0.893] & 0.170 [-0.492, 0.829]\\
LeWM pred. & 0.692 [0.483, 0.917] & 0.822 [0.640, 0.957] & 0.763 [0.334, 0.922]\\
AC enc. & 0.363 [0.338, 0.383] & 0.904 [0.768, 0.956] & 0.791 [0.521, 0.865]\\
AC pred. & 0.359 [0.323, 0.383] & 0.860 [0.566, 0.938] & 0.693 [0.393, 0.855]\\
Cosmos tokenizer & 0.447 [0.392, 0.657] & 0.986 [0.978, 0.993] & 0.912 [0.885, 0.957]\\
Cosmos null & 0.432 [0.376, 0.550] & 0.852 [0.588, 0.958] & 0.823 [0.507, 0.961]\\
Cosmos text & 0.383 [0.307, 0.507] & 0.431 [0.015, 0.779] & 0.207 [-0.312, 0.675]\\
\midrule
\textbf{Confirmation} & & &\\
RGB first future & 0.477 [0.439, 0.688] & 0.977 [0.960, 0.993] & 0.977 [0.960, 0.993]\\
RGB last input & 0.453 [0.371, 0.568] & 0.899 [0.463, 0.987] & 0.877 [0.204, 0.987]\\
VideoSAUR enc. & 0.332 [0.274, 0.413] & 0.050 [-0.168, 0.337] & -0.665 [-1.161, -0.042]\\
C-JEPA pred. & 0.412 [0.268, 0.474] & 0.401 [0.145, 0.710] & -0.107 [-0.540, 0.196]\\
LeWM enc. & 0.543 [0.443, 0.799] & 0.703 [0.278, 0.861] & 0.255 [-0.137, 0.739]\\
LeWM pred. & 0.622 [0.490, 0.910] & 0.829 [0.691, 0.957] & 0.697 [0.312, 0.896]\\
AC enc. & 0.359 [0.339, 0.381] & 0.877 [0.806, 0.950] & 0.779 [0.586, 0.866]\\
AC pred. & 0.357 [0.331, 0.379] & 0.868 [0.498, 0.937] & 0.750 [0.430, 0.855]\\
Cosmos tokenizer & 0.441 [0.391, 0.655] & 0.984 [0.979, 0.992] & 0.919 [0.892, 0.959]\\
Cosmos null & 0.440 [0.381, 0.548] & 0.836 [0.583, 0.952] & 0.798 [0.462, 0.959]\\
Cosmos text & 0.418 [0.278, 0.547] & 0.391 [0.195, 0.857] & 0.242 [-0.150, 0.757]\\
\bottomrule
\end{longtable}
\endgroup

\subsection{Frozen Fits under a Moved Camera}
\label{app:viewpoint}

Every discovery context has a second rendering with unchanged physics and
a moved camera. The camera turns by $45^\circ$ in azimuth and $-15^\circ$
in elevation in spring, collision, bounce and refraction, and by $45^\circ$
and $-8^\circ$ in beam splitting. It turns by $10^\circ$ in azimuth and
$12^\circ$ in elevation in mechanics plus optics and viscoelastic recovery.
In fluid it rises by $7^\circ$ and moves 0.18 m farther away. Heat changes
the bench and back textures, supports, lighting and camera together.
The reference and endpoint observations of each context at the moved view
supply its three references. Each context is scored by the discovery fold
that held it out, with that fold's canonical training-query mean as the
$R^2$ reference, as in Table~\ref{tab:scenes}. Re-scoring the canonical
view through the same code reproduces every saved discovery point loss.
All weights and multipliers are the saved discovery values, and all
observations are saved extractions. Simulator RGB is omitted because its
moved-view first future was never rendered.

\begin{table}[h]
\caption{Frozen discovery fits under the recorded camera changes. Each discovery context is rendered again with unchanged physics and scored by the fold that held it out, using the moved view's own three references. Entries are medians over the same 24 axes. Lower gives the percentage of axes whose viewpoint score falls below the canonical score, compared before rounding.}
\label{tab:viewpoint}
\centering\small
\setlength{\tabcolsep}{3pt}
\begin{tabular}{@{}lrrrrrr@{}}
\toprule
& \multicolumn{3}{c}{Forward $R^2$} & \multicolumn{3}{c}{Reverse $R^2$}\\
\cmidrule(lr){2-4}\cmidrule(l){5-7}
Observation & Canonical & Viewpoint & Lower & Canonical & Viewpoint & Lower\\
\midrule
VideoSAUR enc. & 0.344 & 0.365 & 41.7 & 0.208 & 0.176 & 58.3\\
C-JEPA pred. & 0.363 & 0.374 & 37.5 & 0.455 & 0.449 & 50.0\\
LeWM enc. & 0.619 & 0.560 & 41.7 & 0.659 & 0.456 & 58.3\\
LeWM pred. & 0.742 & 0.687 & 45.8 & 0.760 & 0.852 & 58.3\\
AC enc. & 0.359 & 0.355 & 33.3 & 0.898 & 0.893 & 62.5\\
AC pred. & 0.350 & 0.346 & 37.5 & 0.885 & 0.840 & 58.3\\
Cosmos tokenizer & 0.439 & 0.516 & 16.7 & 0.986 & 0.979 & 58.3\\
Cosmos null & 0.431 & 0.450 & 45.8 & 0.776 & 0.809 & 54.2\\
Cosmos text & 0.410 & 0.430 & 37.5 & 0.530 & 0.417 & 50.0\\
\bottomrule
\end{tabular}
\end{table}

Cosmos null and text share every input, query and reference rule, so their
neural losses pair exactly. Table~\ref{tab:promptpairing} forms axis-wise
null minus text differences in $R^2$ and resamples both configurations with
the same hierarchical scene/context-block draws in each condition.

\begin{table}[h]
\caption{Paired Cosmos null minus text differences in neural $R^2$. Each estimate is the median over 24 axes of axis-wise differences, with a 95\% scene/context-block percentile interval from 1,999 draws shared by both configurations. $>0$ counts axes whose difference is positive, and $\uparrow$/$\downarrow$ count axes whose own paired context-block interval lies wholly above/below zero. The configurations differ in one branch-invariant scene prompt and guidance, and each follows one fixed generation seed. Intervals condition on saved fits and the examined cohorts.}
\label{tab:promptpairing}
\centering\small
\setlength{\tabcolsep}{3pt}
\begin{tabular}{@{}lrrrrrrrr@{}}
\toprule
& \multicolumn{4}{c}{Forward $R^2$} & \multicolumn{4}{c}{Reverse $R^2$}\\
\cmidrule(lr){2-5}\cmidrule(l){6-9}
Condition & Null$-$text & $>0$ & $\uparrow$ & $\downarrow$ & Null$-$text & $>0$ & $\uparrow$ & $\downarrow$\\
\midrule
Discovery CV & 0.062 [-0.001, 0.146] & 16 & 15 & 0 & 0.219 [0.048, 0.483] & 20 & 14 & 0\\
Validation & 0.056 [-0.018, 0.142] & 15 & 13 & 2 & 0.396 [0.071, 0.584] & 22 & 15 & 1\\
Confirmation & 0.064 [-0.007, 0.205] & 17 & 11 & 0 & 0.232 [0.018, 0.477] & 21 & 15 & 1\\
Moved camera & 0.026 [-0.016, 0.101] & 15 & 15 & 0 & 0.294 [0.111, 0.395] & 23 & 12 & 0\\
\bottomrule
\end{tabular}
\end{table}

\subsection{Equal-Calibration Controls}
\label{app:neuralcontrols}

The PL forward uses exactly the same three measurements as the neural
predictor. Its inverse uses the same observed query and the same polyline
projection as the neural reverse, but omits the residual. Table~\ref{tab:neuralcontrols} retains all interfaces and
the number of axes improved. A higher median does not imply improvement
on every axis, and test outcomes never select the primary method.

\begin{table}[h]
\caption{Equal-three-reference comparisons. PL denotes piecewise-linear response prediction or its projected inverse. N denotes the neural residual tool. Scores are axis medians. Gains count axes with a higher neural score.}
\label{tab:neuralcontrols}
\centering\small
\setlength{\tabcolsep}{3pt}
\begin{tabular}{@{}lrrrrrr@{}}
\toprule
Observation & F: PL & F: N & R: PL & R: N & F gains & R gains\\
\midrule
VideoSAUR enc. & 0.311 & 0.344 & -0.280 & 0.208 & 20/24 & 24/24\\
C-JEPA pred. & 0.348 & 0.363 & 0.148 & 0.455 & 19/24 & 24/24\\
LeWM enc. & 0.564 & 0.619 & 0.451 & 0.659 & 20/24 & 22/24\\
LeWM pred. & 0.705 & 0.742 & 0.716 & 0.760 & 22/24 & 22/24\\
AC enc. & 0.317 & 0.359 & 0.876 & 0.898 & 24/24 & 22/24\\
AC pred. & 0.311 & 0.350 & 0.859 & 0.885 & 24/24 & 23/24\\
Cosmos tokenizer & 0.401 & 0.439 & 0.920 & 0.986 & 24/24 & 24/24\\
Cosmos null & 0.411 & 0.431 & 0.694 & 0.776 & 18/24 & 21/24\\
Cosmos text & 0.372 & 0.410 & 0.340 & 0.530 & 20/24 & 22/24\\
RGB last input & 0.378 & 0.433 & 0.741 & 0.877 & 24/24 & 23/24\\
RGB first future & 0.447 & 0.469 & 0.913 & 0.972 & 23/24 & 24/24\\
\bottomrule
\end{tabular}
\end{table}

\begin{table}[h]
\caption{Paired gains conditional on the recorded fits. First three columns are skill relative to the matching PL control. The last is own reverse skill relative to RGB reverse. Each estimate is a median of axis-wise paired loss ratios, with 1,999 hierarchical scene/context-block draws and 95\% percentile intervals. These intervals do not account for tool selection on discovery data or repeated physical trajectories. Axis-wise context-block intervals accompany the analysis records.}
\label{tab:pairedintervals}
\centering\scriptsize
\setlength{\tabcolsep}{3pt}
\begin{tabular}{@{}lrrrr@{}}
\toprule
Observation & $S_F$ & $S_R$ & $S_{RGB}$ & Own vs. RGB\\
\midrule
VideoSAUR enc. & 0.056 [0.024, 0.074] & 0.366 [0.297, 0.422] & -0.075 [-0.134, 0.016] & 0.377 [0.342, 0.459]\\
C-JEPA pred. & 0.032 [0.002, 0.050] & 0.361 [0.245, 0.474] & -0.035 [-0.141, 0.009] & 0.469 [0.358, 0.506]\\
LeWM enc. & 0.059 [0.024, 0.265] & 0.369 [0.174, 0.522] & -0.019 [-0.187, 0.142] & 0.473 [0.353, 0.602]\\
LeWM pred. & 0.059 [0.023, 0.301] & 0.290 [0.166, 0.573] & -0.082 [-0.524, 0.151] & 0.477 [0.343, 0.606]\\
AC enc. & 0.058 [0.047, 0.135] & 0.345 [0.223, 0.433] & -0.056 [-0.551, 0.066] & 0.420 [0.290, 0.561]\\
AC pred. & 0.054 [0.040, 0.114] & 0.312 [0.138, 0.423] & 0.008 [-0.510, 0.121] & 0.363 [0.253, 0.487]\\
Cosmos tokenizer & 0.083 [0.055, 0.106] & 0.757 [0.573, 0.913] & 0.033 [-0.108, 0.246] & 0.639 [0.450, 0.916]\\
Cosmos null & 0.048 [0.006, 0.083] & 0.247 [0.098, 0.331] & 0.102 [0.018, 0.262] & 0.071 [-0.031, 0.171]\\
Cosmos text & 0.043 [0.014, 0.077] & 0.279 [0.138, 0.336] & 0.036 [-0.019, 0.123] & 0.232 [0.115, 0.311]\\
RGB last input & 0.100 [0.021, 0.135] & 0.526 [0.447, 0.719] & 0.447 [0.206, 0.652] & 0.056 [-0.001, 0.214]\\
RGB first future & 0.103 [0.052, 0.122] & 0.623 [0.528, 0.753] & 0.623 [0.528, 0.753] & 0.000 [0.000, 0.000]\\
\bottomrule
\end{tabular}
\end{table}

Paired skills $S=1-\mathrm{SSE}_N/\mathrm{SSE}_{PL}$ and $\Delta R^2$ are computed per axis from the same query
losses before taking a cross-axis median. A difference between two marginal
medians is not this paired contrast. Within an axis, each bootstrap draw
samples 24 complete contexts, retaining all levels. For cross-axis summaries,
each draw samples nine scene blocks, keeps all axes in each sampled scene,
and resamples that scene's contexts jointly across axes and interfaces.
The displayed statistic weights the original 24 axes equally. We use 1,999
draws, seed 20260923, and 2.5th/97.5th percentiles. Physical-scale intervals
resample complete contexts, retaining their anchors and signs. These are
conditional descriptive intervals: no tool is refitted, the fitting procedure
was developed after viewing discovery data, and reused trajectories limit
independent dynamical inference. The intervals are marginal and do not
establish optimal fitting or physical accuracy. Undefined ratios retain
their missing counts.

The fitted weights and validation losses are saved for every fold. A test
that replaces outer-query outputs with extreme values leaves that fold's
weights and selected multipliers unchanged. Forward tests check reference
preservation and the capacity to leave the calibration span. Reverse tests
check dimension-independent inputs, geometric invariance, and exact reuse
of frozen source weights. These tests establish implementation properties.

\subsection{Geometry, Time, and Reuse}
\label{app:geometrydefinitions}

For responses $x_{ci}$ and signed log/raw offsets $s_{ci}$,
$\rho_c=\operatorname{corr}_{i<j}(|s_{ci}-s_{cj}|,\|x_{ci}-x_{cj}\|)$.
The global statistic uses level-mean offsets and responses, with recorded
zero-offset exclusions. Its null permutes level identities. For standardized
offsets $w_{ci}$, let $v_c=\sum_iw_{ci}x_{ci}$ and
$d_c=v_c/\|v_c\|$ when nonzero. Direction consistency is
$\|\sum_cd_c\|/n_{\rm valid}$. Its null flips context signs. Both tests
use 499 draws and threshold $0.05/n_{\rm axes}$. Within-context correlations
are descriptive. Effective rank summarizes direction spread, while signed-side
cosine compares mean effects below/above the reference, with potentially
unequal offsets. None is a continuum derivative.

Reference subtraction cancels additive backgrounds but can retain shadows,
occlusions, and other intervention effects. Neither averaging nor projection
assigns physical meaning: context-dependent physical directions can cancel
along with visually specific ones. The new fits retain these coordinates.

\subsection{Context-Resolved Distances and Local Geometry}
\label{app:contextgeometry}

An analysis resolves the global geometry into individual
contexts and parameter positions for all 216 atlas units. Each scene, axis,
and interface is analyzed separately, using the complete recorded response
at ten levels in 24 discovery contexts. These observations produce 24
distance matrices, 216 adjacent intervals, 192 interior positions, and
276 context pairs per unit. No new model inference, projection, or physical
rendering is used. The analysis concerns the original atlas grid, separately
from the six-magnitude experiment in Section~\ref{sec:scales}.

\paragraph{Individual distances and the mean response.}
Write $x_{ci}=\Delta Y_c(u_{ci})$ and $\bar x_i=24^{-1}\sum_c x_{ci}$.
For each context we retain all 45 pairwise RMS distances and their
association with transformed physical-parameter separation. Distances
between mean responses are compared with the distribution of individual
distances through
\begin{equation}
d_{c,ij}=\|x_{ci}-x_{cj}\|_{\mathrm{RMS}},\quad
\bar d_{ij}=\|\bar x_i-\bar x_j\|_{\mathrm{RMS}},\quad
C=\frac{\sum_{i<j}\bar d_{ij}^{\,2}}
{24^{-1}\sum_c\sum_{i<j}d_{c,ij}^{\,2}}.
\label{eq:contextdistances}
\end{equation}
Here $C\in[0,1]$ measures the fraction of mean squared within-context
separation retained after averaging response vectors. It can decrease
through differences in direction or amplitude. It does not identify the
discarded variation as noise. All ten measured levels are retained,
including 16 zero-offset viscoelastic compression cases excluded from the
older global statistic's level means. That original statistic is reproduced
separately and remains unchanged.

\paragraph{Shape and amplitude across contexts.}
For the 45-entry distance vector $d_c$, set
$A_c=\operatorname{RMS}(d_c)$ and $q_c=d_c/A_c$.
The discrepancy $S(c,c')=\operatorname{RMS}(q_c-q_{c'})$ compares shape
independently of overall amplitude, translation, or orthogonal rotation.
Zero discrepancy denotes proportional distance matrices. We report
discrepancies from the mean-response matrix and between all context pairs,
with amplitude differences summarized separately by
$B=\exp[\operatorname{median}_{c<c'}|\log(A_c/A_{c'})|]$.
Relative parameter-grid spacings match across contexts within each axis.
The distance contrast, $\operatorname{sd}(d_c)/\operatorname{mean}(d_c)$,
helps distinguish structured distance variation from nearly equidistant
geometry. Similarity of distance matrices alone provides neither physical
semantics nor evidence of a shared mechanism.

\paragraph{Local direction and sensitivity.}
Using each context's actual increasing log/raw coordinates $s_{ci}$,
the interval secant is
$v_{ci}=(x_{c,i+1}-x_{ci})/(s_{c,i+1}-s_{ci})$.
We retain its RMS norm on every adjacent interval. At each interior level,
let $v_-,v_+$ be the left and right increasing-parameter secants. Their
angle measures turning, while
$a=(\|v_+\|-\|v_-\|)/(\|v_+\|+\|v_-\|)$ measures signed amplitude
asymmetry and $b=\|v_+-v_-\|/(\|v_+\|+\|v_-\|)$ also captures direction
changes. Because these finite intervals can have unequal lengths, their
normalized secants are not a same-step symmetry test. Large turns do not establish
infinitesimal discontinuity, particularly when response norms are small.
Zero-norm ratios and angles remain undefined, with their counts retained.

\paragraph{Averaging can change the apparent organization.}
For C-JEPA thermal diffusivity, distance association is 0.861 for the mean
response but has median 0.186 within contexts, with 5th--95th percentiles
$[-0.141,0.592]$. Mean squared-distance retention is 0.082. The corresponding
LeWM encoder/predictor within-context medians are 0.995/0.996, with retention
0.800/0.791. Their median context-pair shape discrepancies are 0.035/0.039,
despite amplitude factors of 1.33/1.42. Thus a geometrically organized mean
can coexist with heterogeneous individual responses, while other interfaces
retain that organization within contexts (Table~\ref{tab:contextgeometry}).

\paragraph{Similar distance shapes can have different local behavior.}
Heat AC encoder/predictor context-pair discrepancies are 0.022/0.060, but
distance contrasts are only 0.025/0.044. Their median local turns are
$119.2^\circ/119.1^\circ$, despite small median absolute amplitude
asymmetries of 0.007/0.014. Nearly equal interval norms and similar distance
matrices therefore coexist with large changes in vector direction. Nearly
equidistant triples themselves produce turns near $120^\circ$ under this
definition, so the pattern does not establish a physical discontinuity.

LeWM's diffusivity turns have medians $16.3^\circ/15.8^\circ$ and
95th percentiles $30.5^\circ/30.6^\circ$. Encoder median secant RMS rises
from 0.0240 on levels 1:2 to 0.0276 on levels 6:7, then falls to 0.0227 on
levels 9:10. Predictor values are 0.0258, 0.0324, and 0.0255 in its own
coordinate units. The largest level-wise median turns occur at level 8
for the encoder ($21.2^\circ$) and level 9 for the predictor ($21.7^\circ$).
These profiles locate sensitivity changes along the sampled parameter
range without treating a single global direction as the whole response.

The companion records retain every context's matrix, every interval norm,
and all signed and vector asymmetries. Missing observations remain visible:
the viscoelastic LeWM encoder's $\beta$ response has only 48/192 defined
turns, so its $180^\circ$ median conditions on those positions. All 216
units retain the same physical query identities. The 48 generated-image
units are recomputed directly from RGB, while the other 168 retain their
original observations and statistics. Norms and turning cosines are checked against
the saved distance matrices. These analyses preserve the atlas's reused
physical settings and fixed-realization interpretation.

\begingroup
\small
\setlength{\tabcolsep}{3pt}
\begin{longtable}{@{}llrrrrrrr@{}}
\caption{Context-resolved response geometry for every atlas axis and interface. $\rho_{\rm m}/\widetilde\rho_c$ compares mean-response and median within-context distance association. $C$ is mean squared-distance retention. CV is median within-context distance standard deviation divided by mean, and $S_{\rm p}$ is median normalized distance discrepancy between contexts. $B$ is the median-pair amplitude factor, $\theta$ the median local turning angle in degrees, and $|a|$ the median absolute neighboring-secant amplitude asymmetry. Medians use defined observations. Brackets give their count when incomplete (24 contexts for $\rho$ and CV, 192 positions otherwise). All ten levels are used.} \label{tab:contextgeometry}\\
\toprule
Interface & Axis & $\rho_{\rm m}/\widetilde\rho_c$ & $C$ & CV & $S_{\rm p}$ & $B$ & $\theta$ & $|a|$\\
\midrule
\endfirsthead
\multicolumn{9}{l}{Table \thetable\ continued}\\
\toprule
Interface & Axis & $\rho_{\rm m}/\widetilde\rho_c$ & $C$ & CV & $S_{\rm p}$ & $B$ & $\theta$ & $|a|$\\
\midrule
\endhead
\midrule
\multicolumn{9}{r}{Continued on next page}\\
\endfoot
\bottomrule
\endlastfoot
\multicolumn{9}{l}{\textbf{Spring}}\\*
VideoSAUR enc. & k & 0.474/-0.047 & 0.057 & 0.24 & 0.369 & 1.32 & 117.7 & 0.164\\*
 & $m_A$ & 0.871/0.507 & 0.095 & 0.21 & 0.241 & 1.14 & 111.4 & 0.169\\*
 & $m_B$ & 0.896/0.503 & 0.090 & 0.21 & 0.259 & 1.16 & 111.7 & 0.182\\*
 & amp. & 0.701/0.333 & 0.062 & 0.33 & 0.404 & 1.40 & 106.8 & 0.226\\
C-JEPA pred. & k & 0.405/0.016 & 0.068 & 0.37 & 0.504 & 1.29 & 117.7 & 0.199\\*
 & $m_A$ & 0.852/0.680 & 0.305 & 0.36 & 0.287 & 1.21 & 101.2 & 0.212\\*
 & $m_B$ & 0.852/0.587 & 0.303 & 0.38 & 0.312 & 1.19 & 106.0 & 0.184\\*
 & amp. & 0.576/0.434 & 0.056 & 0.37 & 0.420 & 1.63 & 97.7 & 0.232\\
LeWM enc. & k & 0.641/0.425 & 0.042 & 0.53 & 0.493 & 1.37 & 116.1 [177] & 0.239\\*
 & $m_A$ & 0.809/0.644 & 0.227 & 0.41 & 0.374 & 1.57 & 106.2 & 0.170\\*
 & $m_B$ & 0.784/0.641 & 0.215 & 0.41 & 0.368 & 1.61 & 101.8 & 0.166\\*
 & amp. & 0.949/0.712 & 0.062 & 0.43 & 0.368 & 1.44 & 60.6 & 0.180\\
LeWM pred. & k & 0.823/0.731 & 0.135 & 0.57 & 0.252 & 1.64 & 62.8 & 0.141\\*
 & $m_A$ & 0.837/0.663 & 0.183 & 0.40 & 0.366 & 1.56 & 103.0 & 0.167\\*
 & $m_B$ & 0.794/0.645 & 0.174 & 0.41 & 0.356 & 1.56 & 103.0 & 0.174\\*
 & amp. & 0.970/0.720 & 0.082 & 0.43 & 0.359 & 1.47 & 57.5 & 0.157\\
AC enc. & k & 0.989/0.918 & 0.116 & 0.15 & 0.074 & 1.03 & 111.7 & 0.107\\*
 & $m_A$ & 0.962/0.925 & 0.166 & 0.12 & 0.046 & 1.02 & 113.0 & 0.110\\*
 & $m_B$ & 0.960/0.932 & 0.165 & 0.13 & 0.043 & 1.02 & 112.6 & 0.106\\*
 & amp. & 0.949/0.870 & 0.091 & 0.16 & 0.071 & 1.02 & 107.7 & 0.107\\
AC pred. & k & 0.988/0.911 & 0.099 & 0.14 & 0.079 & 1.04 & 111.8 & 0.104\\*
 & $m_A$ & 0.961/0.930 & 0.181 & 0.13 & 0.048 & 1.03 & 112.5 & 0.113\\*
 & $m_B$ & 0.959/0.930 & 0.182 & 0.13 & 0.047 & 1.03 & 112.5 & 0.104\\*
 & amp. & 0.941/0.859 & 0.091 & 0.16 & 0.072 & 1.03 & 107.7 & 0.110\\
Cosmos tokenizer & k & 0.890/0.761 & 0.433 & 0.46 & 0.083 & 1.17 & 66.8 & 0.117\\*
 & $m_A$ & 0.973/0.884 & 0.406 & 0.17 & 0.088 & 1.07 & 118.4 & 0.140\\*
 & $m_B$ & 0.970/0.874 & 0.393 & 0.16 & 0.088 & 1.08 & 118.7 & 0.135\\*
 & amp. & 0.932/0.813 & 0.186 & 0.33 & 0.037 & 1.06 & 62.5 & 0.092\\
Cosmos null & k & 0.596/0.475 & 0.072 & 0.33 & 0.363 & 1.20 & 107.1 & 0.180\\*
 & $m_A$ & 0.816/0.457 & 0.167 & 0.11 & 0.128 & 1.16 & 113.4 & 0.099\\*
 & $m_B$ & 0.862/0.436 & 0.142 & 0.12 & 0.151 & 1.19 & 113.4 & 0.114\\*
 & amp. & 0.913/0.745 & 0.084 & 0.21 & 0.136 & 1.17 & 95.5 & 0.122\\
Cosmos text & k & 0.520/0.427 & 0.058 & 0.47 & 0.535 & 1.78 & 110.1 & 0.183\\*
 & $m_A$ & 0.543/0.343 & 0.098 & 0.34 & 0.420 & 1.44 & 113.4 & 0.149\\*
 & $m_B$ & 0.686/0.410 & 0.117 & 0.33 & 0.408 & 1.51 & 113.5 & 0.138\\*
 & amp. & 0.588/0.638 & 0.051 & 0.23 & 0.348 & 1.38 & 99.1 & 0.145\\
\midrule
\multicolumn{9}{l}{\textbf{Collision}}\\*
VideoSAUR enc. & $m_A$ & 0.531/0.463 & 0.086 & 0.31 & 0.330 & 1.18 & 110.5 & 0.205\\*
 & $m_B$ & 0.706/0.425 & 0.052 & 0.31 & 0.337 & 1.20 & 109.9 & 0.180\\*
 & e & 0.908/0.515 & 0.059 & 0.32 & 0.388 & 1.22 & 110.7 & 0.243\\*
 & v & 0.868/0.540 & 0.077 & 0.34 & 0.312 & 1.17 & 105.9 & 0.274\\
C-JEPA pred. & $m_A$ & 0.585/0.508 & 0.159 & 0.52 & 0.407 & 1.19 & 98.8 & 0.267\\*
 & $m_B$ & 0.654/0.476 & 0.115 & 0.45 & 0.384 & 1.32 & 98.9 & 0.250\\*
 & e & 0.925/0.737 & 0.050 & 0.42 & 0.376 & 1.34 & 103.9 & 0.234\\*
 & v & 0.743/0.550 & 0.156 & 0.49 & 0.372 & 1.29 & 91.3 & 0.288\\
LeWM enc. & $m_A$ & 0.435/0.247 & 0.189 & 0.47 & 0.441 & 1.61 & 91.8 & 0.189\\*
 & $m_B$ & 0.600/0.337 & 0.157 & 0.44 & 0.432 & 1.63 & 93.6 & 0.201\\*
 & e & 0.791/0.539 & 0.165 & 0.51 & 0.435 & 1.58 & 78.8 & 0.229\\*
 & v & 0.829/0.353 & 0.071 & 0.63 & 0.662 & 1.79 & 131.9 & 0.373\\
LeWM pred. & $m_A$ & 0.482/0.351 & 0.154 & 0.45 & 0.409 & 1.61 & 94.0 & 0.201\\*
 & $m_B$ & 0.566/0.370 & 0.142 & 0.43 & 0.406 & 1.65 & 90.0 & 0.182\\*
 & e & 0.809/0.562 & 0.135 & 0.51 & 0.413 & 1.53 & 80.2 & 0.199\\*
 & v & 0.792/0.560 & 0.079 & 0.47 & 0.449 & 1.90 & 95.2 & 0.216\\
AC enc. & $m_A$ & 0.776/0.759 & 0.132 & 0.18 & 0.065 & 1.03 & 109.5 & 0.076\\*
 & $m_B$ & 0.823/0.745 & 0.113 & 0.18 & 0.067 & 1.03 & 109.5 & 0.106\\*
 & e & 0.903/0.832 & 0.096 & 0.20 & 0.090 & 1.04 & 108.5 & 0.096\\*
 & v & 0.953/0.870 & 0.084 & 0.17 & 0.076 & 1.04 & 109.2 & 0.172\\
AC pred. & $m_A$ & 0.741/0.753 & 0.144 & 0.19 & 0.066 & 1.03 & 109.2 & 0.071\\*
 & $m_B$ & 0.795/0.733 & 0.120 & 0.18 & 0.071 & 1.04 & 109.0 & 0.107\\*
 & e & 0.903/0.832 & 0.106 & 0.20 & 0.090 & 1.04 & 108.3 & 0.100\\*
 & v & 0.944/0.860 & 0.080 & 0.17 & 0.082 & 1.05 & 109.1 & 0.174\\
Cosmos tokenizer & $m_A$ & 0.815/0.770 & 0.390 & 0.30 & 0.072 & 1.13 & 96.6 & 0.107\\*
 & $m_B$ & 0.866/0.753 & 0.317 & 0.32 & 0.072 & 1.14 & 94.4 & 0.137\\*
 & e & 0.891/0.856 & 0.242 & 0.33 & 0.067 & 1.12 & 82.5 & 0.093\\*
 & v & 0.804/0.633 & 0.311 & 0.42 & 0.126 & 1.35 & 89.6 & 0.161\\
Cosmos null & $m_A$ & 0.514/0.595 & 0.110 & 0.21 & 0.170 & 1.20 & 107.6 & 0.116\\*
 & $m_B$ & 0.704/0.560 & 0.085 & 0.19 & 0.165 & 1.19 & 107.2 & 0.123\\*
 & e & 0.815/0.755 & 0.091 & 0.24 & 0.117 & 1.14 & 97.4 & 0.112\\*
 & v & 0.789/0.613 & 0.083 & 0.35 & 0.314 & 1.16 & 96.9 & 0.246\\
Cosmos text & $m_A$ & 0.590/0.458 & 0.079 & 0.20 & 0.469 & 1.45 & 107.1 & 0.126\\*
 & $m_B$ & 0.666/0.530 & 0.100 & 0.21 & 0.417 & 1.44 & 106.5 & 0.127\\*
 & e & 0.749/0.576 & 0.069 & 0.22 & 0.270 & 1.38 & 102.5 & 0.106\\*
 & v & 0.722/0.561 & 0.105 & 0.33 & 0.344 & 1.47 & 102.0 & 0.281\\
\midrule
\multicolumn{9}{l}{\textbf{Bounce}}\\*
VideoSAUR enc. & g & 0.168/0.141 & 0.039 & 0.26 & 0.372 & 1.12 & 118.6 & 0.162\\*
 & h & 0.525/0.147 & 0.051 & 0.26 & 0.375 & 1.12 & 118.1 & 0.197\\*
 & e & 0.565/0.254 & 0.059 & 0.23 & 0.317 & 1.14 & 116.9 & 0.183\\
C-JEPA pred. & g & 0.897/0.362 & 0.115 & 0.31 & 0.402 & 1.28 & 113.4 & 0.181\\*
 & h & 0.896/0.672 & 0.218 & 0.35 & 0.352 & 1.24 & 106.7 & 0.171\\*
 & e & 0.819/0.670 & 0.308 & 0.36 & 0.319 & 1.18 & 98.5 & 0.190\\
LeWM enc. & g & 0.712/0.179 & 0.077 & 0.46 & 0.542 & 1.60 & 113.6 [186] & 0.238\\*
 & h & 0.402/0.188 & 0.064 & 0.39 & 0.485 & 1.37 & 122.2 [191] & 0.248\\*
 & e & 0.729/0.482 & 0.068 & 0.49 & 0.483 & 1.44 & 109.3 & 0.276\\
LeWM pred. & g & 0.980/0.893 & 0.086 & 0.51 & 0.224 & 2.79 & 55.7 & 0.179\\*
 & h & 0.993/0.929 & 0.084 & 0.54 & 0.221 & 2.91 & 62.4 & 0.245\\*
 & e & 0.921/0.861 & 0.113 & 0.63 & 0.198 & 2.84 & 60.6 & 0.138\\
AC enc. & g & 0.995/0.864 & 0.067 & 0.18 & 0.119 & 1.07 & 109.9 & 0.105\\*
 & h & 0.986/0.848 & 0.061 & 0.20 & 0.139 & 1.07 & 108.3 & 0.115\\*
 & e & 0.919/0.814 & 0.078 & 0.20 & 0.139 & 1.07 & 105.6 & 0.146\\
AC pred. & g & 0.994/0.855 & 0.065 & 0.18 & 0.123 & 1.07 & 110.4 & 0.111\\*
 & h & 0.982/0.835 & 0.060 & 0.20 & 0.144 & 1.08 & 109.0 & 0.117\\*
 & e & 0.910/0.812 & 0.076 & 0.20 & 0.148 & 1.07 & 105.4 & 0.167\\
Cosmos tokenizer & g & 0.971/0.924 & 0.343 & 0.49 & 0.116 & 1.13 & 69.9 & 0.168\\*
 & h & 0.972/0.928 & 0.334 & 0.50 & 0.146 & 1.16 & 78.9 & 0.181\\*
 & e & 0.915/0.881 & 0.429 & 0.58 & 0.046 & 1.12 & 76.0 & 0.107\\
Cosmos null & g & 0.948/0.586 & 0.096 & 0.48 & 0.532 & 1.69 & 95.8 & 0.246\\*
 & h & 0.785/0.608 & 0.080 & 0.45 & 0.497 & 1.73 & 100.8 & 0.257\\*
 & e & 0.272/0.489 & 0.100 & 0.47 & 0.538 & 1.62 & 100.5 & 0.299\\
Cosmos text & g & 0.710/0.438 & 0.042 & 0.40 & 0.476 & 1.48 & 107.6 & 0.187\\*
 & h & 0.330/0.343 & 0.051 & 0.37 & 0.453 & 1.50 & 110.0 & 0.193\\*
 & e & 0.088/0.216 & 0.043 & 0.41 & 0.493 & 1.69 & 111.0 & 0.233\\
\midrule
\multicolumn{9}{l}{\textbf{Refraction}}\\*
VideoSAUR enc. & n & 0.608/0.527 & 0.055 & 0.50 & 0.590 & 2.64 & 90.3 & 0.192\\*
 & t & 0.845/0.480 & 0.074 & 0.50 & 0.538 & 1.62 & 96.4 & 0.241\\*
 & $i_0$ & 0.956/0.551 & 0.055 & 0.49 & 0.564 & 2.00 & 98.1 & 0.230\\
C-JEPA pred. & n & 0.575/0.388 & 0.052 & 0.46 & 0.515 & 1.57 & 88.2 & 0.205\\*
 & t & 0.877/0.653 & 0.131 & 0.43 & 0.404 & 1.31 & 95.9 & 0.195\\*
 & $i_0$ & 0.935/0.540 & 0.068 & 0.42 & 0.473 & 1.38 & 93.1 & 0.200\\
LeWM enc. & n & 0.634/0.696 & 0.060 & 0.45 & 0.355 & 1.51 & 52.7 & 0.168\\*
 & t & 0.982/0.940 & 0.206 & 0.50 & 0.229 & 1.46 & 48.2 & 0.097\\*
 & $i_0$ & 0.966/0.856 & 0.146 & 0.49 & 0.312 & 1.65 & 62.3 & 0.129\\
LeWM pred. & n & 0.770/0.702 & 0.076 & 0.46 & 0.340 & 1.55 & 52.9 & 0.141\\*
 & t & 0.968/0.924 & 0.162 & 0.50 & 0.249 & 1.49 & 46.9 & 0.100\\*
 & $i_0$ & 0.980/0.880 & 0.139 & 0.49 & 0.301 & 1.66 & 62.5 & 0.133\\
AC enc. & n & 0.850/0.684 & 0.052 & 0.15 & 0.133 & 1.07 & 115.2 & 0.080\\*
 & t & 0.963/0.903 & 0.087 & 0.15 & 0.055 & 1.04 & 109.5 & 0.027\\*
 & $i_0$ & 0.987/0.927 & 0.057 & 0.16 & 0.075 & 1.05 & 110.4 & 0.061\\
AC pred. & n & 0.859/0.694 & 0.054 & 0.15 & 0.136 & 1.06 & 115.1 & 0.081\\*
 & t & 0.966/0.917 & 0.100 & 0.16 & 0.058 & 1.04 & 108.8 & 0.029\\*
 & $i_0$ & 0.988/0.918 & 0.059 & 0.16 & 0.082 & 1.05 & 109.9 & 0.065\\
Cosmos tokenizer & n & 0.842/0.858 & 0.082 & 0.39 & 0.046 & 1.17 & 49.9 & 0.094\\*
 & t & 0.947/0.840 & 0.180 & 0.20 & 0.040 & 1.11 & 83.3 & 0.038\\*
 & $i_0$ & 0.975/0.919 & 0.099 & 0.30 & 0.041 & 1.10 & 72.8 & 0.020\\
Cosmos null & n & 0.665/0.684 & 0.051 & 0.15 & 0.104 & 1.26 & 101.0 & 0.077\\*
 & t & 0.965/0.891 & 0.151 & 0.21 & 0.156 & 1.24 & 99.5 & 0.028\\*
 & $i_0$ & 0.925/0.839 & 0.054 & 0.12 & 0.087 & 1.17 & 110.5 & 0.024\\
Cosmos text & n & 0.443/0.578 & 0.046 & 0.15 & 0.192 & 1.19 & 104.5 & 0.104\\*
 & t & 0.967/0.891 & 0.128 & 0.22 & 0.174 & 1.21 & 101.5 & 0.054\\*
 & $i_0$ & 0.932/0.781 & 0.054 & 0.14 & 0.170 & 1.20 & 110.1 & 0.069\\
\midrule
\multicolumn{9}{l}{\textbf{Beam splitting}}\\*
VideoSAUR enc. & n & 0.478/0.192 & 0.039 & 0.55 & 0.659 & 1.47 & 113.8 & 0.209\\*
 & $\theta$ & 0.749/0.359 & 0.050 & 0.45 & 0.534 & 1.44 & 108.5 & 0.267\\*
 & $i_0$ & 0.906/0.517 & 0.044 & 0.41 & 0.480 & 1.37 & 107.6 & 0.219\\
C-JEPA pred. & n & 0.502/0.254 & 0.038 & 0.36 & 0.463 & 1.54 & 116.2 & 0.231\\*
 & $\theta$ & 0.614/0.454 & 0.051 & 0.32 & 0.393 & 1.39 & 116.0 & 0.252\\*
 & $i_0$ & 0.927/0.554 & 0.061 & 0.34 & 0.407 & 1.39 & 111.6 & 0.204\\
LeWM enc. & n & 0.898/0.882 & 0.389 & 0.56 & 0.158 & 1.58 & 59.6 & 0.134\\*
 & $\theta$ & 0.994/0.946 & 0.134 & 0.56 & 0.216 & 1.51 & 32.4 & 0.086\\*
 & $i_0$ & 0.988/0.804 & 0.264 & 0.51 & 0.409 & 1.55 & 82.1 & 0.222\\
LeWM pred. & n & 0.903/0.882 & 0.406 & 0.57 & 0.146 & 1.56 & 58.2 & 0.132\\*
 & $\theta$ & 0.987/0.960 & 0.100 & 0.55 & 0.207 & 1.61 & 32.3 & 0.090\\*
 & $i_0$ & 0.990/0.823 & 0.278 & 0.52 & 0.395 & 1.50 & 80.9 & 0.166\\
AC enc. & n & 0.237/0.142 & 0.042 & 0.02 & 0.023 & 1.01 & 119.9 & 0.045\\*
 & $\theta$ & 0.975/0.830 & 0.043 & 0.03 & 0.022 & 1.01 & 119.2 & 0.115\\*
 & $i_0$ & 0.979/0.852 & 0.045 & 0.03 & 0.023 & 1.01 & 118.6 & 0.063\\
AC pred. & n & 0.515/0.087 & 0.041 & 0.05 & 0.074 & 1.02 & 119.5 & 0.048\\*
 & $\theta$ & 0.881/0.608 & 0.044 & 0.06 & 0.067 & 1.02 & 118.6 & 0.115\\*
 & $i_0$ & 0.975/0.581 & 0.045 & 0.05 & 0.063 & 1.02 & 118.4 & 0.064\\
Cosmos tokenizer & n & 0.818/0.620 & 0.047 & 0.05 & 0.047 & 1.06 & 119.4 & 0.044\\*
 & $\theta$ & 0.974/0.901 & 0.049 & 0.19 & 0.047 & 1.10 & 102.8 & 0.073\\*
 & $i_0$ & 0.992/0.946 & 0.058 & 0.15 & 0.031 & 1.04 & 106.0 & 0.041\\
Cosmos null & n & 0.946/0.946 & 0.073 & 0.45 & 0.071 & 1.11 & 98.6 & 0.067\\*
 & $\theta$ & 0.948/0.866 & 0.051 & 0.22 & 0.063 & 1.09 & 91.8 & 0.054\\*
 & $i_0$ & 0.893/0.718 & 0.047 & 0.18 & 0.072 & 1.08 & 88.2 & 0.055\\
Cosmos text & n & 0.326/0.227 & 0.059 & 0.33 & 0.440 & 1.44 & 121.2 & 0.156\\*
 & $\theta$ & 0.594/0.310 & 0.042 & 0.28 & 0.357 & 1.43 & 104.3 & 0.189\\*
 & $i_0$ & 0.742/0.600 & 0.042 & 0.19 & 0.191 & 1.10 & 99.8 & 0.131\\
\midrule
\multicolumn{9}{l}{\textbf{Mechanics + optics}}\\*
VideoSAUR enc. & g & 0.855/0.445 & 0.126 & 0.58 & 0.630 & 1.47 & 113.8 & 0.243\\*
 & h & 0.420/0.202 & 0.049 & 0.43 & 0.588 & 1.79 & 112.8 & 0.188\\
C-JEPA pred. & g & 0.916/0.562 & 0.150 & 0.38 & 0.427 & 1.31 & 110.9 & 0.144\\*
 & h & 0.604/0.208 & 0.048 & 0.38 & 0.501 & 1.62 & 119.4 & 0.182\\
LeWM enc. & g & -0.178/-0.069 & 0.034 & 0.34 & 0.448 & 1.64 & 120.6 & 0.200\\*
 & h & 0.936/0.691 & 0.412 & 0.42 & 0.383 & 1.35 & 107.4 & 0.177\\
LeWM pred. & g & 0.964/0.652 & 0.109 & 0.34 & 0.329 & 1.55 & 108.5 & 0.164\\*
 & h & 0.928/0.770 & 0.394 & 0.40 & 0.369 & 1.38 & 105.5 & 0.155\\
AC enc. & g & 0.993/0.848 & 0.086 & 0.05 & 0.040 & 1.02 & 118.9 & 0.103\\*
 & h & 0.898/0.351 & 0.089 & 0.04 & 0.045 & 1.02 & 118.8 & 0.051\\
AC pred. & g & 0.950/0.634 & 0.090 & 0.07 & 0.091 & 1.03 & 118.5 & 0.125\\*
 & h & 0.918/0.292 & 0.093 & 0.05 & 0.079 & 1.03 & 118.4 & 0.061\\
Cosmos tokenizer & g & 0.953/0.942 & 0.329 & 0.22 & 0.046 & 1.10 & 115.5 & 0.105\\*
 & h & 0.940/0.889 & 0.166 & 0.07 & 0.031 & 1.04 & 116.0 & 0.032\\
Cosmos null & g & 0.958/0.812 & 0.084 & 0.14 & 0.117 & 1.11 & 116.8 & 0.119\\*
 & h & 0.831/0.789 & 0.252 & 0.13 & 0.042 & 1.05 & 112.9 & 0.047\\
Cosmos text & g & 0.715/0.394 & 0.053 & 0.21 & 0.306 & 1.30 & 115.7 & 0.138\\*
 & h & 0.820/0.535 & 0.145 & 0.18 & 0.253 & 1.28 & 113.5 & 0.083\\
\midrule
\multicolumn{9}{l}{\textbf{Heat}}\\*
VideoSAUR enc. & $\alpha$ & 0.567/0.209 & 0.064 & 0.41 & 0.556 & 1.43 & 114.9 & 0.205\\*
 & temp. & 0.663/0.347 & 0.103 & 0.41 & 0.495 & 1.32 & 114.2 & 0.218\\
C-JEPA pred. & $\alpha$ & 0.861/0.186 & 0.082 & 0.31 & 0.419 & 1.33 & 122.0 & 0.152\\*
 & temp. & 0.885/0.683 & 0.261 & 0.34 & 0.337 & 1.21 & 113.7 & 0.155\\
LeWM enc. & $\alpha$ & 0.996/0.995 & 0.800 & 0.59 & 0.035 & 1.33 & 16.3 & 0.046\\*
 & temp. & 0.989/0.986 & 0.666 & 0.55 & 0.090 & 1.13 & 19.7 & 0.057\\
LeWM pred. & $\alpha$ & 0.996/0.996 & 0.791 & 0.59 & 0.039 & 1.42 & 15.8 & 0.041\\*
 & temp. & 0.975/0.982 & 0.623 & 0.57 & 0.102 & 1.19 & 19.6 & 0.069\\
AC enc. & $\alpha$ & 0.991/0.731 & 0.077 & 0.02 & 0.022 & 1.01 & 119.2 & 0.007\\*
 & temp. & 0.979/0.476 & 0.053 & 0.02 & 0.024 & 1.01 & 120.1 & 0.008\\
AC pred. & $\alpha$ & 0.989/0.397 & 0.074 & 0.04 & 0.060 & 1.01 & 119.1 & 0.014\\*
 & temp. & 0.970/0.251 & 0.052 & 0.05 & 0.062 & 1.02 & 120.1 & 0.014\\
Cosmos tokenizer & $\alpha$ & 0.986/0.973 & 0.221 & 0.12 & 0.020 & 1.03 & 117.8 & 0.007\\*
 & temp. & 0.997/0.988 & 0.330 & 0.18 & 0.023 & 1.03 & 116.4 & 0.006\\
Cosmos null & $\alpha$ & 0.981/0.982 & 0.823 & 0.54 & 0.028 & 1.04 & 63.7 & 0.032\\*
 & temp. & 0.993/0.993 & 0.869 & 0.55 & 0.027 & 1.03 & 61.4 & 0.025\\
Cosmos text & $\alpha$ & 0.974/0.965 & 0.734 & 0.49 & 0.124 & 1.08 & 86.1 & 0.080\\*
 & temp. & 0.988/0.984 & 0.814 & 0.51 & 0.071 & 1.05 & 79.6 & 0.059\\
\midrule
\multicolumn{9}{l}{\textbf{Fluid}}\\*
VideoSAUR enc. & re & 0.876/0.277 & 0.083 & 0.54 & 0.670 & 1.39 & 108.2 & 0.292\\
C-JEPA pred. & re & 0.874/0.488 & 0.152 & 0.33 & 0.381 & 1.15 & 114.5 & 0.227\\
LeWM enc. & re & 0.667/0.718 & 0.668 & 0.48 & 0.223 & 1.12 & 82.4 & 0.188\\
LeWM pred. & re & 0.601/0.616 & 0.633 & 0.46 & 0.257 & 1.10 & 75.9 & 0.174\\
AC enc. & re & 0.802/0.816 & 0.152 & 0.06 & 0.021 & 1.01 & 117.8 & 0.182\\
AC pred. & re & 0.833/0.668 & 0.151 & 0.07 & 0.062 & 1.02 & 117.4 & 0.193\\
Cosmos tokenizer & re & 0.783/0.825 & 0.306 & 0.31 & 0.019 & 1.03 & 100.6 & 0.154\\
Cosmos null & re & 0.823/0.820 & 0.235 & 0.33 & 0.086 & 1.09 & 96.5 & 0.153\\
Cosmos text & re & 0.830/0.844 & 0.176 & 0.30 & 0.086 & 1.06 & 100.4 & 0.163\\
\midrule
\multicolumn{9}{l}{\textbf{Viscoelastic recovery}}\\*
VideoSAUR enc. & $\beta$ & 0.407/0.227 & 0.041 & 0.59 & 0.723 & 4.01 & 96.1 & 0.331\\*
 & compr. & 0.916/0.431 & 0.091 & 0.42 & 0.517 & 1.30 & 103.5 & 0.386\\
C-JEPA pred. & $\beta$ & 0.420/0.368 & 0.047 & 0.48 & 0.546 & 2.01 & 91.8 & 0.235\\*
 & compr. & 0.965/0.747 & 0.186 & 0.41 & 0.366 & 1.35 & 100.7 & 0.287\\
LeWM enc. & $\beta$ & 0.128/0.094 [13] & 0.038 & 0.89 [13] & 0.890 & 2.39 & 180.0 [48] & 0.315 [76]\\*
 & compr. & 0.997/0.969 & 0.191 & 0.57 & 0.187 & 1.41 & 31.5 & 0.079\\
LeWM pred. & $\beta$ & 0.726/0.794 & 0.173 & 0.51 & 0.156 & 2.12 & 77.7 & 0.155\\*
 & compr. & 0.992/0.956 & 0.123 & 0.54 & 0.210 & 1.42 & 38.6 & 0.103\\
AC enc. & $\beta$ & 0.958/0.887 & 0.055 & 0.28 & 0.153 & 1.11 & 93.1 & 0.105\\*
 & compr. & 0.997/0.946 & 0.106 & 0.19 & 0.045 & 1.03 & 107.4 & 0.226\\
AC pred. & $\beta$ & 0.985/0.877 & 0.068 & 0.30 & 0.195 & 1.16 & 93.1 & 0.113\\*
 & compr. & 0.996/0.926 & 0.119 & 0.21 & 0.094 & 1.04 & 106.3 & 0.205\\
Cosmos tokenizer & $\beta$ & 0.911/0.888 & 0.610 & 0.45 & 0.083 & 1.13 & 70.7 & 0.102\\*
 & compr. & 0.990/0.957 & 0.286 & 0.49 & 0.019 & 1.04 & 54.6 & 0.051\\
Cosmos null & $\beta$ & 0.869/0.791 & 0.062 & 0.48 & 0.407 & 2.46 & 107.6 & 0.112\\*
 & compr. & 0.993/0.977 & 0.259 & 0.44 & 0.105 & 1.12 & 77.4 & 0.109\\
Cosmos text & $\beta$ & 0.702/0.255 & 0.052 & 0.42 & 0.443 & 1.88 & 111.0 & 0.149\\*
 & compr. & 0.563/0.486 & 0.073 & 0.37 & 0.401 & 1.53 & 98.0 & 0.276\\
\end{longtable}
\endgroup

\section{Perturbation Design, Errors, and Coverage}
\label{app:scales}

\subsection{Repeatability and Magnitude Sweeps}

The preliminary study examines repeated execution on the same input,
paired responses across random seeds, and changing perturbation magnitude
at a fixed realization. Its paired-seed experiment contains 58 input records
and 722 completed calls. The six-magnitude pilot contains 78 records and
630 calls. The main spring, heat, and bounce studies each contain 1,080
records and 7,560 calls: 24 contexts, three anchors, 13 physical branches,
and two reference re-renders per anchor. The completed work totals
$722+630+3(7{,}560)=24{,}032$ calls, excluding interrupted attempts.

The fine-scale analysis uses the declared first observation at each primary
interface. Including both RGB references, each scene yields 648
context-anchor-observation families and 6,480 finer signed queries. The first output is
one native unit: a slot/state vector, AC's two-frame tubelet, an input-tokenizer
latent slice, or one generated RGB image. Native temporal support therefore
need not be one RGB frame.
All 72 families per observation remain in evaluation. The atlas retains its
additional LeWM and AC encoder interfaces, whereas the scale study reports
their predictors.

\subsection{Fitting Diagnostics and Repeatability}

Table~\ref{tab:scaleerror} reports the finest-magnitude error for every
subject. We restore the first atlas fold's saved weights and multiplier for
spring stiffness, heat diffusivity, and bounce gravity, respectively. No
fine-scale observation is used to select this fold or update a weight.
New references normalize each context-anchor family by exactly the same
rules as in the atlas. The entire native output is predicted and scored.
Pointwise ratio quartiles and equal-calibration PL scores appear in
Table~\ref{tab:heatprofiles}.

\begin{table}[h]
\caption{Full-response median error ratios at $|s|=\log(1.5)/32$. The neural predictor is frozen from atlas fold zero and receives only three coarse references in each new context. Each cell includes 144 signed queries. A ratio of one equals the unchanged-output baseline.}
\label{tab:scaleerror}
\centering\small
\setlength{\tabcolsep}{3pt}
\begin{tabular}{@{}lrrr@{}}
\toprule
Observation & Spring & Heat & Bounce\\
\midrule
VideoSAUR enc. & 0.982 & 1.279 & 0.975\\
C-JEPA pred. & 1.020 & 1.832 & 0.989\\
LeWM pred. & 1.009 & 0.562 & 1.014\\
AC pred. & 0.974 & 1.042 & 0.967\\
Cosmos tokenizer & 1.021 & 0.967 & 1.003\\
Cosmos null & 1.007 & 0.970 & 0.991\\
Cosmos text & 0.986 & 0.958 & 0.988\\
RGB last input & 1.001 & 0.446 & 0.990\\
RGB first future & 0.998 & 0.417 & 0.990\\
\bottomrule
\end{tabular}
\end{table}

\begin{table}[h]
\caption{Finest-scale heat prediction. Neural and PL columns use complete outputs and the same three references. Quartiles describe neural pointwise error ratios. Reverse uses the same frozen atlas tool and is scored against zero signed intervention. This denominator differs from the atlas training-mean score.}
\label{tab:heatprofiles}
\centering\small
\setlength{\tabcolsep}{3pt}
\begin{tabular}{@{}lrrrrr@{}}
\toprule
Observation & Neural & PL & $q_{25}$ & $q_{75}$ & Reverse $R^2_0$\\
\midrule
VideoSAUR enc. & 1.279 & 0.972 & 0.985 & 2.752 & -11.426\\
C-JEPA pred. & 1.832 & 0.912 & 1.183 & 3.620 & -7.608\\
LeWM pred. & 0.562 & 0.551 & 0.484 & 0.616 & 0.990\\
AC pred. & 1.042 & 0.986 & 1.017 & 1.076 & -87.156\\
Cosmos tokenizer & 0.967 & 0.975 & 0.960 & 0.972 & -28.363\\
Cosmos null & 0.970 & 0.971 & 0.962 & 0.980 & 0.531\\
Cosmos text & 0.958 & 0.983 & 0.949 & 0.965 & -22.853\\
RGB last input & 0.446 & 0.210 & 0.310 & 0.472 & 0.997\\
RGB first future & 0.417 & 0.207 & 0.277 & 0.531 & 0.998\\
\bottomrule
\end{tabular}
\end{table}

Spring's 144 reference comparisons per subject have identical inputs and
outputs. In heat, all 144 comparisons differ under separate rendering streams.
All 24 preliminary same-input, same-seed generation repeats reproduce the
first RGB frame exactly. Fine-response direction agreement across three
generation seeds varies by scene and configuration: median pairwise cosine
ranges from $-0.006$ to 0.542. These preliminary observations do not establish
seed-independent response laws. The main sweeps follow fixed realizations.

Contact timing provides an additional physical description of bounce.
Before model inference, each input receives a source-event classification.
Among 864 nonzero conditions, 285 change contact signatures, 451 retain them,
and 128 have a sampled frame within one simulation step of contact.
Near-contact classification takes precedence. Endpoint and coarse-calibration
strata are recorded independently. Intermediate events require observations
inside those intervals.

\subsection{What the Scale Scores Compare}

Forward ratios compare each query error with the magnitude of that query's
complete response. They do not divide by a fitted low-rank component or
omit an unpredictable remainder. Zero response makes that ratio undefined.
Absolute errors remain available for every query. The
144 points at each scale give equal weight to 24 contexts, three anchors,
and two signs. Their dependence precludes treating them as independent
replicates for significance testing.

Reverse uses the frozen native or RGB-trained tool with the same three
references. Its fine-scale score is $R^2_0=1-\sum(\widehat s-s)^2/\sum s^2$,
relative to the known zero-intervention baseline. A narrow symmetric intervention range can make even small
parameter errors large under this score. These definitions and complete
point losses are retained separately from the atlas results.

This supplementary experiment tests a fixed learned law after coarse local calibration.
It does not test whether fitting on the fine-scale targets could improve
performance, nor whether success at the sampled steps extends to all smaller
steps. That extension requires additional regularity assumptions.

\subsection{Direct True-Future Comparisons}
\label{app:directfuture}

The direct comparison uses the same 24 contexts, three reference anchors,
two signs and five fine magnitudes. Each of the four predictive observations
has 720 queries per scene, or 2,160 across scenes. The true future is frame
16 at 0.08 seconds after the original sixteen-frame input. Its source
conditions and observation time match the model's first prediction. All
three scenes retain complete first-future RGB coverage. Cosmos null/text
use their saved images with HWC coordinates converted to the simulator's
CHW ordering before any subtraction. No response estimator is fitted.

\small
\begin{longtable}{@{}llrrrrr@{}}
\caption{Generated futures compared directly with true first-future RGB, with no fitted response estimator. All errors use every pixel in native $[0,1]$ display units. Future RMS compares complete images. Effect error compares branch-minus-reference responses. True effect is the simulator response RMS. Ratio is the median of pointwise effect-error/true-effect ratios. Each row contains 144 queries.}\label{tab:physicalscales}\\
\toprule
Scene & Subject & $|s|/\delta_0$ & Future RMS & Effect error & True effect & Ratio\\
\midrule
\endfirsthead
\toprule
Scene & Subject & $|s|/\delta_0$ & Future RMS & Effect error & True effect & Ratio\\
\midrule
\endhead
Spring & Cosmos null & 1/2 & 0.0480 & 0.0314 & 0.0354 & 0.916\\
Spring & Cosmos null & 1/4 & 0.0497 & 0.0288 & 0.0312 & 1.063\\
Spring & Cosmos null & 1/8 & 0.0520 & 0.0257 & 0.0237 & 1.111\\
Spring & Cosmos null & 1/16 & 0.0501 & 0.0181 & 0.0165 & 1.124\\
Spring & Cosmos null & 1/32 & 0.0502 & 0.0117 & 0.0105 & 1.118\\
Spring & Cosmos text & 1/2 & 0.0333 & 0.0393 & 0.0354 & 1.195\\
Spring & Cosmos text & 1/4 & 0.0312 & 0.0357 & 0.0312 & 1.273\\
Spring & Cosmos text & 1/8 & 0.0339 & 0.0307 & 0.0237 & 1.339\\
Spring & Cosmos text & 1/16 & 0.0335 & 0.0236 & 0.0165 & 1.420\\
Spring & Cosmos text & 1/32 & 0.0325 & 0.0180 & 0.0105 & 1.708\\
Heat & Cosmos null & 1/2 & 0.0070 & 0.0020 & 0.0040 & 0.499\\
Heat & Cosmos null & 1/4 & 0.0070 & 0.0017 & 0.0020 & 0.862\\
Heat & Cosmos null & 1/8 & 0.0070 & 0.0016 & 0.0010 & 1.606\\
Heat & Cosmos null & 1/16 & 0.0070 & 0.0015 & 0.0005 & 3.022\\
Heat & Cosmos null & 1/32 & 0.0070 & 0.0015 & 0.0002 & 5.843\\
Heat & Cosmos text & 1/2 & 0.0084 & 0.0044 & 0.0040 & 1.122\\
Heat & Cosmos text & 1/4 & 0.0083 & 0.0040 & 0.0020 & 2.023\\
Heat & Cosmos text & 1/8 & 0.0084 & 0.0038 & 0.0010 & 3.813\\
Heat & Cosmos text & 1/16 & 0.0083 & 0.0037 & 0.0005 & 7.582\\
Heat & Cosmos text & 1/32 & 0.0083 & 0.0036 & 0.0002 & 14.605\\
Bounce & Cosmos null & 1/2 & 0.1802 & 0.0283 & 0.0234 & 1.262\\
Bounce & Cosmos null & 1/4 & 0.1806 & 0.0227 & 0.0189 & 1.243\\
Bounce & Cosmos null & 1/8 & 0.1769 & 0.0193 & 0.0166 & 1.157\\
Bounce & Cosmos null & 1/16 & 0.1792 & 0.0150 & 0.0119 & 1.137\\
Bounce & Cosmos null & 1/32 & 0.1792 & 0.0112 & 0.0096 & 1.191\\
Bounce & Cosmos text & 1/2 & 0.0080 & 0.0095 & 0.0234 & 0.482\\
Bounce & Cosmos text & 1/4 & 0.0073 & 0.0074 & 0.0189 & 0.528\\
Bounce & Cosmos text & 1/8 & 0.0069 & 0.0069 & 0.0166 & 0.665\\
Bounce & Cosmos text & 1/16 & 0.0067 & 0.0065 & 0.0119 & 0.752\\
Bounce & Cosmos text & 1/32 & 0.0068 & 0.0056 & 0.0096 & 0.922\\
\bottomrule
\end{longtable}
\normalsize

For C-JEPA, the frozen VideoSAUR target encoder is recurrent. The direct
test therefore replays the original history locally at float32 precision,
predicts its first future slots with the unchanged C-JEPA checkpoint, and
continues the same unquantized slot state with the true future image to
obtain the target. Future pixels never enter the predictor. This paired
replay avoids comparing a target from one recurrent trajectory with a
prediction from another: small cross-device discrepancies in historical
slot extraction can accumulate over the input sequence. No slot matching,
target projection, or learned alignment is added. The original atlas
observations remain unchanged and separate from this paired replay.

LeWM's saved first predictor output precedes its released prediction
projection. We apply that fixed \texttt{pred\_proj} to each absolute output,
then compare it with the true future encoded by the released encoder and
\texttt{projector}. Projection is applied before branch-minus-reference
subtraction. Bilinear antialiased resizing and original normalization are
retained. The true target is stored in float32, while the saved prediction
retains its original fp16 precision. Replayed historical LeWM encodings and
known one-step C-JEPA state transitions each pass a relative RMS tolerance
of 0.002, with per-record errors retained. Neither target encoder is trained.

\small
\begin{longtable}{@{}llrrrrr@{}}
\caption{Predicted latents compared with encoded true first-future frames. Each row retains 144 queries. C-JEPA predicts from the original history in a local frozen replay, and its target continues the same float32 slot state. LeWM applies its released prediction projector before comparison with the target encoder. Future and persistence errors are in each model's own latent units. Ratios are medians of pointwise ratios. Values above one lose to the respective baseline.}\label{tab:latentfutures}\\
\toprule
Scene & Subject & $|s|/\delta_0$ & Future RMS & Persist. RMS & Future/persist. & Effect ratio\\
\midrule
\endfirsthead
\toprule
Scene & Subject & $|s|/\delta_0$ & Future RMS & Persist. RMS & Future/persist. & Effect ratio\\
\midrule
\endhead
Spring & C-JEPA & 1/2 & 0.4122 & 0.0681 & 5.81 & 1.035\\
Spring & C-JEPA & 1/4 & 0.4083 & 0.0664 & 6.09 & 1.044\\
Spring & C-JEPA & 1/8 & 0.4119 & 0.0673 & 6.18 & 1.028\\
Spring & C-JEPA & 1/16 & 0.4052 & 0.0641 & 6.27 & 1.025\\
Spring & C-JEPA & 1/32 & 0.4139 & 0.0617 & 6.39 & 1.032\\
Spring & LeWM & 1/2 & 0.8279 & 0.0175 & 46.05 & 0.999\\
Spring & LeWM & 1/4 & 0.8274 & 0.0185 & 43.45 & 1.056\\
Spring & LeWM & 1/8 & 0.8280 & 0.0154 & 53.66 & 1.045\\
Spring & LeWM & 1/16 & 0.8283 & 0.0110 & 74.46 & 1.038\\
Spring & LeWM & 1/32 & 0.8282 & 0.0096 & 82.45 & 1.033\\
Heat & C-JEPA & 1/2 & 0.4158 & 0.0533 & 7.97 & 1.031\\
Heat & C-JEPA & 1/4 & 0.4166 & 0.0541 & 7.84 & 1.032\\
Heat & C-JEPA & 1/8 & 0.4166 & 0.0525 & 7.99 & 1.029\\
Heat & C-JEPA & 1/16 & 0.4173 & 0.0535 & 7.83 & 1.027\\
Heat & C-JEPA & 1/32 & 0.4180 & 0.0527 & 7.92 & 1.029\\
Heat & LeWM & 1/2 & 0.4262 & 0.0023 & 183.05 & 0.972\\
Heat & LeWM & 1/4 & 0.4261 & 0.0023 & 182.71 & 0.973\\
Heat & LeWM & 1/8 & 0.4261 & 0.0023 & 181.51 & 0.977\\
Heat & LeWM & 1/16 & 0.4261 & 0.0023 & 181.72 & 0.989\\
Heat & LeWM & 1/32 & 0.4261 & 0.0023 & 181.48 & 1.013\\
Bounce & C-JEPA & 1/2 & 0.4347 & 0.0546 & 7.17 & 1.118\\
Bounce & C-JEPA & 1/4 & 0.4333 & 0.0419 & 10.35 & 1.117\\
Bounce & C-JEPA & 1/8 & 0.4422 & 0.0522 & 8.28 & 1.120\\
Bounce & C-JEPA & 1/16 & 0.4415 & 0.0492 & 8.66 & 1.121\\
Bounce & C-JEPA & 1/32 & 0.4423 & 0.0563 & 7.61 & 1.065\\
Bounce & LeWM & 1/2 & 0.6264 & 0.0132 & 47.24 & 1.004\\
Bounce & LeWM & 1/4 & 0.6251 & 0.0082 & 80.58 & 1.003\\
Bounce & LeWM & 1/8 & 0.6256 & 0.0093 & 74.92 & 1.035\\
Bounce & LeWM & 1/16 & 0.6271 & 0.0066 & 95.22 & 1.007\\
Bounce & LeWM & 1/32 & 0.6268 & 0.0079 & 81.17 & 1.004\\
\bottomrule
\end{longtable}
\normalsize

Both latent interfaces retain every coordinate and compare whole-future error
with persistence of the last observed latent. At the finest heat scale,
LeWM's median future RMS is 0.4261 and persistence RMS is 0.0023. The
median pointwise ratio is 181.48. Large ratios can therefore reflect both a poor prediction and a
small physical temporal change. Response error in Equation~\ref{eq:trueeffect}
compares branch-minus-reference futures and normalizes by the true
intervention effect. These two denominators answer different questions.
Undefined ratios remain missing with their absolute errors retained.

AC's first prediction represents frames 16 and 17. Only frame 16 has been
rendered under all matched conditions, so a same-time target tubelet cannot
yet be constructed. Repeating frame 16 or appending the final input would
define a different target and is not used. VideoSAUR and tokenizer are not
future predictors in the declared observation set. The resulting missing
comparisons are not counted as prediction failures.

\subsection{Effect and Precision Diagnostics}
\label{app:effectdiagnostics}

The pointwise amplitude ratio and cosine use the complete effect vectors.
Their exact relation to error is
$e^2=A^2+1-2Ac$ when both effects are nonzero. We retain zero-denominator
missing counts. Odd and even mismatch vectors are half the difference and
half the sum of the positive/negative effect errors. Their squared RMS
energies add exactly before any median or bootstrap. Reusing the last-input
branch-minus-reference effect supplies a same-space persistence-effect
baseline, different from either zero effect or temporal persistence of a
whole observation. Table~\ref{tab:effectdiagnostics} retains all scales.

\small
\begin{longtable}{@{}llrrrrr@{}}
\caption{Complete-effect diagnostics at every scale. $A=\|\Delta Z\|/\|\Delta T\|$, $c$ is their cosine, $e$ is the physical-effect error ratio, and $e_{\rm persist}$ reuses the last-input intervention effect in the same coordinates. Brackets give 95\% paired context-block intervals. All summaries are pointwise medians with 144 queries.}\label{tab:effectdiagnostics}\\
\toprule
Scene & Subject & Scale & $A$ & $c$ & $e$ [95\% interval] & $e_{\rm persist}$\\
\midrule
\endfirsthead
\toprule
Scene & Subject & Scale & $A$ & $c$ & $e$ [95\% interval] & $e_{\rm persist}$\\
\midrule
\endhead
Spring & C-JEPA & 1/2 & 0.352 & 0.038 & 1.035 [1.022, 1.083] & 0.560\\
Spring & C-JEPA & 1/4 & 0.303 & -0.008 & 1.044 [1.028, 1.072] & 0.779\\
Spring & C-JEPA & 1/8 & 0.259 & 0.023 & 1.028 [1.012, 1.048] & 0.712\\
Spring & C-JEPA & 1/16 & 0.251 & 0.016 & 1.025 [1.011, 1.048] & 0.629\\
Spring & C-JEPA & 1/32 & 0.265 & 0.018 & 1.032 [1.018, 1.083] & 0.775\\
Spring & LeWM & 1/2 & 0.277 & 0.175 & 0.999 [0.988, 1.008] & 0.874\\
Spring & LeWM & 1/4 & 0.306 & -0.035 & 1.056 [1.041, 1.067] & 1.567\\
Spring & LeWM & 1/8 & 0.218 & -0.123 & 1.045 [1.036, 1.054] & 1.356\\
Spring & LeWM & 1/16 & 0.202 & -0.109 & 1.038 [1.029, 1.048] & 1.291\\
Spring & LeWM & 1/32 & 0.202 & -0.078 & 1.033 [1.026, 1.040] & 1.357\\
Spring & Cosmos null & 1/2 & 1.011 & 0.574 & 0.916 [0.880, 0.941] & 1.129\\
Spring & Cosmos null & 1/4 & 0.953 & 0.389 & 1.063 [0.996, 1.095] & 1.514\\
Spring & Cosmos null & 1/8 & 0.838 & 0.115 & 1.111 [1.081, 1.147] & 1.385\\
Spring & Cosmos null & 1/16 & 0.660 & 0.033 & 1.124 [1.100, 1.156] & 1.337\\
Spring & Cosmos null & 1/32 & 0.628 & 0.018 & 1.118 [1.096, 1.155] & 1.341\\
Spring & Cosmos text & 1/2 & 1.264 & 0.436 & 1.195 [1.096, 1.343] & 1.129\\
Spring & Cosmos text & 1/4 & 1.112 & 0.266 & 1.273 [1.147, 1.344] & 1.514\\
Spring & Cosmos text & 1/8 & 1.091 & 0.095 & 1.339 [1.236, 1.454] & 1.385\\
Spring & Cosmos text & 1/16 & 1.147 & 0.041 & 1.420 [1.361, 1.564] & 1.337\\
Spring & Cosmos text & 1/32 & 1.456 & 0.024 & 1.708 [1.575, 2.023] & 1.341\\
Heat & C-JEPA & 1/2 & 0.246 & -0.004 & 1.031 [1.020, 1.049] & 0.673\\
Heat & C-JEPA & 1/4 & 0.259 & -0.017 & 1.032 [1.021, 1.045] & 0.693\\
Heat & C-JEPA & 1/8 & 0.243 & -0.022 & 1.029 [1.020, 1.047] & 0.748\\
Heat & C-JEPA & 1/16 & 0.233 & -0.007 & 1.027 [1.019, 1.047] & 0.784\\
Heat & C-JEPA & 1/32 & 0.248 & -0.000 & 1.029 [1.015, 1.048] & 0.798\\
Heat & LeWM & 1/2 & 0.381 & 0.272 & 0.972 [0.963, 0.983] & 0.067\\
Heat & LeWM & 1/4 & 0.382 & 0.267 & 0.973 [0.966, 0.984] & 0.078\\
Heat & LeWM & 1/8 & 0.385 & 0.261 & 0.977 [0.965, 0.986] & 0.111\\
Heat & LeWM & 1/16 & 0.408 & 0.236 & 0.989 [0.979, 0.999] & 0.183\\
Heat & LeWM & 1/32 & 0.468 & 0.216 & 1.013 [0.994, 1.026] & 0.368\\
Heat & Cosmos null & 1/2 & 1.092 & 0.895 & 0.499 [0.485, 0.507] & 0.106\\
Heat & Cosmos null & 1/4 & 1.312 & 0.756 & 0.862 [0.853, 0.881] & 0.117\\
Heat & Cosmos null & 1/8 & 1.880 & 0.536 & 1.606 [1.580, 1.622] & 0.129\\
Heat & Cosmos null & 1/16 & 3.176 & 0.312 & 3.022 [2.981, 3.062] & 0.130\\
Heat & Cosmos null & 1/32 & 5.933 & 0.169 & 5.843 [5.733, 5.955] & 0.126\\
Heat & Cosmos text & 1/2 & 1.462 & 0.651 & 1.122 [1.077, 1.189] & 0.106\\
Heat & Cosmos text & 1/4 & 2.222 & 0.425 & 2.023 [1.942, 2.154] & 0.117\\
Heat & Cosmos text & 1/8 & 3.929 & 0.232 & 3.813 [3.722, 3.997] & 0.129\\
Heat & Cosmos text & 1/16 & 7.615 & 0.129 & 7.582 [7.257, 7.739] & 0.130\\
Heat & Cosmos text & 1/32 & 14.676 & 0.068 & 14.605 [14.163, 14.960] & 0.126\\
Bounce & C-JEPA & 1/2 & 0.678 & 0.095 & 1.118 [1.078, 1.156] & 0.529\\
Bounce & C-JEPA & 1/4 & 0.615 & 0.038 & 1.117 [1.080, 1.197] & 0.639\\
Bounce & C-JEPA & 1/8 & 0.595 & 0.047 & 1.120 [1.066, 1.235] & 0.726\\
Bounce & C-JEPA & 1/16 & 0.556 & 0.031 & 1.121 [1.072, 1.220] & 0.737\\
Bounce & C-JEPA & 1/32 & 0.460 & 0.055 & 1.065 [1.044, 1.130] & 0.826\\
Bounce & LeWM & 1/2 & 0.318 & 0.181 & 1.004 [0.992, 1.026] & 0.678\\
Bounce & LeWM & 1/4 & 0.312 & 0.165 & 1.003 [0.994, 1.020] & 0.703\\
Bounce & LeWM & 1/8 & 0.364 & 0.140 & 1.035 [1.007, 1.049] & 0.991\\
Bounce & LeWM & 1/16 & 0.349 & 0.129 & 1.007 [0.998, 1.034] & 0.972\\
Bounce & LeWM & 1/32 & 0.305 & 0.136 & 1.004 [0.999, 1.017] & 0.922\\
Bounce & Cosmos null & 1/2 & 1.141 & 0.300 & 1.262 [1.131, 1.368] & 0.642\\
Bounce & Cosmos null & 1/4 & 1.068 & 0.302 & 1.243 [1.117, 1.445] & 0.802\\
Bounce & Cosmos null & 1/8 & 0.985 & 0.255 & 1.157 [1.087, 1.241] & 1.070\\
Bounce & Cosmos null & 1/16 & 0.957 & 0.188 & 1.137 [1.110, 1.175] & 1.190\\
Bounce & Cosmos null & 1/32 & 0.947 & 0.193 & 1.191 [1.157, 1.242] & 1.207\\
Bounce & Cosmos text & 1/2 & 1.038 & 0.892 & 0.482 [0.398, 0.550] & 0.642\\
Bounce & Cosmos text & 1/4 & 1.058 & 0.869 & 0.528 [0.432, 0.611] & 0.802\\
Bounce & Cosmos text & 1/8 & 1.056 & 0.822 & 0.665 [0.519, 0.804] & 1.070\\
Bounce & Cosmos text & 1/16 & 1.053 & 0.755 & 0.752 [0.650, 0.879] & 1.190\\
Bounce & Cosmos text & 1/32 & 1.055 & 0.667 & 0.922 [0.744, 1.104] & 1.207\\
\bottomrule
\end{longtable}
\normalsize

For each context-anchor family, subtracting the reference error
$Z(0)-T(0)$ from every prediction gives corrected whole-future error
$[Z(s)-T(s)]-[Z(0)-T(0)]$, exactly the effect-error vector. This removes
a constant coordinate offset without fitting on query truths, but requires
the physical reference truth and is not a deployable forecast. Target-space projection, time indices and
known-history replay checks are verified, but no matched training-domain
positive-control run has been performed.

\small
\begin{longtable}{@{}llrrrrrr@{}}
\caption{Parity and constant-offset diagnostics. Even share is the median fraction of paired squared error from the even mismatch. Odd and even energies sum exactly before aggregation. True even is the median fraction of paired true-effect energy in its even part. Debiased future subtracts the reference branch's prediction-minus-target vector, using no query fit, and therefore equals absolute effect error. The last three columns are observation-specific RMS errors. Temporal persistence is a distinct baseline from persistence of intervention effects.}\label{tab:parityoffset}\\
\toprule
Scene & Subject & Scale & Even share & True even & Future & Debiased & Persist.\\
\midrule
\endfirsthead
\toprule
Scene & Subject & Scale & Even share & True even & Future & Debiased & Persist.\\
\midrule
\endhead
Spring & C-JEPA & 1/2 & 0.653 & 0.658 & 0.4122 & 0.2334 & 0.0681\\
Spring & C-JEPA & 1/4 & 0.620 & 0.599 & 0.4083 & 0.2061 & 0.0664\\
Spring & C-JEPA & 1/8 & 0.640 & 0.655 & 0.4119 & 0.1726 & 0.0673\\
Spring & C-JEPA & 1/16 & 0.626 & 0.616 & 0.4052 & 0.1877 & 0.0641\\
Spring & C-JEPA & 1/32 & 0.616 & 0.606 & 0.4139 & 0.1535 & 0.0617\\
Spring & LeWM & 1/2 & 0.579 & 0.596 & 0.8279 & 0.0234 & 0.0175\\
Spring & LeWM & 1/4 & 0.489 & 0.515 & 0.8274 & 0.0167 & 0.0185\\
Spring & LeWM & 1/8 & 0.323 & 0.308 & 0.8280 & 0.0115 & 0.0154\\
Spring & LeWM & 1/16 & 0.147 & 0.153 & 0.8283 & 0.0063 & 0.0110\\
Spring & LeWM & 1/32 & 0.113 & 0.101 & 0.8282 & 0.0032 & 0.0096\\
Spring & Cosmos null & 1/2 & 0.728 & 0.685 & 0.0480 & 0.0314 & 0.0321\\
Spring & Cosmos null & 1/4 & 0.718 & 0.616 & 0.0497 & 0.0288 & 0.0310\\
Spring & Cosmos null & 1/8 & 0.608 & 0.561 & 0.0520 & 0.0257 & 0.0279\\
Spring & Cosmos null & 1/16 & 0.495 & 0.478 & 0.0501 & 0.0181 & 0.0239\\
Spring & Cosmos null & 1/32 & 0.411 & 0.390 & 0.0502 & 0.0117 & 0.0233\\
Spring & Cosmos text & 1/2 & 0.687 & 0.685 & 0.0333 & 0.0393 & 0.0321\\
Spring & Cosmos text & 1/4 & 0.700 & 0.616 & 0.0312 & 0.0357 & 0.0310\\
Spring & Cosmos text & 1/8 & 0.664 & 0.561 & 0.0339 & 0.0307 & 0.0279\\
Spring & Cosmos text & 1/16 & 0.596 & 0.478 & 0.0335 & 0.0236 & 0.0239\\
Spring & Cosmos text & 1/32 & 0.565 & 0.390 & 0.0325 & 0.0180 & 0.0233\\
Heat & C-JEPA & 1/2 & 0.422 & 0.424 & 0.4158 & 0.0279 & 0.0533\\
Heat & C-JEPA & 1/4 & 0.430 & 0.431 & 0.4166 & 0.0198 & 0.0541\\
Heat & C-JEPA & 1/8 & 0.314 & 0.308 & 0.4166 & 0.0109 & 0.0525\\
Heat & C-JEPA & 1/16 & 0.242 & 0.236 & 0.4173 & 0.0054 & 0.0535\\
Heat & C-JEPA & 1/32 & 0.164 & 0.158 & 0.4180 & 0.0031 & 0.0527\\
Heat & LeWM & 1/2 & 0.007 & 0.006 & 0.4262 & 0.0051 & 0.0023\\
Heat & LeWM & 1/4 & 0.003 & 0.002 & 0.4261 & 0.0025 & 0.0023\\
Heat & LeWM & 1/8 & 0.006 & 0.000 & 0.4261 & 0.0013 & 0.0023\\
Heat & LeWM & 1/16 & 0.019 & 0.000 & 0.4261 & 0.0006 & 0.0023\\
Heat & LeWM & 1/32 & 0.053 & 0.000 & 0.4261 & 0.0003 & 0.0023\\
Heat & Cosmos null & 1/2 & 0.619 & 0.011 & 0.0070 & 0.0020 & 0.0047\\
Heat & Cosmos null & 1/4 & 0.669 & 0.004 & 0.0070 & 0.0017 & 0.0047\\
Heat & Cosmos null & 1/8 & 0.710 & 0.001 & 0.0070 & 0.0016 & 0.0047\\
Heat & Cosmos null & 1/16 & 0.720 & 0.000 & 0.0070 & 0.0015 & 0.0047\\
Heat & Cosmos null & 1/32 & 0.730 & 0.000 & 0.0070 & 0.0015 & 0.0047\\
Heat & Cosmos text & 1/2 & 0.677 & 0.011 & 0.0084 & 0.0044 & 0.0047\\
Heat & Cosmos text & 1/4 & 0.709 & 0.004 & 0.0083 & 0.0040 & 0.0047\\
Heat & Cosmos text & 1/8 & 0.722 & 0.001 & 0.0084 & 0.0038 & 0.0047\\
Heat & Cosmos text & 1/16 & 0.729 & 0.000 & 0.0083 & 0.0037 & 0.0047\\
Heat & Cosmos text & 1/32 & 0.741 & 0.000 & 0.0083 & 0.0036 & 0.0047\\
Bounce & C-JEPA & 1/2 & 0.679 & 0.672 & 0.4347 & 0.2396 & 0.0546\\
Bounce & C-JEPA & 1/4 & 0.644 & 0.604 & 0.4333 & 0.2024 & 0.0419\\
Bounce & C-JEPA & 1/8 & 0.640 & 0.657 & 0.4422 & 0.1841 & 0.0522\\
Bounce & C-JEPA & 1/16 & 0.653 & 0.658 & 0.4415 & 0.1715 & 0.0492\\
Bounce & C-JEPA & 1/32 & 0.657 & 0.702 & 0.4423 & 0.1413 & 0.0563\\
Bounce & LeWM & 1/2 & 0.571 & 0.565 & 0.6264 & 0.0382 & 0.0132\\
Bounce & LeWM & 1/4 & 0.631 & 0.653 & 0.6251 & 0.0336 & 0.0082\\
Bounce & LeWM & 1/8 & 0.575 & 0.561 & 0.6256 & 0.0243 & 0.0093\\
Bounce & LeWM & 1/16 & 0.592 & 0.586 & 0.6271 & 0.0157 & 0.0066\\
Bounce & LeWM & 1/32 & 0.438 & 0.415 & 0.6268 & 0.0100 & 0.0079\\
Bounce & Cosmos null & 1/2 & 0.653 & 0.659 & 0.1802 & 0.0283 & 0.0094\\
Bounce & Cosmos null & 1/4 & 0.704 & 0.639 & 0.1806 & 0.0227 & 0.0085\\
Bounce & Cosmos null & 1/8 & 0.643 & 0.585 & 0.1769 & 0.0193 & 0.0095\\
Bounce & Cosmos null & 1/16 & 0.571 & 0.511 & 0.1792 & 0.0150 & 0.0095\\
Bounce & Cosmos null & 1/32 & 0.558 & 0.436 & 0.1792 & 0.0112 & 0.0098\\
Bounce & Cosmos text & 1/2 & 0.653 & 0.659 & 0.0080 & 0.0095 & 0.0094\\
Bounce & Cosmos text & 1/4 & 0.713 & 0.639 & 0.0073 & 0.0074 & 0.0085\\
Bounce & Cosmos text & 1/8 & 0.706 & 0.585 & 0.0069 & 0.0069 & 0.0095\\
Bounce & Cosmos text & 1/16 & 0.696 & 0.511 & 0.0067 & 0.0065 & 0.0095\\
Bounce & Cosmos text & 1/32 & 0.662 & 0.436 & 0.0068 & 0.0056 & 0.0098\\
\bottomrule
\end{longtable}
\normalsize

Independent input-render streams yield 144 generated-reference differences
per scene and Cosmos configuration. Their magnitudes are sensitivity controls.
All 24 saved same-input/same-seed generation repeats are exactly identical.
No same-condition true-future rerender or encoded-target repeat is available
to identify a matched physical-effect noise floor. Accordingly, no variance
is subtracted from the physical errors.

The 8-bit code step $q=1/255$ bounds individual channel quantization.
We round each true image to this
grid, then compare its induced contrast with the original float contrast.
We also report the conservative residual
$\|\max(|\Delta Z-\Delta T|-q,0)\|_{\rm RMS}$ after allowing a full code
step per contrast coordinate. At the finest heat scale this residual is
0.00017/0.00175 for Cosmos null/text. The largest-to-smallest error plateau remains an empirical
finite-range observation (Table~\ref{tab:precision}).

\begin{table}[h]
\caption{Finest-scale RGB absolute errors and precision diagnostics. Rerender is generated-reference variation across independent input-render streams. Round-only is the error induced by rounding the true images to 8-bit. Interval lower is the remaining error after allowing one code step per contrast coordinate. All 24 same-input/same-seed repeats are identical. No future-truth rerender or encoded-target repeat is available to identify a matched noise floor.}
\label{tab:precision}
\centering\small
\setlength{\tabcolsep}{3pt}
\begin{tabular}{@{}llrrrr@{}}
\toprule
Scene & Subject & Rerender RMS & Effect error & Round-only & Interval lower\\
\midrule
Spring & Cosmos null & 0.00000 & 0.01166 & 0.00000 & 0.01110\\
Spring & Cosmos text & 0.00000 & 0.01799 & 0.00000 & 0.01718\\
Heat & Cosmos null & 0.00329 & 0.00146 & 0.00033 & 0.00017\\
Heat & Cosmos text & 0.00532 & 0.00359 & 0.00033 & 0.00175\\
Bounce & Cosmos null & 0.00208 & 0.01122 & 0.00000 & 0.01023\\
Bounce & Cosmos text & 0.00278 & 0.00558 & 0.00000 & 0.00477\\
\bottomrule
\end{tabular}
\end{table}

\subsection{Self-Description in the Physical Target Space}
\label{app:samespace}

Both ratios in Figure~\ref{fig:samespace} use the same saved latent observation:
LeWM after \texttt{pred\_proj}, or C-JEPA's consistent paired replay. We
reapply the common fitting procedure.
The available target-space records contain reference and offsets down from
$\pm\delta_0/2$. We therefore fix
$0,\pm\delta_0/2$ as the three references for every scene and both subjects,
and query $\pm\delta_0/j$ for $j=4,8,16,32$. Six context folds keep all
three anchors together. Architecture, training budget, multiplier selection
and full-coordinate scoring are unchanged. Only saved observations are used.
This is a new fit, with no new world-model call or rendering.

Each self-description ratio normalizes by the subject's own response norm.
The physical ratio normalizes by the encoded true effect norm. The figure shows all six scene-interface pairs
at every held-out magnitude, and Table~\ref{tab:samespace} lists the values with their context-block intervals.

\small
\begin{longtable}{@{}llrrrr@{}}
\caption{Self-description and physical correspondence in identical target coordinates: LeWM after its prediction projection, C-JEPA paired replay. References are $0,\pm\log(1.5)/2$, the largest saved target-space offsets. Four finer magnitudes are held out by context. PL and the common neural procedure use the same references. All three ratios retain full coordinates but normalize by their own task's response magnitude. Brackets are context-block 95\% intervals conditional on the fits.}\label{tab:samespace}\\
\toprule
Scene & Subject & Scale & Own: PL & Own: neural & Physical effect\\
\midrule
\endfirsthead
\toprule
Scene & Subject & Scale & Own: PL & Own: neural & Physical effect\\
\midrule
\endhead
Spring & C-JEPA & 1/4 & 0.834 [0.787, 0.861] & 0.827 [0.781, 0.874] & 1.044 [1.028, 1.072]\\
Spring & C-JEPA & 1/8 & 0.916 [0.888, 0.940] & 0.922 [0.887, 0.950] & 1.028 [1.012, 1.048]\\
Spring & C-JEPA & 1/16 & 0.962 [0.946, 0.982] & 0.958 [0.934, 0.977] & 1.025 [1.011, 1.048]\\
Spring & C-JEPA & 1/32 & 0.975 [0.967, 0.990] & 0.969 [0.958, 0.991] & 1.032 [1.018, 1.083]\\
Spring & LeWM & 1/4 & 0.769 [0.702, 0.818] & 0.731 [0.687, 0.770] & 1.056 [1.041, 1.067]\\
Spring & LeWM & 1/8 & 0.969 [0.902, 1.016] & 0.898 [0.861, 0.966] & 1.045 [1.036, 1.054]\\
Spring & LeWM & 1/16 & 1.009 [0.965, 1.059] & 0.950 [0.929, 0.973] & 1.038 [1.029, 1.048]\\
Spring & LeWM & 1/32 & 1.038 [0.996, 1.094] & 0.968 [0.943, 0.988] & 1.033 [1.026, 1.040]\\
Heat & C-JEPA & 1/4 & 0.554 [0.425, 0.600] & 0.554 [0.425, 0.600] & 1.032 [1.021, 1.045]\\
Heat & C-JEPA & 1/8 & 0.754 [0.652, 0.848] & 0.758 [0.652, 0.841] & 1.029 [1.020, 1.047]\\
Heat & C-JEPA & 1/16 & 0.848 [0.737, 0.930] & 0.848 [0.737, 0.929] & 1.027 [1.019, 1.047]\\
Heat & C-JEPA & 1/32 & 0.842 [0.751, 0.925] & 0.844 [0.751, 0.925] & 1.029 [1.015, 1.048]\\
Heat & LeWM & 1/4 & 0.099 [0.088, 0.105] & 0.096 [0.080, 0.103] & 0.973 [0.966, 0.984]\\
Heat & LeWM & 1/8 & 0.193 [0.167, 0.223] & 0.192 [0.164, 0.220] & 0.977 [0.965, 0.986]\\
Heat & LeWM & 1/16 & 0.347 [0.308, 0.385] & 0.347 [0.308, 0.388] & 0.989 [0.979, 0.999]\\
Heat & LeWM & 1/32 & 0.555 [0.481, 0.600] & 0.556 [0.484, 0.594] & 1.013 [0.994, 1.026]\\
Bounce & C-JEPA & 1/4 & 0.819 [0.732, 0.865] & 0.826 [0.734, 0.864] & 1.117 [1.080, 1.197]\\
Bounce & C-JEPA & 1/8 & 0.917 [0.883, 0.939] & 0.910 [0.879, 0.951] & 1.120 [1.066, 1.235]\\
Bounce & C-JEPA & 1/16 & 0.965 [0.942, 0.985] & 0.971 [0.928, 1.009] & 1.121 [1.072, 1.220]\\
Bounce & C-JEPA & 1/32 & 0.981 [0.967, 0.999] & 0.982 [0.958, 1.007] & 1.065 [1.044, 1.130]\\
Bounce & LeWM & 1/4 & 0.748 [0.702, 0.826] & 0.825 [0.776, 0.912] & 1.003 [0.994, 1.020]\\
Bounce & LeWM & 1/8 & 0.930 [0.866, 0.974] & 0.991 [0.936, 1.042] & 1.035 [1.007, 1.049]\\
Bounce & LeWM & 1/16 & 0.976 [0.960, 0.989] & 1.027 [0.994, 1.108] & 1.007 [0.998, 1.034]\\
Bounce & LeWM & 1/32 & 0.991 [0.975, 1.004] & 1.012 [0.989, 1.107] & 1.004 [0.999, 1.017]\\
\bottomrule
\end{longtable}
\normalsize

\section{Mathematical Arguments}
\label{app:proofs}

The exact finite-scale identities below apply to the recorded observations.
Norms use a fixed inner product, including the RMS weighting of each interface.
Derivative recovery requires a first-order remainder for the declared
intervention-to-observation map. Stronger smoothness is needed only for
stronger rates. Native quantization and event transitions are properties of
that map, distinct from additional observation or export error.

\subsection{Absolute Error, Derivatives and Refinement}
\label{app:scalerecovery}

\paragraph{Exact decomposition.}
Fix context, anchor, interface and random realization. For $h>0$, define
$\Delta_\pm=Y(\pm h)-Y(0)$,
$D_h=(\Delta_+-\Delta_-)/(2h)$ and
$K_h=(\Delta_++\Delta_-)/(2h)$. Thus
$\Delta_+=h(D_h+K_h)$ and $\Delta_-=h(-D_h+K_h)$ exactly.
For any coefficient $L$, the errors are $h[(L-D_h)-K_h]$ and
$h[-(L-D_h)-K_h]$. Adding their squared norms cancels the cross terms and
proves Equation~\ref{eq:absolutescale} without assuming orthogonality of
$D_h,K_h$. Retaining the predictor's even component gives
Equation~\ref{eq:neuralpaired}: the errors are
$h[(\widehat D_h-D_h)+(\widehat K_h-K_h)]$ and
$h[-(\widehat D_h-D_h)+(\widehat K_h-K_h)]$, whose squared cross terms cancel.
The paired response energy is
$2h^2(\|D_h\|^2+\|K_h\|^2)$. For the fixed coefficient $D_0=D_{h_0}$
and positive response energy, division yields
\begin{equation}
\mathcal R_h^2=
\frac{\|hD_0-\Delta_+\|^2+\|{-hD_0}-\Delta_-\|^2}
{\|\Delta_+\|^2+\|\Delta_-\|^2}
=\frac{\|D_0-D_h\|^2+\|K_h\|^2}{\|D_h\|^2+\|K_h\|^2}.
\label{eq:pairedscale}
\end{equation}
Subtracting denominator from numerator proves
$\mathcal R_h<1$ iff $2\langle D_0,D_h\rangle>\|D_0\|^2$.
This paired relative error differs from $R^2$ and median signed error.
Zero response leaves the absolute identity valid and the ratio undefined.

\paragraph{Connection to measured physical-effect error.}
For a fixed context and anchor in Section~\ref{sec:scales}, let
$\Delta_\pm=T_W(\pm h)-T_W(0)$ and
$\widehat\Delta_\pm=Z_W(\pm h)-Z_W(0)$, using the same observation
space and RMS norm as Equation~\ref{eq:trueeffect}. When the paired target
energy is positive, Equation~\ref{eq:neuralpaired} gives
\begin{equation}
\mathcal R_{W,h}^2:=
\frac{\|\widehat\Delta_+-\Delta_+\|^2+\|\widehat\Delta_--\Delta_-\|^2}
{\|\Delta_+\|^2+\|\Delta_-\|^2}
=\frac{\|\widehat D_h-D_h\|^2+\|\widehat K_h-K_h\|^2}
{\|D_h\|^2+\|K_h\|^2}.
\label{eq:pairedphysical}
\end{equation}
If both signed target responses are nonzero, this is the mean of
$e_W(h)^2$ and $e_W(-h)^2$, weighted by their true-response energies.
Figure~\ref{fig:scales}
reports medians over individual signed queries, so its points are
not substituted into this paired identity. Each magnitude supplies 72
context-anchor pairs from the 144 signed queries. The identity specifies
how their errors can be resolved into two components. The reported aggregate
ratios alone do not determine either contribution. A zero-effect predictor
sets $\widehat D_h=\widehat K_h=0$ and has $\mathcal R_{W,h}=1$.
The same ratio can arise from a nonzero response, for example
$\widehat\Delta_\pm=2\Delta_\pm$. Relative effect error therefore
needs amplitude and direction comparisons to diagnose its source.

\paragraph{An exact tolerance test and an odd-predictor floor.}
For $V_h=\|D_h\|^2+\|K_h\|^2>0$ and $0<\tau<1$, rearranging
Equation~\ref{eq:pairedscale} gives
\begin{equation}
\mathcal R_h\le\tau\quad\Longleftrightarrow\quad
\|D_0-D_h\|^2+(1-\tau^2)\|K_h\|^2\le\tau^2\|D_h\|^2.
\label{eq:exactscaletolerance}
\end{equation}
Any odd predictor on this pair has outputs $\pm hL$ for some vector $L$.
Minimizing the numerator over $L$ sets $L=D_h$, hence
\begin{equation}
\inf_L\frac{\|hL-\Delta_+\|^2+\|-hL-\Delta_-\|^2}
{\|\Delta_+\|^2+\|\Delta_-\|^2}
=\frac{\|K_h\|^2}{V_h}=E_h.
\label{eq:oddresponsefloor}
\end{equation}
Thus $E_h\le\tau^2$ is necessary within the odd class, even with an
evaluation-side optimum. The corresponding minimum absolute error is
$h\|K_h\|$. Predictors with an even term are not subject to this restriction.

\paragraph{An equidistant reference case.}
Suppose the five outputs at $0,\pm h_0,\pm h$, with $0<h<h_0$, have
common pairwise distance $\ell>0$. For distinct nonreference indices $i,j$,
polarization gives
$\langle Y_i-Y_0,Y_j-Y_0\rangle=\ell^2/2$, while each contrast has
squared norm $\ell^2$. Expanding the sums and differences therefore yields
\begin{equation}
\begin{aligned}
\|D_h\|^2&=\frac{\ell^2}{4h^2},&
\|K_h\|^2&=\frac{3\ell^2}{4h^2},\\
\langle D_{h_0},D_h\rangle&=0,&
\langle K_{h_0},K_h\rangle&=\frac{\ell^2}{2h_0h}.
\end{aligned}
\label{eq:equidistantparity}
\end{equation}
Consequently $E_h=3/4$, $\cos(K_{h_0},K_h)=2/3$ and the original
coarse rule has $\mathcal R_h^2=1+h^2/(4h_0^2)$. These equalities
describe a finite distance configuration.
Even-component alignment alone therefore cannot establish smooth curvature.

\paragraph{A derivative requires agreement of both sides.}
The one-sided increasing-parameter slopes are $D_h+K_h$ and $D_h-K_h$.
Both tend to $J$ exactly when $D_h\to J$ and $K_h\to0$, proving the
main-text characterization of differentiability, including $J=0$.
For $Y(s)=|s|$, $D_h=0,K_h=1$. For $Y(s)=s^2$, $D_h=0,K_h=h$.
The former has a cusp, whereas the latter correctly has zero derivative.
For a fixed coefficient and nonzero $J$, the relative error tends to
\begin{equation}
\lim_{h\to0}\mathcal R_h=\frac{\|D_0-J\|}{\|J\|}.
\label{eq:coarsebiaslimit}
\end{equation}
This is a property of keeping $D_0$ fixed. No division by $\|J\|$ is needed for absolute recovery.

\paragraph{First-order remainder and updated calibration.}
Assume $Y(s)=Y(0)+sJ+R(s)$ and
$\|R(s)\|\le |s|\omega(|s|)$ for $|s|\le r$, with nondecreasing
$\omega$ tending to zero. Equation~\ref{eq:absolutescale} with $L=J$ gives
\begin{equation}
\mathcal B_h(J)^2=\|D_h-J\|^2+\|K_h\|^2
=\frac{\|R(h)\|^2+\|R(-h)\|^2}{2h^2}\le\omega(h)^2.
\end{equation}
The triangle inequality in the product space gives, for $2h\le r$,
$\mathcal B_h(D_{2h})\le\mathcal B_h(J)+\|D_{2h}-J\|
\le\omega(h)+\omega(2h)$, proving Equation~\ref{eq:modulusrefinement}.
If $Y$ is absolutely continuous and
$\|Y'(s)-J\|\le H|s|^\alpha$ almost everywhere, $0<\alpha\le1$,
integration gives $\omega(h)=Hh^\alpha/(1+\alpha)$.
Thus derivative error is $O(h^\alpha)$ and updated response error is
$O(h^{1+\alpha})$. For $Y(s)=s+\operatorname{sign}(s)|s|^{1+\alpha}$,
$D_h=1+h^\alpha$, so central differencing does not supply a second-order
rate under only these assumptions.

\paragraph{Finite steps and a continuum limit.}
For $h_j=h_0 2^{-j}$, let $b_j=\|D_{h_{j+1}}-D_{h_j}\|$.
If $\sum_jb_j<\infty$, the slopes are Cauchy and their limit $J_*$ obeys
$\|D_{h_j}-J_*\|\le\sum_{k\ge j}b_k$. A bound
$b_{k+1}\le qb_k$ for every subsequent $k$, $0\le q<1$, makes this tail
at most $b_j/(1-q)$. A fitted finite-prefix ratio does not establish that premise.
Even exact dyadic agreement can miss between-scale variation:
$Y(s)=s\sin(2\pi\log_2(|s|/h_0))$, $Y(0)=0$, is locally Lipschitz,
has $D_{h_j}=K_{h_j}=0$, and is not differentiable at zero.
Extending sampled convergence therefore also requires control between steps.
These examples limit continuum inference.

\subsection{Finite Probes and Recovered Laws}
\label{app:finiteproberecovery}

\paragraph{A deterministic derivative certificate.}
Let $F:U\subset\mathbb R^m\to\mathbb R^d$ satisfy
$\|F(u_0+v)-F(u_0)-Jv\|\le\|v\|\omega(\|v\|)$ uniformly on a
neighborhood, with $J$ linear and $\omega$ as above. Use $n$ directions
$\|a_i\|\le1$ and step $h_0>0$ within that neighborhood.
Set $A=[a_1\ \cdots\ a_n]$ and
$\kappa_n=\lambda_{\min}(AA^\top/n)>0$.
If each measured endpoint has norm error at most $\nu$, form central slopes
$S_i$ and $\widehat J=[S_1\ \cdots\ S_n]A^\top(AA^\top)^{-1}$.
Then
\begin{equation}
\|\widehat J-J\|_{\operatorname{op}}
\le\beta_F:=\frac{\omega(h_0)+\nu/h_0}{\sqrt{\kappa_n}}.
\label{eq:finitejacobian}
\end{equation}
Each slope equals $Ja_i+r_i$, with
$\|r_i\|\le\omega(h_0)+\nu/h_0$. The residual matrix norm is at most
$\sqrt n$ times this bound, whereas
$\|A^\top(AA^\top)^{-1}\|_{\operatorname{op}}=1/\sqrt{n\kappa_n}$.
Their product proves the claim. The unit-direction budget implies
$\kappa_n\le1/m$, and rank deficiency prevents full-Jacobian recovery without
extra structure. At fixed directional coverage, additional measurement error
must satisfy $\nu(h_0)/h_0\to0$ for this bound to vanish. Native quantization
belongs to $F$. The recorded
runs supply no uniform $\nu$ or $\omega$.

\paragraph{From recovered tangents to finite responses.}
Let $Z$ be an observation and $T$ a target at the same context, intervention
coordinate and declared observation times. Suppose their derivative estimates
have bounds $(\beta_Z,\beta_T)$ from Equation~\ref{eq:finitejacobian}.
For a fixed linear response readout $W$, put
\begin{equation}
\gamma_W=\|\widehat J_T-W\widehat J_Z\|_{\operatorname{op}}
+\beta_T+\|W\|_{\operatorname{op}}\beta_Z.
\end{equation}
If both maps have first-order remainder moduli $\omega_T,\omega_Z$ on a
common neighborhood, then for $\|v\|\le1$ and $h>0$ within it,
\begin{equation}
\|\Delta T(hv)-W\Delta Z(hv)\|
\le h[\gamma_W+\omega_T(h)+\|W\|_{\operatorname{op}}\omega_Z(h)].
\label{eq:finiterecovery}
\end{equation}
Triangle inequality first bounds the true derivative discrepancy by
$\gamma_W$, then the two remainder bounds give the result.
For self-description, take $T=Y$, the exactly known $Z(u)=u$, and
$W=\widehat J_Y$. The error is at most $h[\beta_Y+\omega_Y(h)]$.
Absolute-output accuracy also needs the baseline. A physical target requires
declared common units and observation times.

\subsection{Energy and Predictor Classes}

\paragraph{Scores of complementary observables.}
Let $P$ project onto the training-derived response subspace and $I-P$ onto
its orthogonal complement, denoted $\parallel$ and $\perp$. Applying the same
projector to targets and predictions gives, by Pythagoras,
$\mathrm{SSE}=\mathrm{SSE}_{\parallel}+\mathrm{SSE}_{\perp}$. Compatible
reference centering likewise gives
$\mathrm{SST}=\mathrm{SST}_{\parallel}+\mathrm{SST}_{\perp}$.
When both component SSTs are positive, substituting
$R^2=1-\mathrm{SSE}/\mathrm{SST}$ proves
\begin{equation}
R^2_{\rm full}=\eta R^2_{\parallel}+(1-\eta)R^2_{\perp},
\qquad \eta=\mathrm{SST}_{\parallel}/\mathrm{SST}_{\rm full}.
\label{eq:r2energy}
\end{equation}
The whole-response score is a variation-weighted average of its component
scores. Summing foldwise sums of squares retains
the identity when each fold has its own training-derived projector.
If a component SST vanishes, its $R^2$ is undefined. Retain the undivided
SSE/SST decomposition. Total
$R^2$ is undefined when the full SST vanishes.

\paragraph{Error outside the calibration span.}
Let $v=Y(s)-Y(0)$ and let $\widehat v$ be any predictor, including the
neural response tool. For an orthogonal projector $P$, Pythagoras gives
\begin{equation}
\|\widehat v-v\|^2=\|P(\widehat v-v)\|^2+
\|(I-P)(\widehat v-v)\|^2.
\label{eq:projection}
\end{equation}
Only when $P\widehat v=\widehat v$ does the second term reduce to
$\|(I-P)v\|^2$, a target-only residual floor. PL interpolation satisfies
this restriction for the calibration span, but the neural coordinate residual
need not. We therefore do not apply that special floor to the new neural
scores. Component-normalized ratios also require nonzero component targets.
The undivided identity remains valid when one vanishes. Projection alone
assigns no physical interpretation to either component.

\paragraph{Predictable variation and readout class.}
For a random target $T$ with finite second moment, let
$\mathcal H_0\subseteq\mathcal H_1$ be nested closed linear spaces of
predictor functions under a common data distribution. The projection
$P_{\mathcal H_i}T$ minimizes mean squared error in space $i$. Orthogonality
gives
\begin{equation}
\inf_{f\in\mathcal H_0}\mathbb E\|T-f\|^2-
\inf_{f\in\mathcal H_1}\mathbb E\|T-f\|^2
=\|P_{\mathcal H_1}T-P_{\mathcal H_0}T\|_{L^2}^2.
\end{equation}
Here $\|Z\|_{L^2}^2=\mathbb E\|Z\|^2$ for a random vector $Z$. The identity
expresses the gain from a richer predictor space as the squared magnitude of
the additional predictable component. It concerns population optima in nested
closed linear spaces.

\subsection{Continuity, Derivatives, and Joint Changes}

\paragraph{The implemented response.}
At fixed context and realization, let $G$ map source inputs through
the adapter, model, and observation extraction. For Lipschitz constants
$L_G,L_\kappa$, the composition $F=G\circ\kappa_h$ satisfies
$\|F(u)-F(v)\|\le L_GL_\kappa\|u-v\|$. Apply the bound for $G$ to the
two compiled inputs and then the bound for $\kappa_h$. Subtracting a common
reference preserves the result. Where both maps are differentiable on suitable
neighborhoods, sensitivity follows the chain rule:
\begin{equation}
D_uF(u)=D_xG(\kappa_h(c,u))\,D_u\kappa_h(c,u).
\label{eq:sensitivity}
\end{equation}
Here $D_u$ and $D_x$ differentiate with respect to the intervention and
input coordinates. Both stimulus construction and model inference contribute
to the observed sensitivity. For example, with input quantization
$\kappa(u)=q\operatorname{round}(u/q)$ and downstream identity $G(x)=x$,
both inputs at $\pm\delta$ coincide when $0<\delta<q/2$. Their measured
secant is zero although the ideal unquantized response has derivative one.
Experimental resolution has merged the inputs before they reach inference.
For stochastic subjects, this construction describes a fixed realization.
Distributional continuity uses a distance between conditional output laws.

\paragraph{From a finite net to a domain.}
At fixed context, suppose $F$ and $\widehat F$ are respectively
$L$- and $\widehat L$-Lipschitz on a declared intervention domain.
Let a $\rho$-net place a sample within $\rho$ of every domain point,
with error at most $\epsilon$ simultaneously at all samples. Then
\begin{equation}
\sup_u\|F(u)-\widehat F(u)\|\le\epsilon+(L+\widehat L)\rho.
\label{eq:net}
\end{equation}
To prove this, choose a net point $v$ within $\rho$ of any $u$.
The three terms
$\|F(u)-F(v)\|$, $\|F(v)-\widehat F(v)\|$, and
$\|\widehat F(v)-\widehat F(u)\|$ are bounded by $L\rho$, $\epsilon$,
and $\widehat L\rho$. Their sum yields the result. Finite samples alone
admit an unsampled narrow peak even in a smooth response. The regularity
constants bound how much variation can lie between samples.

\paragraph{What values constrain about derivatives.}
The uniform norm $\|e\|_\infty$ is the largest value-error norm over the domain.
If it is at most $\epsilon$, the symmetric difference of $e$ over step
$\delta>0$ is bounded by $\epsilon/\delta$. Value accuracy therefore controls
fixed-step differences, with amplification as the step shrinks. The example
$e_n(u)=n^{-1}\sin(nu)$ converges uniformly to zero while $e_n'(0)=1$,
showing why derivative convergence calls for further regularity.

The directions sampled by an experiment matter as well. With time $t$ and
physical parameter $u$, the error $e(t,u)=b(t)(u-u_0)$ vanishes along the
entire trajectory at $u_0$ and has arbitrary intervention derivative $b(t)$.
For two parameters, $e(u_1,u_2)=\gamma u_1u_2$ vanishes along both coordinate
axes and has arbitrary mixed derivative $\gamma$. Temporal accuracy and
single-axis accuracy constrain their respective slices. Joint interventions
provide observations of the interaction.

\section{Independent Temporal, Visibility and Material Experiments}
\label{app:historical}

The revised neural analysis fits discovery observations and evaluates the saved
fits on validation and confirmation contexts (Appendix~\ref{app:neuralstages}).
The studies below retain their original estimators, experimental criteria and
saved results, separately from that cross-stage evaluation. Earlier
sparse-component and ridge-atlas appearance analyses also remain in the
historical records, and Appendix~\ref{app:viewpoint} rescores the
moved-camera renders with the neural fits. None of the studies below is
relabeled as a result of the three-reference neural tool.

The collision scene and all four of its axes remain in the complete-response
atlas. Its historical rank-one/rank-eight analysis uses training-fold
level-mean response directions and ridge readouts, with separate component,
complement and full-width scores. Component predictability answers a
different question from complete-response prediction, and even its full-width
readout uses a different fitting and calibration procedure. Those scores
are therefore kept as separate historical evidence.

\subsection{Temporal Windows and Event Opportunities}
\label{app:temporal}

The bounce study observes each physical branch through two histories: one
ending just before the first contact, and another ending at the second
contact. Restitution cannot affect motion in the first history, but can in
the second. Gravity and initial height can affect both. The registered
readouts therefore test supported post-contact restitution, unsupported
pre-contact restitution, and supported pre-contact gravity and height.
C-JEPA, both AC interfaces, and Cosmos tokenizer meet that pattern in
discovery and in the independent eight-context validation and confirmation
stages. The generated-RGB configurations do not have independent-stage
confirmation at that observation interface.

The temporal window of readability also depends on the model interface.
For bounce gravity, the original LeWM encoder readout has support only in
the active window, whereas its predictor has support only in the event-free
window. In the original atlas's 15 eligible axes, AC encoder and predictor
window labels agree on all 15, while LeWM's agree on nine. These are separate
within-window decisions. They show
why a law's temporal applicability cannot be inferred from its scene identity
or from a single aggregate response score. The adjacent-frame RGB comparison
in Section~\ref{sec:relations} addresses what is visible at a selected
observation instant.

\subsection{Visibility Dose}
\label{app:visibility}

The bounce visibility study uses 12 contexts and saves 2,508 runs at each of
seven subjects. Five rungs set the ball's
measured Weber contrast to 0, 0.075, 0.15, 0.225 and 0.3, and a point-light
projection keeps its shadow in every rung. The zero rung removes the ball
and keeps the shadow. Post-contact restitution is read from each
interface's response by ridge regression that holds out whole contexts in
eight folds, with a within-context permutation null of 200 draws. These
original readouts are separate from the neural fits.
Table~\ref{tab:visibility} reports every interface at every rung.

\begin{table}[h]
\caption{Bounce visibility dose for post-contact restitution. Entries are held-out ridge $R^2$ with within-context permutation $p$ values from 200 draws in parentheses. Columns give the ball's measured Weber contrast, with its shadow kept at every rung, and Native is the unmodified rendering. The zero rung removes the ball. Each entry uses 96 queries from 12 contexts in eight context folds. Cosmos null and text use the first generated RGB frame. $^\dagger$Prediction spread below 1\% of the target spread.}
\label{tab:visibility}
\centering\scriptsize
\setlength{\tabcolsep}{2.5pt}
\begin{tabular}{@{}lrrrrrr@{}}
\toprule
Observation & 0 & 0.075 & 0.15 & 0.225 & 0.3 & Native\\
\midrule
VideoSAUR enc. & -0.002$^\dagger$ (0.791) & -0.030 (0.990) & 0.004 (0.025) & -0.002$^\dagger$ (0.826) & -0.349 (1.000) & -0.538 (1.000)\\
C-JEPA pred. & -0.240 (0.995) & -0.287 (0.995) & 0.529 (0.005) & 0.590 (0.005) & 0.565 (0.005) & 0.616 (0.005)\\
LeWM enc. & -0.000$^\dagger$ (0.473) & 0.000$^\dagger$ (0.015) & 0.000$^\dagger$ (0.005) & 0.000$^\dagger$ (0.005) & 0.000$^\dagger$ (0.005) & 0.001$^\dagger$ (0.020)\\
LeWM pred. & -0.002$^\dagger$ (1.000) & 0.002$^\dagger$ (0.010) & 0.015 (0.005) & 0.036 (0.005) & 0.076 (0.005) & 0.217 (0.005)\\
AC enc. & 0.230 (0.005) & 0.707 (0.005) & 0.660 (0.005) & 0.674 (0.005) & 0.648 (0.005) & 0.615 (0.005)\\
AC pred. & 0.193 (0.005) & 0.662 (0.005) & 0.610 (0.005) & 0.629 (0.005) & 0.614 (0.005) & 0.585 (0.005)\\
Cosmos tokenizer & 0.778 (0.005) & 0.897 (0.005) & 0.910 (0.005) & 0.914 (0.005) & 0.912 (0.005) & 0.876 (0.005)\\
Cosmos null & -1.540 (1.000) & 0.011 (0.010) & -0.251 (1.000) & -0.420 (1.000) & -0.733 (1.000) & -0.544 (1.000)\\
Cosmos text & -0.926 (1.000) & -0.323 (1.000) & -1.770 (1.000) & -0.621 (1.000) & -0.362 (0.995) & -1.083 (1.000)\\
\bottomrule
\end{tabular}
\end{table}

\subsection{Material Futures after Exact Reset}
\label{app:material}

The viscoelastic study separates a visible-compression task from a
material-dependent future task. Frames 0--13 evolve under branch-specific
damping $\beta$. Frames 14--15 reset finite-element positions and velocities
to a common state while preserving that damping. Earlier recovery motion
can reveal the material. The target is the eight-step height response in
frames 16--23. A fixed ridge
readout with $\alpha=1$ predicts this target from the model observation.
The benchmark is the strongest registered RGB-history, dense-flow-history,
or flow-warped-rollout readout. Positive controls, including true height
history and future-supervised features, establish learnable future information.

The preregistered 24-context extension uses three compression settings with
eight appearance contexts each. C-JEPA and AC satisfy the visible-compression
readout and response criteria, with forward $R^2=0.125$ and 0.031. Their
material-future $R^2$ values are $-0.066$ and 0.339, below the 0.816616
benchmark by more than 0.02, with Holm-adjusted $p=0.0004$ for both
inferiority comparisons. Reset-only and pre-opportunity upper bounds are
below 0.05. AC retains readable $\beta$, whereas C-JEPA's one-sided
95th-percentile readout bound is below 0.05. Thus visible response structure
and success on the material-future criterion dissociate within this experiment.
The fixed readout limits the conclusion: failure is not proof that no
nonlinear readout could recover the relevant material information.

Stages remain distinct. The initial eight-context validation used an
incorrect oracle-failure criterion and was instrument-invalid. The subsequent
eight-context confirmation established only AC's physical-arm shortfall.
The 24-context extension established the declared contrast at both interfaces.
These results are retained under their original procedures.

\section{Single-Frame Reference Comparisons}
\label{app:simulatorreference}

RGB is another subject, measured through a complete single image. Last-input
and first-future RGB provide separate reference observations. The latter
is the common comparator in Table~\ref{tab:scenes}. Baseline and intervention
images always use the same frame index.
All observations cover the same 24 physical axes. Branch-specific atlas
anchors and fixed fine-scale times are retained. Native tubelets and VAE
units keep their temporal support.

\paragraph{Common evaluation.}
Subjects share intervention coordinates, sample ordering, outer folds,
calibration-only normalization, training rules, budgets, scoring and controls
(Appendix~\ref{app:estimators}). Geometry uses identical distances, signed
weights and zero handling. Native forward retains each subject's complete
three-reference observations, while reverse retains the common physical
parameter target. The 48 RGB and 216 learned units give 1,320 procedure scores.
These discovery comparisons use the same 50,512 eligible query records per
procedure. External-stage results use the same frozen fits and are reported
separately in Appendix~\ref{app:neuralstages}. Differences are descriptive.

\paragraph{Applying RGB-fitted descriptions to other outputs.}
Each outer fold trains and selects the reverse network $g_R$ exclusively from
future RGB. The common geometric descriptor $\Psi$ maps different output
dimensions to the same nineteen inputs, so a target subject $W$ receives
\begin{equation}
\widehat u_W=g_R\!\left(\Psi(\mathcal D_c^W,Y_W)\right).
\label{eq:rgbtransfer}
\end{equation}
Its deterministic normalization uses the three local references. It
normalizes intervention extent and response magnitude. We restore the exact
source network and multiplier for every target. The target's query
observation enters as reverse input. Its parameter is used only for scoring.
The transfer includes all eleven
observations and 24 axes, with no target-side regression or hyperparameter
selection. The 24 RGB self-transfer scores equal the native reverse scores
exactly, and all fold weights are retained.

\begin{table}[h]
\caption{RGB-trained correction beyond the target's own three-reference inverse. PL and RGB columns are median $R^2$ scores. The paired skill $S_{RGB}=1-\mathrm{SSE}_{RGB}/\mathrm{SSE}_{PL}$ isolates the effect of adding the frozen RGB correction. Brackets are 95\% scene/context-block percentile intervals conditional on saved fits. Wins are paired axes.}
\label{tab:alignedreference}
\centering\small
\setlength{\tabcolsep}{3pt}
\begin{tabular}{@{}lrrrr@{}}
\toprule
Observation & Target PL $R^2$ & RGB $R^2$ & $S_{RGB}$ [95\% interval] & Wins\\
\midrule
VideoSAUR enc. & -0.280 & -0.334 & -0.075 [-0.134, 0.016] & 10/24\\
C-JEPA pred. & 0.148 & -0.061 & -0.035 [-0.141, 0.009] & 8/24\\
LeWM enc. & 0.451 & 0.158 & -0.019 [-0.187, 0.142] & 12/24\\
LeWM pred. & 0.716 & 0.645 & -0.082 [-0.524, 0.151] & 11/24\\
AC enc. & 0.876 & 0.780 & -0.056 [-0.551, 0.066] & 11/24\\
AC pred. & 0.859 & 0.721 & 0.008 [-0.510, 0.121] & 12/24\\
Cosmos tokenizer & 0.920 & 0.914 & 0.033 [-0.108, 0.246] & 14/24\\
Cosmos null & 0.694 & 0.768 & 0.102 [0.018, 0.262] & 19/24\\
Cosmos text & 0.340 & 0.266 & 0.036 [-0.019, 0.123] & 13/24\\
RGB last input & 0.741 & 0.873 & 0.447 [0.206, 0.652] & 23/24\\
RGB first future & 0.913 & 0.972 & 0.623 [0.528, 0.753] & 24/24\\
\bottomrule
\end{tabular}
\end{table}

\begin{figure}[p]
\centering
\includegraphics[width=\linewidth]{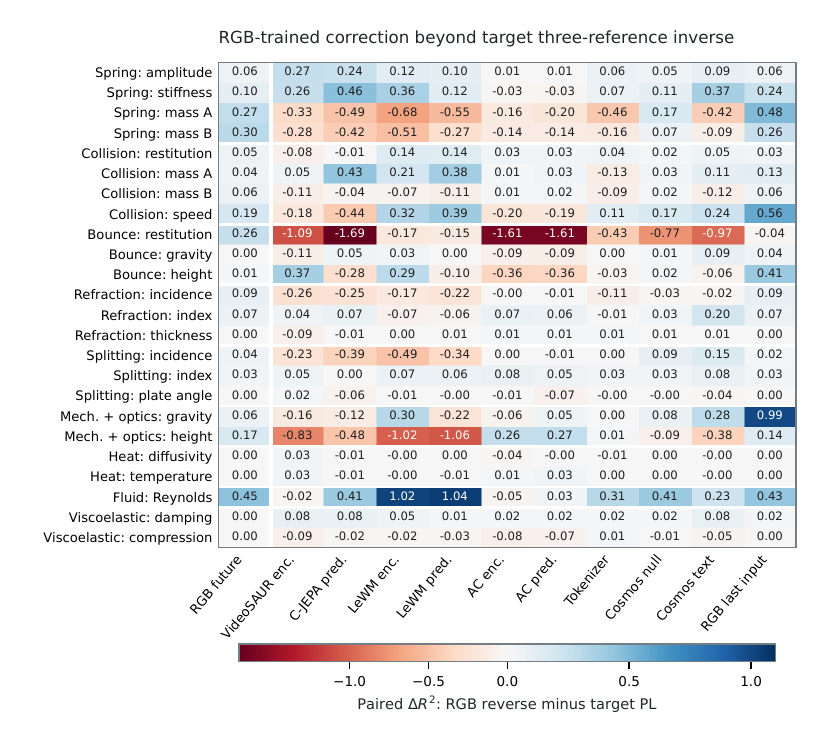}
\caption{Paired axis-wise $\Delta R^2$ from adding the frozen RGB correction
to each target's own PL inverse. Positive entries improve the target PL.
Negative entries worsen it. The source RGB column is a self-check. All
axes and negative values are retained. High absolute reverse scores alone
do not establish a transferred correction.}
\label{appfig:rgbgain}
\end{figure}

In discovery, the source's median reverse score is 0.972. At the learned interfaces,
negative transfer scores occur on 22/24 axes for VideoSAUR, 13/24 for
C-JEPA, 11/24 for LeWM encoder, and 4/24 for LeWM predictor. AC
encoder/predictor have 2/24 and 3/24. Cosmos tokenizer has none, null
has 1/24, and text has 9/24. These counts are descriptive and retain all
axes. A negative score means the frozen readout loses to the declared
training-mean reference.

\paragraph{Paired transfer contrasts.}
Cosmos null improves on its target PL inverse on 19/24 axes. Last-input RGB
has paired skill 0.447 $[0.206,0.652]$. Null's paired $R^2$ advantage over
text is 0.300 $[0.100,0.728]$, but its relative-skill advantage is only
0.023 $[-0.029,0.191]$. Their raw contrast therefore does not isolate
superior transfer of the learned correction.

\paragraph{Supplemented observations and coverage.}
The first future frame uses the original physical conditions and sampling
interval. Fluid continuation first reproduces all 406 saved states exactly.
Viscoelastic recovery rerenders all 504 baseline/intervention conditions
from saved finite-element states, using the same rendering procedure for
both RGB observations. Historical pixel equality is not required.
All nine scenes have complete last-input and first-future references.
The external-stage supplement adds 3,984 matched conditions, each with both
RGB observations, covering all 24 axes in validation and confirmation.
It reuses the original physical settings, observation times and renderer
versions. Viscoelastic reference and intervention images follow the same
paired-rerender policy as discovery. No new world-model inference or response
estimator training is required for the cross-stage evaluation.
The direct future comparison additionally encodes the existing true frames
and replays C-JEPA consistently as specified in Appendix~\ref{app:directfuture}.
Original inputs, archived outputs and independent experiments remain unchanged.

\end{document}